\documentclass[10pt,journal,compsoc]{IEEEtran}
\ifCLASSOPTIONcompsoc
  \usepackage[nocompress]{cite}
\else
  \usepackage{cite}
\fi
\ifCLASSINFOpdf
\else
\fi

\usepackage{amsmath,amsfonts}
\usepackage{algorithmic}
\usepackage{algorithm}
\usepackage{array}
\usepackage{tabularx}
\usepackage[caption=false,font=normalsize,labelfont=sf,textfont=sf]{subfig}
\usepackage{textcomp}
\usepackage{stfloats}
\usepackage{url}
\usepackage{verbatim}
\usepackage{graphicx}
\usepackage{cite}
\usepackage{xcolor}

\usepackage{lipsum}

\usepackage[utf8]{inputenc} % allow utf-8 input
\usepackage[T1]{fontenc}    % use 8-bit T1 fonts
\usepackage{hyperref} 
\hypersetup{
hidelinks,
colorlinks=true,
citecolor=blue,
urlcolor=purple
}
\usepackage{url}            % simple URL typesetting
\usepackage{booktabs}       % professional-quality tables
\usepackage{nicefrac}       % compact symbols for 1/2, etc.
\usepackage{microtype}      % microtypography
\usepackage{braket}

\usepackage{float}
\usepackage{times}
\usepackage{multirow}
\usepackage{amstext}
\usepackage{amssymb}
\usepackage{bm}
\usepackage{caption}
\usepackage{epsfig}
\usepackage{epstopdf}
\usepackage{multirow}
\usepackage{multicol}
\usepackage{eucal}%
\usepackage{makecell}%bold the line
\usepackage{fnbreak} %
\usepackage{xcolor} %
\usepackage{hyperref} 
\usepackage[marginal]{footmisc}
\usepackage{enumerate}

\usepackage{eso-pic}
\hypersetup{pdftitle={Tensor Networks Meet Neural Networks: A Survey and Future Perspectives},
pdfauthor={Maolin Wang, Yu Pan, Zenglin Xu, Guangxi Li, Xiangli Yang, Danilo P. Mandic, Andrzej Cichocki}}
\newcommand{\arxivcopyrightnotice}{%
  \AddToShipoutPictureBG*{%
    \AtPageUpperLeft{%
      \put(\LenToUnit{0.5\paperwidth},\LenToUnit{-5mm}){%
        \makebox[0pt][c]{\parbox[t]{0.86\paperwidth}{\centering\scriptsize
          This article has been accepted for publication in \emph{Proceedings of the IEEE}. This is the author's accepted version; the final published version is available at \url{https://doi.org/10.1109/JPROC.2026.3756230}.\\
          \copyright~2026 IEEE. Personal use of this material is permitted. Permission from IEEE must be obtained for all other uses, in any current or future media, including reprinting/republishing this material for advertising or promotional purposes, creating new collective works, for resale or redistribution to servers or lists, or reuse of any copyrighted component of this work in other works.}}}}}}

\usepackage{amsthm}
\newcommand{\tbscalar}[1]{\MakeLowercase{#1}}
\newcommand{\tbvector}[1]{\mathbf{\MakeLowercase{#1}}}
\newcommand{\tbmatrix}[1]{\mathbf{\MakeUppercase{#1}}}
\newcommand{\tbtensor}[1]{\bm{\mathcal{\MakeUppercase{#1}}}}
\newcommand{\tbdim}[1]{\MakeUppercase{#1}}

\newcommand{\rwone}[1]{#1}
\newcommand{\rwtwo}[1]{#1}
\newcommand{\rwthree}[1]{#1}
\newcommand{\rwfour}[1]{#1}

\definecolor{AEOrange}{HTML}{000000}
\definecolor{R1Red}{HTML}{000000}
\definecolor{R2Blue}{HTML}{000000}
\definecolor{R3Green}{HTML}{000000}
\newcommand{\aehl}[1]{#1}
\newcommand{\ronehl}[1]{#1}

\newcommand{\rthreehl}[1]{#1}
\begin{document}
\arxivcopyrightnotice
\title{
Tensor Networks Meet Neural Networks: \\
A Survey and Future Perspectives

}

\author{Maolin~Wang,
        Yu~Pan,         Zenglin~Xu{*},~\IEEEmembership{Senior~Member,~IEEE},
        Guangxi~Li,
        Xiangli~Yang,  \\
        Danilo Mandic,~\IEEEmembership{Fellow,~IEEE},
and~Andrzej~Cichocki,~\IEEEmembership{Fellow,~IEEE}% <-this % stops a space
% \thanks{* Equal Contribution}% <-this % stops a space
\thanks{* Corresponding Author}
\IEEEcompsocitemizethanks{\IEEEcompsocthanksitem M. Wang is with the Hong Kong Institute of AI for Science, City University of Hong Kong, HKSAR, China.
E-mail: morin.w98@gmail.com
\IEEEcompsocthanksitem Y. Pan is with the Harbin Institute of Technology Shenzhen, Shenzhen, China.
E-mail: iperryuu@gmail.com
\IEEEcompsocthanksitem Z. Xu is with the Fudan University and the Shanghai Academy of AI for Science, Shanghai, China.
E-mail: zenglin@gmail.com
\IEEEcompsocthanksitem G. Li is with the Quantum Science Center of Guangdong-Hong Kong-Macao Greater Bay Area, Shenzhen, China.
E-mail: gxli2017@gmail.com
\IEEEcompsocthanksitem X. Yang is with the Information Engineering University in Zhengzhou, China.
E-mail: xlyang@std.uestc.edu.cn
\IEEEcompsocthanksitem D. Mandic is with the Department of Electrical and Electronic Engineering, Imperial College London, UK, E-mail: d.mandic@imperial.ac.uk
\IEEEcompsocthanksitem A. Cichocki is with the Systems Research Institute, Polish Academy of Sciences, Newelska 6, 01-447 Warsaw, Poland, and also with Artificial Intelligence Project,  Riken,  103-0027 Tokyo, Japan, E-mail: a.cichocki@riken.jp
}% <-this % stops a space
}

\markboth{Proceedings of the IEEE (Accepted Version)}%
{Wang \MakeLowercase{\textit{et al.}}: Tensor Networks Meet Neural Networks: A Survey and Future Perspectives}





\IEEEtitleabstractindextext{%
\begin{abstract}

Tensor networks (TNs) and neural networks (NNs) provide complementary approaches to high-dimensional learning. \rwfour{TNs represent high-order arrays as contractions of smaller cores and can reduce otherwise prohibitive storage and computation when suitable low-rank structure is present. NNs, by contrast, learn flexible nonlinear representations through architectural inductive biases and data-driven optimization. Their connection does not arise from multilinearity alone; rather, TNs provide a principled framework for analyzing and restructuring the multilinear components that occur within neural architectures.} We refer to neural models that incorporate such tensor-network structure as tensorial neural networks~(TNNs). This survey examines TNNs from both data-processing and model-architecture perspectives. The data perspective covers multi-source fusion, multimodal pooling, tensor compression, multi-task learning, and quantum-state data representation. The model perspective considers convolutional and recurrent networks, Transformers, graph neural networks, large language and mixture-of-experts models, and quantum neural networks. We further review training and rank-selection strategies, contraction and hardware considerations, interpretability, and software support. \rthreehl{The central contribution is a cross-architectural synthesis of how tensorization, rank, topology, contraction cost, optimization, inductive bias, and deployment constraints interact across model families.} This synthesis identifies when tensorization functions as compression, structured parameter sharing, or a task-aligned inductive bias, and it motivates concrete directions for reproducible evaluation and future research. We provide a curated list of TNNs at \url{https://github.com/tnbar/awesome-tensorial-neural-networks}.

\end{abstract}

}

\maketitle 
\thispagestyle{empty}% arXiv version: page 1 carries the accepted-version/copyright notice instead of the running header

% is not selected <OR> if conference mode is selected - because compsoc
% conference papers position the abstract like regular (non-compsoc)
\IEEEdisplaynontitleabstractindextext

% \fi
\IEEEpeerreviewmaketitle

\ifCLASSOPTIONcompsoc
\IEEEraisesectionheading{\section{Introduction}\label{sec:introduction}}
\else
\section{Introduction}
\label{sec:introduction}
\fi

\IEEEPARstart{T}{ensors} are \rwthree{multi-dimensional (or equivalently, multi-order)} arrays that represent multiway interactions among multiple modal sources. In contrast, vectors (i.e., first-order tensors) and matrices (i.e., second-order tensors) are accessed in only one or two modes, respectively. As a common data type, tensors have been widely observed in several scenarios
~\cite{cyganski1985applications,koniusz2021tensor,tang2016tri,tang2019social}.
For instance, functional magnetic resonance imaging (fMRI) samples are inherently fourth-order tensors that are composed of three-dimensional voxels that change over time~\cite{davidson2013network,acar2017tensor,belyaeva2024learning}. In quantum physics, variational wave functions used to study many-body quantum systems are also high-order tensors~\cite{biamonte2017tensor,rudolph2023synergistic}. For spatiotemporal traffic analysis, road flow/speed information, which is collected from multiple roads over several weeks, can also be structured as a third-order tensor (road segment$\times$day$\times$time of day)~\cite{chen2019bayesian}. 
% Effective analysis of tensors requires the development of models and associated algorithms that can identify the core relations that exist among the different tensor modes. 
However, for higher-order tensors, when the number of modes increases, the total number of elements in the tensors grows exponentially, which is prohibitive for storing and processing tensors,
% Such a phenomenon 
which is also recognized as the ``curse of dimensionality''~\cite{cichocki2016tensor}.
Tensor networks have emerged as an effective approach to mitigate this problem.
% which can be resolved by tensor networks. 

% \quad

\textbf{Tensor Networks (TNs).} TNs~\cite{biamonte2017tensor, cichocki2016tensor,cichocki2017tensor} represent a high-order tensor through contractions of smaller component tensors, also called blocks, factors, or cores. \aehl{When the target admits an appropriate low-rank or low-entanglement structure, this representation can replace exponential storage in the tensor order with a number of parameters that scales polynomially in the mode sizes and bond dimensions. The benefit is therefore conditional on the tensorization, topology, and ranks rather than an unconditional resolution of the curse of dimensionality.} TNs also distribute computation across their cores, enabling structured storage and processing of tensors that would otherwise be prohibitively large.
Commonly used TN formats include CANDECOMP/PARAFAC (CP)~\cite{carroll1970analysis}, Tucker decomposition~\cite{tucker1966some}, Block-term Tucker (BTT) decomposition~\cite{de2008decompositions1}, Matrix Product State (MPS)/Tensor Train (TT) decomposition~\cite{oseledets2011tensor,verstraete2008matrix}, Matrix Product Operators (MPO)~\cite{oseledets2011tensor}, Tensor Ring (TR) decomposition~\cite{zhao2016tensor}, Tree TN/Hierarchical Tucker (HT) decomposition~\cite{kressner2012htucker}, Projected Entangled Pair State (PEPS)/Tensor Grid decomposition~\cite{schuch2007computational}, and multiscale entanglement renormalization~\cite{cincio2008multiscale}.
For the purpose of understanding the interconnected structures of TNs, a TN diagram was developed as a straightforward graphical diagram (which is discussed in Section~\ref{Sec:Tensor Diagrams}).
A TN can provide a theoretical and computational framework for the analysis of some computationally prohibitive tasks. For example, based on the low-rank structures of TNs, Pan \textit{et al.}~\cite{pan2022simulation} were able to solve the quantum random circuit sampling problem in 15 hours using 512 graphics processing units (GPUs); this problem was previously believed to require over 10,000 years on the most powerful classic electronic supercomputer and effectively challenge the quantum supremacy of Google's quantum computer called ``Sycamore''.
Other applications include brain analysis~\cite{brainzhang2019tensor}, 
% quantum chemistry calculation~\cite{sharma2015multireference}, 
% human face clustering~\cite{cao2014robust},
dimensionality reduction~\cite{zare2018extension}, 
% missing value estimation~\cite{liu2012tensor}, 
latent factor analysis~\cite{tensorXuYQ15}, 
subspace learning~\cite{zhang2017low}, etc.
\begin{table*}[t]
\centering
\color{AEOrange}
\small
\setlength{\tabcolsep}{3.5pt}
\renewcommand{\arraystretch}{1.35}
\caption{An overview of TNNs from both data and model perspectives. To reduce redundancy, we report representative rather than exhaustive examples.}
\label{table:TNNs}
\begin{tabularx}{\textwidth}{@{}>{\centering\arraybackslash}m{0.070\textwidth}>{\raggedright\arraybackslash}p{0.170\textwidth}>{\raggedright\arraybackslash}X>{\centering\arraybackslash}p{0.065\textwidth}@{}}
\hline
\textbf{Category} & \textbf{Subcategory} & \textbf{Detailed Models/Techniques} & \textbf{Section} \\ \hline
\multirow{5}{*}{\rotatebox[origin=c]{90}{\footnotesize\bfseries Data Processing}}
& Multi-source Fusion & TFL~\cite{zadeh2017tensor}, LMF~\cite{liu2018efficient}, PTP~\cite{hou2019deep}, InTense~\cite{varshneya2024interpretable} & ~\ref{Sec:TFL} \\ \cline{2-4}
& Multimodal Pooling & MCB~\cite{fukui2016multimodal}, MLB~\cite{kim2016hadamard}, MUTAN~\cite{ben2017mutan}, CTI~\cite{do2019compact} & ~\ref{Sec:VQA} \\ \cline{2-4}
& Data Compression & BNTD~\cite{he2018knowledge}, TensorCodec~\cite{kwon2023tensorcodec}, NeuKron~\cite{kwon2023neukron}, TTHRESH~\cite{ballester2019tthresh} & ~\ref{Sec:Data_Compression} \\ \cline{2-4}
& Multi-task Training & TTMT/TMT~\cite{yang2017deep}, PEPS-like TN~\cite{wang2020concatenated}, GTTN~\cite{zhang2021multi}, MTCN~\cite{duan2020novel} & ~\ref{Sec:Multi-task} \\ \cline{2-4}
& Quantum Data & Quantum State Mapping~\cite{stoudenmire2016supervised,biamonte2017tensor}, Word Quantum Embedding~\cite{ganmorphte12,li2019cnm,zhang2018quantum,miller2020tensor} & ~\ref{sec:quantum1} \\ \hline
\multirow{6}{*}{\rotatebox[origin=c]{90}{\footnotesize\bfseries Model Architecture}}
& CNNs & CP-CNN~\cite{denton2014exploiting}, Tucker-CNN~\cite{DBLP:journals/ijon/PanWX22}, BTT-CNN~\cite{li2017bt}, TT-CNN~\cite{liu2022tt}, TR-CNN~\cite{DBLP:conf/cvpr/WangSEWA18} & ~\ref{Sec:CNN} \\ \cline{2-4}
& RNNs & TT-RNN~\cite{DBLP:conf/icml/YangKT17}, TR-RNN~\cite{DBLP:conf/aaai/PanXWYWBX19}, BTT-RNN~\cite{Ye_2018_CVPR}, HT-RNN~\cite{DBLP:conf/cvpr/YinLLW021}, Conv-TT-LSTM~\cite{DBLP:conf/nips/SuBKHKA20} & ~\ref{Sec:RNN} \\ \cline{2-4}
& Transformers & MPO-Transformer~\cite{DBLP:conf/acl/LiuGZXLW20}, Tucker-Bert~\cite{wangExploringExtremeParameter2021}, Tuformer~\cite{liuTuformerDataDrivenDesign2021}, \rthreehl{Tucker Attention~\cite{klein2026tuckerattention}, TPA~\cite{zhang2025tensornew}} & ~\ref{Sec:transformer} \\ \cline{2-4}
& GNNs & TGNN~\cite{hua2022high}, RTGNN~\cite{zhao2022multi}, Nimble GNN~\cite{yin2022nimble}, \rthreehl{T-GINEE~\cite{wang2026tginee}} & ~\ref{Sec:GNNs} \\ \cline{2-4}
& QNNs and Circuits & MPS Models~\cite{torchmps}, ConvAC~\cite{cohen2016expressive}, ANTN~\cite{chen2023antn}, \rthreehl{Tensor Factorizations/Circuits~\cite{loconte2026circuits}} & ~\ref{Sec:TQNNs} \\ \cline{2-4}
& LLMs and MoE & \rthreehl{TRAC~\cite{ye2026trac}, TD-MoE~\cite{xu2026tdmoe}, LeSTD~\cite{li2026lestd}, TIME~\cite{xu2026time}, ReFTA~\cite{zheng2026refta}} & ~\ref{Sec:LLMs} \\ \hline
\end{tabularx}
\end{table*}
\begin{table*}[t]
\centering
\color{AEOrange}
\small
\setlength{\tabcolsep}{3.5pt}
\renewcommand{\arraystretch}{1.35}
\caption{An overview of TNN utility. We keep representative methods only, which makes the table more readable and reduces citation overload.}
\label{table:TNN-Utility}
\begin{tabularx}{\textwidth}{@{}>{\centering\arraybackslash}m{0.070\textwidth}>{\raggedright\arraybackslash}p{0.180\textwidth}>{\raggedright\arraybackslash}X>{\centering\arraybackslash}p{0.075\textwidth}@{}}
\hline
\textbf{Category} & \textbf{Subcategory} & \textbf{Detailed Models/Techniques} & \textbf{Section} \\ \hline
\multirow{4}{*}{\rotatebox[origin=c]{90}{\footnotesize\bfseries Lifecycle}}
& Stable Training & Mixed Precision~\cite{DBLP:journals/pieee/PanagakisKCONAZ21}, Yu Initialization~\cite{DBLP:conf/icml/0005SLW0X22}, MANGO~\cite{pan2023reusing} & ~\ref{sec:stable-training} \\ \cline{2-4}
& Rank/Structure Search & PSTRN~\cite{li2021heuristic}, TR-RL~\cite{cheng2020novel}, TT-Bayes~\cite{hawkins2021bayesian}, RGTN~\cite{wang2026renormalization}, \rthreehl{MTNL~\cite{zeng2026mtnl}} & ~\ref{Sec:select} \\ \cline{2-4}
& Model Adaptation & \rthreehl{TRAC~\cite{ye2026trac}, TD-MoE~\cite{xu2026tdmoe}, LeSTD~\cite{li2026lestd}, TIME~\cite{xu2026time}, ReFTA~\cite{zheng2026refta}} & ~\ref{Sec:LLMs} \\ \cline{2-4}
& Hardware Speedup & TIE~\cite{DBLP:conf/isca/DengSQLWY19}, TT-Engine~\cite{DBLP:journals/tcad/QuDWCLLLZX22}, ETTE~\cite{gong2023ette}, \rthreehl{LEGO~\cite{xu2026lego}} & ~\ref{sec:hardware-implementation} \\ \hline
\multirow{3}{*}{\rotatebox[origin=c]{90}{\footnotesize\bfseries Toolboxes}}
& Basic Tensor Ops & Tensorly~\cite{DBLP:journals/jmlr/KossaifiPAP19}, Tensor Toolbox~\cite{kolda2006matlab}, Tntorch~\cite{tntorch}, ITensor~\cite{itensor}, TensorNetwork~\cite{roberts2019tensornetwork} & ~\ref{sec:basictensortoolbox} \\ \cline{2-4}
& Deep-Model Tools & Tensorly-Torch~\cite{DBLP:journals/jmlr/KossaifiPAP19}, TedNet~\cite{DBLP:journals/ijon/PanWX22} & ~\ref{sec:tensorneuraltoolbox} \\ \cline{2-4} 
& Quantum Simulators & TensorNetwork~\cite{roberts2019tensornetwork}, ITensor~\cite{itensor}, TeD-Q~\cite{JDquantum}, TensorNetworkQuantumSimulator.jl~\cite{tindall2024tnqsjl} & ~\ref{sec:quantumtoolbox} \\ \hline
{\scriptsize\bfseries Analysis} & Interpretability & MPS explanations~\cite{puljak2025tensor}, tensorized model analysis~\cite{pareja2025tensorization}, \rthreehl{TN-SHAP-G~\cite{heidari2026tnshap}} & ~\ref{sec:interpretability} \\ \hline
\end{tabularx}
\end{table*}
\textbf{Neural Networks (NNs).} NNs are powerful learning structures that enable machines to acquire knowledge from observed data~\cite{hopfield1982neural,rumelhart1988learning}. Deep Neural Networks (DNNs)~\cite{lecun2015deep,schmidhuber2015deep} have revolutionized artificial intelligence by capturing complex patterns from hierarchical data. Representative architectural paradigms include convolutional neural networks (CNNs), recurrent neural networks (RNNs), and Transformers~\cite{lecun2015deep,vaswani2017attention,DBLP:conf/naacl/DevlinCLT19}. Their impact spans computer vision, language, and scientific discovery, including large-scale vision models and AlphaFold2~\cite{dosovitskiy2020image,jumper2021highly}.
The field of natural language processing has witnessed a paradigm shift with the emergence of large language models (LLMs) built upon Transformer architectures~\cite{vaswani2017attention}. Their scale and deployment constraints have made parameter efficiency, compression, and hardware-aware inference central design concerns.
The impact of deep learning continues to expand across numerous scientific and practical domains, underscoring why parameter efficiency and structured model design have become increasingly important.

\textbf{Tensor Networks Meet Neural Networks.} Tensor Networks (TNs) and Neural Networks (NNs), while stemming from distinct scientific foundations, have each demonstrated unique capabilities across diverse domains. Cichocki \textit{et al.}~\cite{cichocki2016tensor} and Mondelli and Montanari~\cite{mondelli2019connection} establish connections between multilinear tensor structure and neural computation. These connections motivate tensorial neural networks (TNNs): neural architectures whose parameters, activations, or data-processing modules are explicitly structured through tensor networks. We argue that integrating TNs with NNs can enhance model performance and sustainability from both data and model perspectives:

{(1) Effective Data Representation:} Accurate modeling of higher-order interactions from multi-source data is critical in advancing performance and promoting responsible AI practices~\cite{biamonte2017tensor}. Integrating TNs within NN frameworks offers a principled way to capture such interactions without materializing the full high-order tensor, which is particularly useful in multimodal, multiview, and multitask settings~\cite{zadeh2017tensor,hou2019deep,ben2017mutan}. Tensorized representations are likewise useful for compression-oriented data models such as TensorCodec~\cite{kwon2023tensorcodec}.

{(2) Compact Model Structures:} TNs also offer a sustainable route to model compression by replacing dense weight tensors with structured low-rank factors~\cite{DBLP:journals/ijon/PanWX22,Ye_2018_CVPR,DBLP:conf/aaai/PanXWYWBX19}. This can dramatically reduce parameters and memory traffic while preserving useful architectural priors. Representative examples include tensorized convolutional and recurrent layers as well as TR-LSTM-style recurrent compression~\cite{DBLP:conf/nips/HayashiYSM19,DBLP:conf/cvpr/WangSEWA18,DBLP:conf/aaai/PanXWYWBX19}.

We refer to this family of approaches that connect TNs with NNs as \textbf{Tensorial Neural Networks (TNNs)}.
Although this combination holds significant promise for sustainable AI by offering efficient parameter compression and structured representations, TNNs also present new training challenges that require careful consideration. These challenges include numerical stability issues during optimization, particularly for high-order tensor operations and decompositions, complex hyperparameter selection especially for determining optimal tensor ranks and network architectures, and hardware acceleration requirements to efficiently handle tensor contractions and parallel computations. Therefore, it is necessary to redesign traditional neural network training techniques to address these TNN-specific challenges. 

\aehl{Memmel \textit{et al.}~\cite{memmel24a} position tensor networks as a methodological asset for Green AI and argue that tensorized models should be assessed not only by accuracy but also by efficiency, inclusivity, and deployment-aware metrics.}
\aehl{While several excellent surveys and book chapters have explored tensor decompositions in deep learning or quantum tensor networks separately, our survey focuses on their \emph{intersection}. We investigate how tensor structures are adapted into neural architectures, what trade-offs recur across architectures, and which lessons follow from the literature on model design and deployment.}

\rthreehl{More specifically, Biamonte and Bergholm~\cite{biamonte2017tensor} emphasize quantum-TN foundations, graphical notation, contraction, and entanglement. Panagakis \textit{et al.}~\cite{DBLP:journals/pieee/PanagakisKCONAZ21} survey tensor methods for representation learning, computer vision, and deep learning, whereas their later book chapter~\cite{panagakis2024tensor_methods_deep_learning} provides a broader methodological account that also includes quantum-network material. Our survey asks a different organizing question: how do tensor-network structures become trainable neural components across CNNs, RNNs, Transformers, GNNs, LLMs, and QNNs, and which design principles recur across these families? Accordingly, our contribution is not a claim that the individual decompositions or applications have not been reviewed, but a cross-architectural synthesis that connects tensorization and topology choices with contraction cost, optimization, software support, inductive bias, and deployment constraints.}

\rthreehl{\textbf{What do we learn from this review?} Beyond cataloguing applications, the surveyed literature supports four recurring messages. First, decomposition design is not only about choosing ranks: mode permutations, tensorization shapes, and topology choices can materially change the compression--accuracy frontier~\cite{ghadiri2023core_shapes,li2022tnls}. Second, the central architectural trade-off is often \emph{expressiveness versus contractibility}: richer loopy structures can represent more complex correlations, but they require substantially better approximate contraction algorithms to remain practical~\cite{slattery2024hyperoptimized}. Third, modern TNN progress is increasingly algorithmic rather than merely applicational, spanning structure search, sampling-based decomposition, and uncertainty-aware tensor modeling~\cite{bharadwaj2024leverage_tt,tao2023upmtd}. Fourth, tensorization should be viewed not only as compression, but also as an inductive bias that can improve robustness and generalization when aligned with the underlying problem structure~\cite{wang2023transformed_lowrank}.}

An overview of both their data processing capabilities and model architectures of TNNs is shown in Table~\ref{table:TNNs}. From the data processing perspective, TNNs demonstrate versatility across multiple domains including: multi-source fusion for integrating heterogeneous data sources, multimodal pooling for efficient feature combination, data compression for reducing storage requirements while preserving information fidelity, multi-task training for simultaneous learning of related objectives, and quantum data processing for handling quantum state representations. From the model architecture perspective, TNNs have been successfully integrated into various deep learning frameworks including CNNs, RNNs, Transformers, GNNs, Large Language Models (LLMs), and Quantum Neural Networks (QNNs), each offering unique advantages in their respective application domains.

The remaining sections are organized as follows. Section~\ref{Sec:pre} provides the fundamentals of tensor notation, tensor diagrams, and TN formats. Section~\ref{Sec:IFTNN} examines TNNs for data representation, fusion, compression, multi-task learning, and quantum-state data. Section~\ref{Sec:CTDNN} discusses tensorized model architectures. Section~\ref{Sec:strategies} reviews training, rank and structure selection, and hardware acceleration. Section~\ref{Sec:Tool} introduces software toolboxes and interpretability methods, followed by future perspectives in Section~\ref{sec:discussion}.

\section{Tensor Basis}

\label{Sec:pre}

\begin{figure}[t]
\centering
\includegraphics[width=0.46\textwidth]{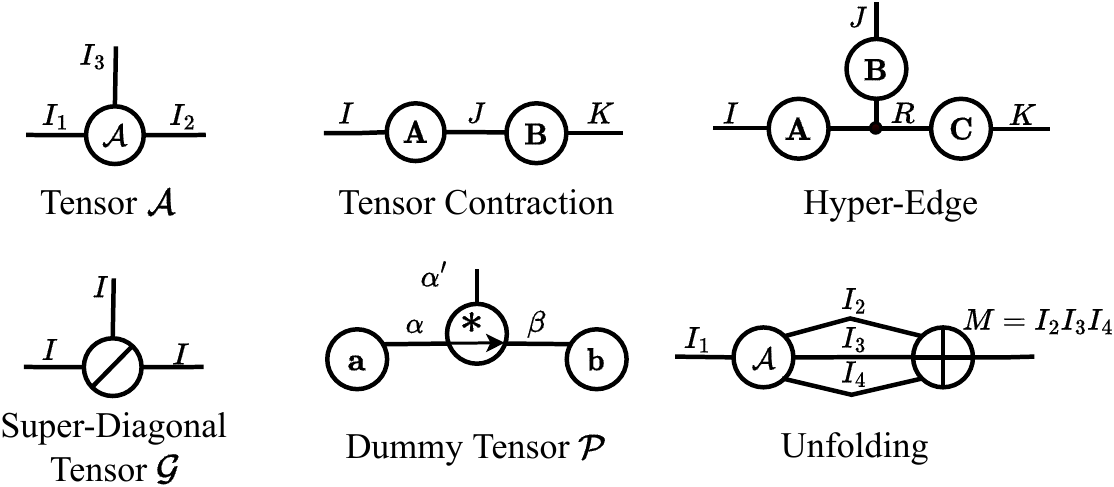}
\caption{
%Editor: Please note that we have not checked the text in this figure or any other figures that are not editable in Word. Please contact our support staff and provide text that can be directly modified in Microsoft Word or PowerPoint for all figure text that you would like revised.
Basic symbols for TN diagrams.
 For more details about TNs, refer to ~\cite{biamonte2017tensor} and ~\cite{cichocki2016tensor}.}
\label{fig:pre:basic_symbol}
\end{figure}

\begin{table}[t]
\centering
\caption{Tensor notations}
\label{table:notation}
\begin{tabular}{c|c}
\hline
Symbol & Explanation \\
\hline
$\tbscalar{a}$      & scalar      \\
$\tbvector{a}$      & vector      \\
$\tbmatrix{a}$      & matrix      \\
$\tbtensor{a}$      & tensor     \\
$\tbdim{a}$      & dimensionality     \\
$\circledast$      & convolution operation     \\
$\circ$      &  outer product operation    \\
$\odot$      &  Hadamard (element-wise) product    \\
$\otimes$      &  Kronecker product / matrix product of factor matrices in TN    \\
$\circledcirc$      &  Khatri-Rao product (column-wise Kronecker product)    \\
 $<\cdot,\cdot>$&  inner product of two tensors   \\
$|\cdot\rangle$      & quantum-state ket vector~(normalized complex column vector)   \\
$\langle\cdot|$      & quantum-state bra vector~(normalized complex row vector)   \\
$\langle\cdot| \cdot \rangle$      & inner product of two quantum state vectors \\
\hline
\end{tabular}
\end{table}

\subsection{Tensor Notations}

A tensor~\cite{xu2012infinite,zhe2016dintucker}, also known as a multiway array, can be viewed as a higher-order extension of a vector (i.e., a first-order tensor) or a matrix (i.e., a second-order tensor). Like the rows and columns in a matrix, an $N$th-order tensor $\bm{\tbtensor X}\in\mathbb R^{I_1\times I_2 \cdots\times I_N}$ has $N$ modes (i.e., ways, orders, or indices) whose lengths (i.e., dimensions) are represented by $I_1$ to $I_N$, respectively.
As shown in Table~\ref{table:notation}, lowercase letters denote scalars, e.g., $\tbscalar{a}$, boldface lowercase letters denote vectors, e.g., $\tbvector{a}$, boldface capital letters denote matrices, e.g., $\tbmatrix{A}$ and boldface Euler script letters denote higher-order tensors, e.g., $\tbtensor A$. In this paper, we define a ``tensor'' in a broad sense that includes scalars, vectors, and matrices.

\subsection{Tensor Diagrams}
\label{Sec:Tensor Diagrams}

In this subsection, we introduce TN diagrams and their corresponding mathematical operations.
TN diagrams were first developed by Roger Penrose~\cite{penrose1971applications} in the early 1970s and are now commonly used to describe quantum algorithms~\cite{biamonte2017tensor,huckle2013computations} and machine learning algorithms~\cite{cichocki2017tensor,stoudenmire2016supervised,DBLP:conf/nips/HayashiYSM19}.
Within these diagrams, tensors are denoted graphically by nodes with edges~\cite{cichocki2017tensor}, which enables intuitive presentation and convenient representation of complex tensors. As both the data and weights in the deep learning field are tensors, tensor diagrams are also promising for use as general network analysis tools in this area. An overview of the basic symbols of tensors is shown in Fig.~\ref{fig:pre:basic_symbol}.

\subsubsection{Tensor Nodes}
A tensor is denoted as a node with edges, as illustrated in Fig.~\ref{fig:pre:basic_symbol}.
The number of edges denotes the modes of a tensor,
and a value on an edge represents the dimension of the corresponding mode. For example, a one-edge node denotes a vector $\mathbf{a} \in \mathbb{R}^{I}$, a two-edge node denotes a matrix $\mathbf{A} \in \mathbb{R}^{I\times J}$ and a three-edge node denotes a tensor $\tbtensor{A}\in \mathbb{R}^{I_1\times I_2\times I_3}$.

%%cpon5order2

% \begin{figure*}[t]
% \centering
% \includegraphics[width=0.91\textwidth]{figure/tensorcube.pdf}
% \caption{Implementation of CP on a 5th-order tensor $\tbtensor{T}\in \mathbb{R}^{I_1\times I_2\times I_3\times I_4\times I_5}$. We show the higher-order tensor in a matrix representation of size $I_1\times I_2$, whose elements are 3rd-order tensors of size $I_2\times I_3\times I_4$. The main idea of CP is to decompose $\tbtensor{T}$ into five factor vectors that encode the modes of $\tbtensor{T}$. Then, it is feasible to apply outer production on these factor vectors to recover $\tbtensor{T}$. Moreover, by regarding such an outer production as a component, CP with a rank of $R$ implements summation on $R$ components to further improve the expressive power of the tensor.}
% \label{fig:cpon5order}
% \end{figure*}

\subsubsection{Tensor Contraction}

Tensor contraction refers to the operation whereby two tensors are contracted into one tensor along their associated pairs of indices. As a result, the corresponding connected edges disappear while the dangling edges persist.
Tensor contraction can be formulated as a tensor product
\begin{align}
    & \tbtensor{C} = \tbtensor{A} \times_{M+1, M+2, \dots, M+N}^{1, 2, \dots N} \tbtensor{B}  
\end{align}
with the elements of $\tbtensor{C}$ are computed via
\begin{align}
    \tbtensor{C}_{p_1, \ldots, p_{K+M}} = \sum_{i_{1},i_{2},\ldots i_{N}} \tbtensor{A}_{i_1,i_2,\ldots i_{N},*}\quad\tbtensor{B}_{*,i_1,i_2,\ldots i_{N}},
\end{align}
where $\tbtensor{A} \in\mathbb{R}^{I_1\times \dots I_N \times P_1 \times \dots  P_K}$, $\tbtensor{B} \in\mathbb{R}^{ P_{K+1} \times  \dots  P_{K+M} \times I_1 \times \dots I_N}$, and $\tbtensor{C} \in \mathbb{R}^{ P_1 \times  \dots  P_K \times P_{K+1} \dots  P_{K+M}}$.
Fig.~\ref{fig:pre:basic_symbol} also shows a diagram of the matrix multiplication operation, which is the most classic tensor contraction situation, given by:
\begin{align}
    \tbmatrix{C} = \tbmatrix{A} \tbmatrix{B} = \tbmatrix{A} \times_{2}^{1} \tbmatrix{B}.
\end{align}
Tensor contractions among multiple tensors (e.g., TNs) can be computed by sequentially performing tensor contractions between each pair of tensors.
It is worth mentioning that the contracting sequence must be determined to achieve better calculation efficiency~\cite{xie2017optimized}.

\subsubsection{Dummy Tensor}

Hayashi \textit{et al.}~\cite{DBLP:conf/nips/HayashiYSM19} introduce a dummy tensor to represent convolution operations. As depicted in Fig.~\ref{fig:pre:basic_symbol}, a node with the star and arrow symbols denotes a dummy tensor. This operation is formulated as
\begin{equation}
\label{eq:dummy}
\mathbf{y}_{j^{'}} 
= \sum_{j=0}^{\alpha-1}\sum_{k=0}^{\beta-1} \tbtensor{P}_{j,j^{'},k}\mathbf{a}_{j} \mathbf{b}_{k},
\end{equation}
where $\mathbf{a}\in\mathbb{R}^{\alpha}$ denotes a vector that will be processed by a convolutional weight $\mathbf{b}\in\mathbb{R}^{\beta}$,
and $\mathbf{y}\in\mathbb{R}^{\alpha '}$ is an output. The symbol
$\tbtensor{P} \in \left\{ 0, 1 \right\}^{\alpha \times {\alpha '} \times \beta}$ denotes a binary tensor with elements defined as $\tbtensor{P}_{j,j^{'},k}=1$ if $j = sj' + k - p$ and $0$ otherwise, where $s$ and $p$ represent the stride and padding size, respectively. Thus, $\tbtensor{P}$ can be applied to any two tensors to induce a well-defined convolutional relationship between them.

\begin{figure*}[t]
\centering

\includegraphics[width=0.8\textwidth]{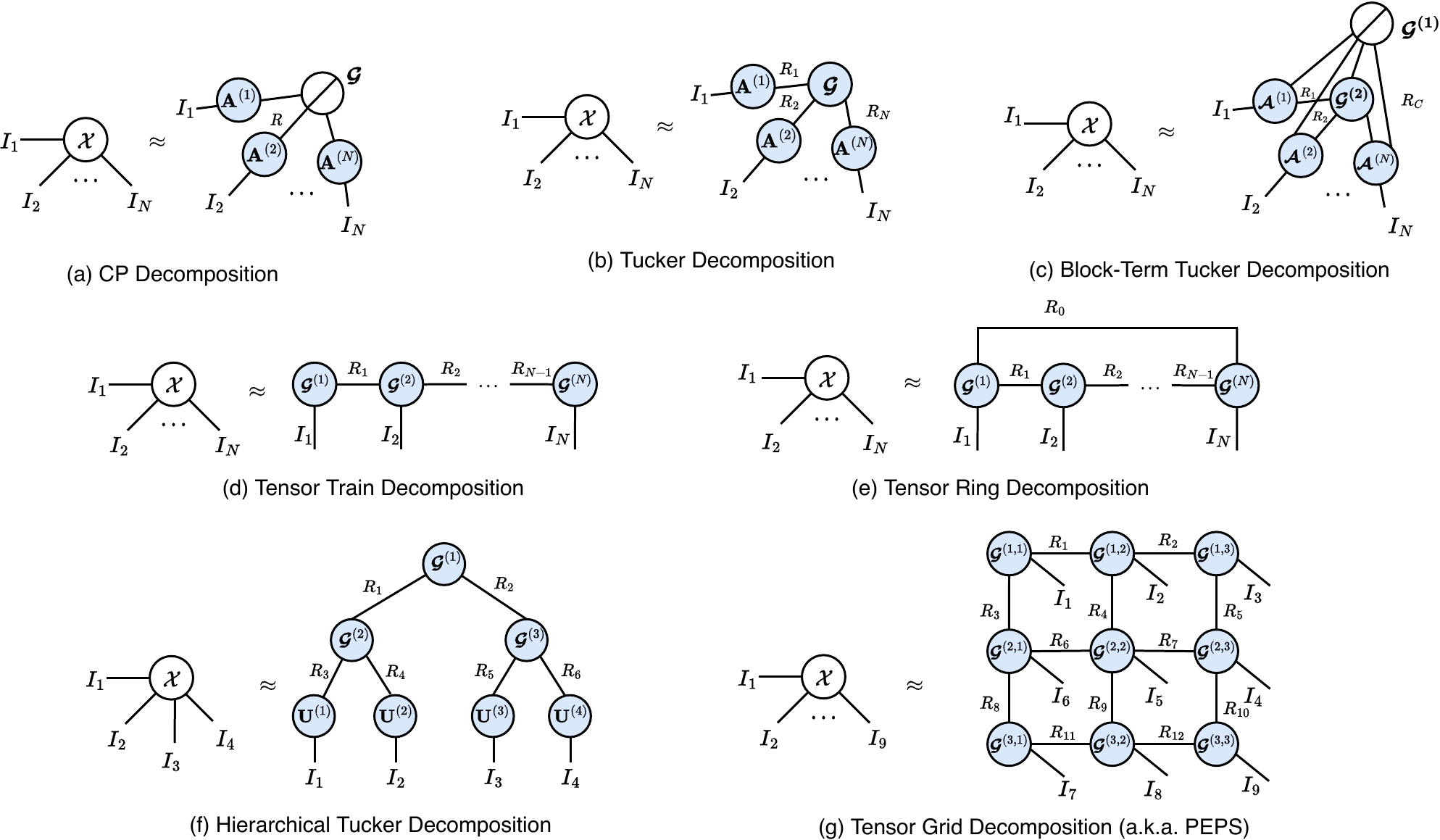}
    \label{fig:cpd}

\caption{TN diagrams of some popular TN decompositions.
(a) The CP format decomposes a tensor ${\tbtensor X}$ into a sum of several rank-1 tensors $\bm a^{(1)}_{:,r}\circ \bm a^{(2)}_{:,r}\circ\cdots \circ \bm a^{(N)}_{:,r}$.
(b) Tucker decomposition decomposes a tensor $\tbtensor{X}$ into a core tensor $\tbtensor{G}$ multiplied by a matrix $\bm{A}^{(n)}$ along the $n$th mode.
(c) Block term decomposition decomposes a tensor ${\tbtensor X}$ into a sum of several Tucker decompositions (on the right) with low Tucker ranks.
(d) TT decomposition decomposes a tensor ${\tbtensor X}$ into a linear multiplication of a set of 3rd-order core tensors $\tbtensor{G}^{(2)}\cdots\tbtensor{G}^{(N-1)}$ and two matrices $\tbtensor{G}^{(1)},\quad\tbtensor{G}^{(N)}$.
(e) TR decomposition decomposes a tensor ${\tbtensor X}$ into a set of 3rd-order core tensors and contracts them into a ring structure.
(f) HT decomposition represents a tensor ${\tbtensor X}$ as a tree-like diagram. For more basic knowledge about TNs, we refer to \cite{biamonte2017tensor} and \cite{cichocki2016tensor}.
(g) Tensor Grid Decomposition (a.k.a. PEPS) represents a high-dimensional tensor as a two-dimensional grid of interconnected lower-rank tensors, where each node connects to its neighbors to efficiently capture spatial correlations in systems with local interactions.
}
\label{fig:tensor-formats}
\end{figure*}

\subsubsection{Hyperedge}
Fig.~\ref{fig:pre:basic_symbol} illustrates the hyperedge that was also introduced by Hayashi \textit{et al.}~\cite{DBLP:conf/nips/HayashiYSM19}. An example of a hyperedge with a size of $R$ can be formulated as
\begin{equation}
\label{eq:hyper}
\tbtensor{Y}_{ijk}
= \sum_{r=1}^{R} \tbmatrix{A}_{ir}\tbmatrix{B}_{jr} \tbmatrix{C}_{kr},
\end{equation}
where $\tbmatrix{A}\in\mathbb{R}^{I\times R}, \tbmatrix{B}\in\mathbb{R}^{J\times R}$ and $\tbmatrix{C} \in\mathbb{R}^{K\times R}$ are three matrices. $\tbtensor{Y} \in\mathbb{R}^{I \times J \times K}$ denotes the results of applying a hyperedge on $\tbmatrix{A}$, $\tbmatrix{B}$, and $\tbmatrix{C}$.
A hyperedge represents contraction with a copy tensor, i.e., an identity super-diagonal tensor whose entry is one when all incident indices are equal and zero otherwise. It therefore enforces equality among the connected bond indices and sums over their shared value, as expressed in Eq.~\eqref{eq:hyper}; it is not an addition operator in isolation. Hayashi \textit{et al.}~\cite{DBLP:conf/nips/HayashiYSM19} use dummy tensors together with such hyperedges to represent convolutional operations and show that TCNN architectures can be expressed systematically as tensor-network diagrams.

\subsubsection{Super-diagonal Tensor}

A super-diagonal tensor has nonzero entries only when all of its indices are equal. For an $N$th-order tensor $\tbtensor{G} \in \mathbb{R}^{\overbrace{I\times I\times \cdots\times I}^{N}}$, this means that $\tbtensor{G}_{i_1,i_2,\ldots,i_N}=0$ unless $i_1=i_2=\cdots=i_N$. As shown in Fig.~\ref{fig:pre:basic_symbol}, a super-diagonal tensor is designated by a node with a skew line in TN diagrams. The identity or copy tensor is the special case whose super-diagonal entries are all one; contracting with it enforces equality of the incident indices, as in the hyperedge construction above.

\subsubsection{Tensor Unfolding}

Tensor unfolding is an operation that virtually flattens a tensor into a high-dimensional but low-order tensor. Matricization is a special case of tensor unfolding. To be more specific, given an $N$th-order tensor $\tbtensor{A}$ $\in\mathbb R^{I_1\times I_2 \cdots\times I_N}$, its mode-$n$ unfolding process yields a matrix $\bm A_{(n)}\in \mathbb R^{ I_n\times I_1  I_2\cdots I_{n-1} I_{n+1}\ldots I_N}$. Such an operation can also be regarded as performing tensor contraction with a specifically designed tensor. A fourth-order tensor unfolding diagram is illustrated in Fig.~\ref{fig:pre:basic_symbol}.

\subsection{Tensor Decomposition Formats}

The commonly used terminology ``tensor decomposition'' (TD) is equivalent to ``tensor network'' to some extent. While TD was employed primarily in signal processing fields~\cite{kolda2009tensor, sidiropoulos2017tensor}, TNs were originally utilized largely in the physics and quantum circuit fields~\cite{penrose1971applications, biamonte2017tensor}. Traditional TD models, such as CP~\cite{Hars1970,carroll1970analysis,kiers2000towards} and Tucker decomposition~\cite{tucker1966some,tucker1963implications}, can be viewed as basic kinds of TNs. In the realm of signal processing, several powerful TNs architectures for quantum analysis have also been introduced. For instance, the MPS decomposition~\cite{schollwock2013matrix} was defined as a TT decomposition~\cite{oseledets2011tensor} and had tremendous success in several applications~\cite{cichocki2017tensor}. After years of collaboration and progress across different research fields, there is no significant distinction between these two terminologies. Therefore, TD and TNs are treated in a unified way in this paper. We briefly introduce some basic TDs by employing TN diagrams.
\rthreehl{Hong \textit{et al.}~\cite{hong2020generalizedcp}, Ghadiri \textit{et al.}~\cite{ghadiri2023core_shapes}, and Li \textit{et al.}~\cite{li2022tnls} show that the design space of a decomposition is richer than its format label: the loss function, core shape, and mode permutation can all substantially affect practical performance.}

\subsubsection{CANDECOMP/PARAFAC}

The CP decomposition~\cite{Hars1970,carroll1970analysis,kiers2000towards} factorizes a higher-order tensor into a sum of several rank-1 tensor components. For instance, given an $N$th-order tensor $\tbtensor{X}\in\mathbb R^{I_1\times I_2\cdots I_N}$, each of its elements in the CP format can be formulated as
\begin{align}\label{eq:pre:CP}
	\tbtensor{X}_{i_1,i_2,\ldots,i_N} \approx \sum_{r=1}^R \tbtensor{G}_{r} \prod^N_{n=1}\tbmatrix{A}^{(n)}_{i_n,r},
\end{align}
where $R$ denotes the CP rank (defined as the smallest possible number of rank-1 tensors \cite{kolda2009tensor}), $\tbtensor{G}$ denotes the $N$-th-order super-diagonal tensor, and $\tbmatrix{A}^{(n)}\in\mathbb R^{I_n\times R}$ denotes a series of factor matrices.
The TN diagram for CP is illustrated in Fig.~\ref{fig:tensor-formats}~(a). 
\rwthree{\textit{Notation:} In this and subsequent TN diagrams, the upper and lower numbers adjacent to a tensor core denote the dimensions of the corresponding modes, with upper indices representing input dimensions and lower indices representing output or bond dimensions. This convention follows the standard TN literature~\cite{biamonte2017tensor}.}
% We also provide a detailed visualization of CP in Fig.~\ref{fig:cpon5order} as an illustrative case of a TN.

When calculating a CP format, the first issue that arises is how to determine the number of rank-1 tensor components, i.e., the CP rank $R$. Actually, this is an NP-hard problem~\cite{hillar2013most}. Hence, in practice, a numerical value is usually assumed in advance (i.e., as a hyperparameter), to fit various CP-based models~\cite{kolda2009tensor}.
After that, the diagonal core tensor $\tbtensor{G}$ and the factor matrices $\tbmatrix{A}^{(n)}$ can be directly solved by employing algorithmic iteration, which usually involves the alternating least-squares (ALS) method that was originally proposed in \cite{Hars1970,carroll1970analysis}.
\rthreehl{Hong \textit{et al.}~\cite{hong2020generalizedcp} extend CP factorization beyond the standard least-squares setting to broader loss functions and data types, which is especially relevant for non-Gaussian observations and task-specific objectives.}

\subsubsection{Tucker Decomposition}

Tucker decomposition~\cite{tucker1966some,tucker1963implications} factorizes a higher-order tensor into a core tensor multiplied by a corresponding factor matrix along each mode. To be more specific, given an $N$th-order tensor $\tbtensor{X} \in\mathbb R^{I_1\times I_2\cdots I_N}$, the Tucker decomposition can be formulated in an elementwise manner as
\begin{align}\label{eq:pre:Tucker}
	\tbtensor{X}_{i_1,i_2,\ldots,i_N} 
\approx \sum_{r_1,\ldots,r_N=1}^{R_1,\ldots,R_N} \tbtensor{G}_{r_1,r_2,\ldots,r_N} \prod^N_{n=1}\tbmatrix{A}^{(n)}_{i_n,r_n},
\end{align}	
where $\{R_1,R_2,\ldots,R_N\}$ denotes a series of Tucker ranks, $\tbtensor{G} \in \mathbb R^{R_1\times R_2\ldots R_N}$ denotes the dense core tensor and $\tbmatrix{A}^{(n)}\in\mathbb R^{I_n\times R_n}$ denotes a factor matrix. The TN diagram for Tucker decomposition is illustrated in Fig.~\ref{fig:tensor-formats}~(b). Please note that compared with the CP rank, $R_1, R_2, \ldots, R_N$ can take different numerical values.

Tucker decomposition is commonly used and can be degraded to CP by setting the core tensor $\tbtensor{G}$ as a super diagonal tensor. In addition, the original Tucker decomposition lacks constraints on its factors, leading to the nonuniqueness of its decomposition results, which is typically undesirable for practical applications due to the lack of explainability. Consequently, orthogonality constraints are always imposed on the component matrices, yielding the well-known higher-order singular value decomposition (HOSVD) algorithm~\cite{de2000multilinear}.

\subsubsection{Block-Term Decompositions}

\aehl{Block-term formats interpolate between CP and Tucker by representing a tensor as a sum of low-multilinear-rank Tucker blocks~\cite{de2008decompositions1,de2000multilinear}. In the BTT construction shown in Fig.~\ref{fig:tensor-formats}~(c), $R_C$ controls the number of blocks and $R_T$ controls their shared Tucker rank. Rank-one blocks recover a CP-like representation, whereas a single block recovers Tucker. General BTD allows each block to have its own multilinear-rank profile. Supplementary Note~S1 gives the detailed BTT and BTD formulations.}

\subsubsection{Tensor Train and Matrix Product Operator}

\aehl{The TT decomposition~\cite{oseledets2011tensor,cichocki2017tensor}, also known as a Matrix Product State (MPS) in quantum physics~\cite{schollwock2011density,schollwock2013matrix}, factorizes an $N$th-order tensor into a chain of third-order cores. Its representative elementwise form is}
\begin{align}\label{eq:TN:TTD}
	&  \tbtensor{X}_{i_1,i_2,\ldots,i_N} \approx \sum_{r_{1},r_{2},\ldots,r_{N-1}=1}^{R_1,R_2,\ldots,R_{N-1}} \notag \\
	& ~~ \tbtensor{G}^{(1)}_{1,i_1,r_1}  \tbtensor{G}^{(2)}_{r_1,i_2,r_2} \tbtensor{G}^{(3)}_{r_2,i_3,r_3}\cdots  \tbtensor{G}^{(N)}_{r_{N-1},i_N,1},
\end{align}	
\aehl{Here, $\tbtensor{G}^{(n)}\in\mathbb R^{R_{n-1}\times I_n\times R_n}$ and $R_0=R_N=1$. The chain in Fig.~\ref{fig:tensor-formats}~(d) supports sequential SVD construction and contractions whose cost grows linearly with tensor order for fixed mode sizes and ranks~\cite{huckle2013computations}.}
\rthreehl{Bharadwaj \textit{et al.}~\cite{bharadwaj2024leverage_tt} use leverage-score-based sampling to reduce the cost of constructing TT approximations while preserving high-quality subspace information.}
\aehl{For tensorized linear operators, MPO pairs local input and output modes in fourth-order cores $\tbtensor{G}^{(n)}\in\mathbb R^{R_{n-1}\times I_n\times J_n\times R_n}$~\cite{pirvu2010matrix,DBLP:conf/icml/YangKT17,ju2019tensorizing,benyouformer}. Its full expression is moved to Supplementary Note~S2.}

\subsubsection{Tensor Ring Decomposition}

\aehl{TR closes the two boundary bonds of a TT to form the cycle shown in Fig.~\ref{fig:tensor-formats}~(e)~\cite{zhao2016tensor,sedighin2021image,qiu2024tensor}. Its cores retain the shape $\tbtensor{G}^{(n)}\in\mathbb R^{R_{n-1}\times I_n\times R_n}$, but the cyclic condition $R_0=R_N$ removes the distinguished rank-one endpoints. This symmetry provides flexible mode ordering while making rank selection, gauge handling, and optimization more involved than in an open chain. Supplementary Note~S3 records the cyclic contraction formula.}
\rthreehl{This additional flexibility makes rank and latent-space design especially important in TR models. Yuan \textit{et al.}~\cite{yuan2019trlatent} minimize rank in the latent space to obtain compact and stable TR representations for completion and learning problems.}

\subsubsection{Hierarchical Tucker Decomposition}
\aehl{HT organizes tensor modes with a dimension tree rather than a chain~\cite{kressner2012htucker}. Leaf frames represent individual modes and internal transfer tensors combine child subsets recursively, producing the structure in Fig.~\ref{fig:tensor-formats}~(f). For a balanced tree with uniform mode dimension $d$ and rank bounded by $R$, its storage is commonly written as $O(NdR+NR^3)$, compared with $O(NdR^2)$ for TT. HT can therefore exploit meaningful hierarchical mode groupings, but its performance and cost depend on the selected dimension tree. The recursive formulation is provided in Supplementary Note~S4.}
\rthreehl{Falco \textit{et al.}~\cite{falco2015geometric} study tensors of fixed or bounded tree-based rank as manifold-structured subsets of ambient tensor spaces. This geometric view connects Tucker, HT, and TT representations and clarifies why rank-constrained optimization has exploitable local structure. Because the cited work remains a preprint, we use it as mathematical background rather than as empirical evidence.}
\subsubsection{PEPS Decomposition}

\aehl{PEPS arranges local cores on a two-dimensional grid~\cite{biamonte2017tensor,schuch2007computational,sedighin2020matrix,wang2020concatenated}. Each core contains one physical mode and up to four bond modes connecting its left, right, upper, and lower neighbors, as illustrated in Fig.~\ref{fig:tensor-formats}~(g). The loopy lattice can encode multidimensional dependencies beyond a chain or tree, but exact contraction is \#P-hard in general and practical use relies on controlled approximations. Supplementary Note~S5 gives the grid-indexed reconstruction formula.}
\rthreehl{Most current TNNs employ low-treewidth structures, including TT/MPS chains, TR cycles, and HT trees, because their contractions remain tractable. PEPS provides additional correlation pathways through a loopy lattice, but topology and contraction algorithm must be designed together. Alkabetz and Arad~\cite{alkabetz2021tensor} analyze belief propagation as a tensor-network contraction procedure, while Tindall \textit{et al.}~\cite{tindall2024efficient} demonstrate message-passing-based simulation on loopy lattice structures. These results motivate approximate-contraction methods whose numerical error and task-level effect are reported explicitly~\cite{huang2025innovative}.}

\aehl{To reduce preliminary material in the main article, the detailed formulations of BTT/BTD, MPO, TR, HT, and PEPS are moved to the supplementary material. The tensor-network diagrams in Fig.~\ref{fig:tensor-formats} and the concise format descriptions above retain the structural information needed for the subsequent TNN sections without introducing another comparison table. The supplementary material records the omitted formulas, while the remainder of the survey uses the common topological, rank, and contraction properties summarized in the text.}

\vspace{3mm}
% \newpage
\section{SUSTAINABLE AI THROUGH TNN in Data Aspect: Effective Data Representation}
\label{Sec:IFTNN}

In real-world data analysis, information often comes from multiple sources, such as vision, sound, and text in video data~\cite{lahat2015multimodal,antol2015vqa}. A prime example is the Visual Question Answering (VQA) task, where the key challenge lies in effectively modeling interactions between textual and visual information. Processing such diverse data sources uniformly is impractical, necessitating specialized architectures with multiple input channels to handle multimodal sources - an approach known as information fusion. While traditional feature-level and decision-level fusion pipelines were popular in early stages~\cite{DBLP:conf/icmi/MorencyMD11}, these linear approaches failed to effectively model intramodality dynamics. Tensor Neural Networks~(TNNs) have emerged as a solution, leveraging their natural multilinear properties to model intramodality dynamics and process higher-order data. TNNs provide effective frameworks for tensor operations, making them naturally suited for expressing and generalizing information fusion modules commonly found in deep learning, such as attention mechanisms and vector concatenation~\cite{zhou2016linked}. As a result, numerous studies have adopted TNNs to capture higher-order interactions among data or parameters.
In the following sections, we will explore various TNN-based approaches for data representation and processing. First, we examine advanced data compression techniques that leverage tensor network architectures to achieve significant parameter reduction while preserving critical information structures. We then investigate novel tensor fusion layers (Section~\ref{Sec:TFL}) designed to facilitate deep feature interactions and transformations across modalities, followed by sophisticated multimodal data pooling mechanisms (Section~\ref{Sec:VQA}) that effectively integrate information across different data types.

\subsection{Multi-source Data fusion}
\label{Sec:TFL}

\rwtwo{The connection between Hadamard products and CP decomposition was established in the polynomial networks framework~\cite{chrysos2020p, chrysos2021deep}. In multimodal fusion, the Hadamard product enables efficient low-rank bilinear pooling~\cite{kim2016hadamard}, avoiding the exponential parameter growth of outer-product-based methods. For a broader discussion of multiplicative interactions in deep learning, see Jayakumar \textit{et al.}~\cite{jayakumar2020multiplicative}.}

\begin{figure}[t]
\centering
\includegraphics[width=0.45\textwidth]{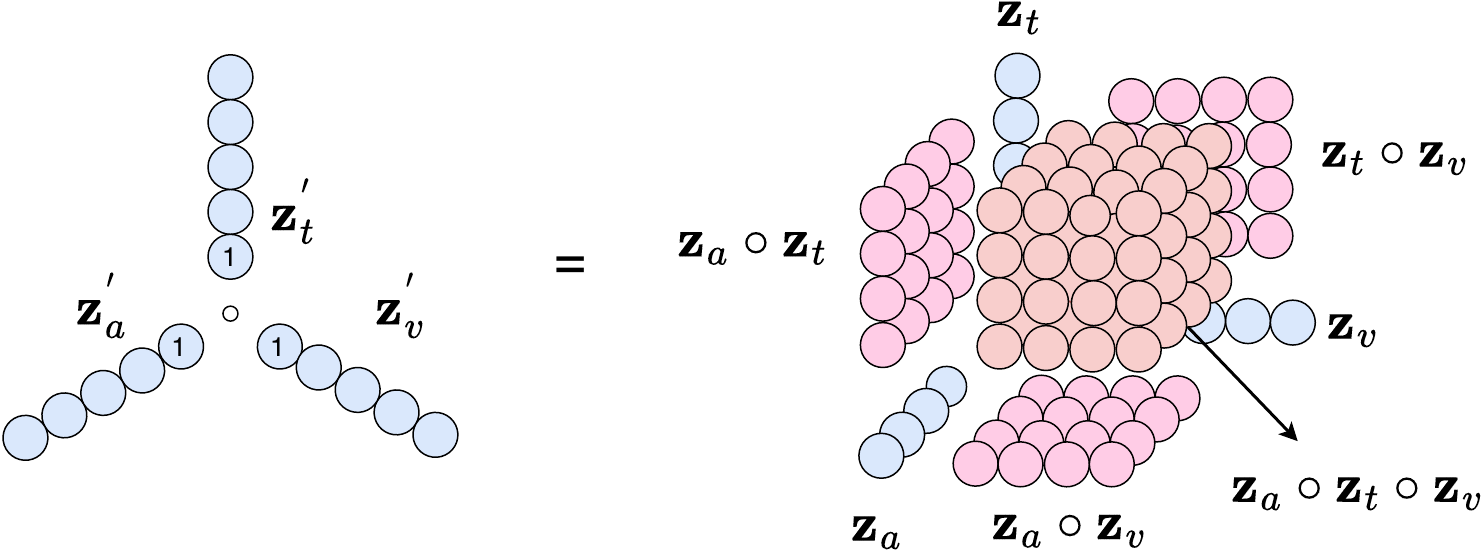}
\caption{Illustration of the tensor fusion process in Eq.~\eqref{eq:tensorfusion}. Different from a TN diagram, each circle corresponds to a value.}
\label{fig:Fusion1}
\end{figure}

Multimodal sentiment analysis is a task containing three communicative modalities, i.e., the textual modality, visual modality, and acoustic modality~\cite{zadeh2017tensor}. Addressing multimodal sentiment analysis, Zadeh \textit{et al.}~\cite{zadeh2017tensor} proposed novel TNNs with deep information fusion layers named tensor fusion layers (TFLs), which can easily learn intramodality dynamics and intermodality dynamics and are able to aggregate multimodal interactions, thereby efficiently fusing the three communicative modalities.
Specifically, a TFL first takes embedded feature vectors $\tbvector{z_t}$, $\tbvector{z_v}$ and $\tbvector{z_a}$ derived by embedding networks rather than the original three data types.
Then, the TFL concatenates a scalar $1$ with each embedded feature vector as
\begin{equation}
\tbvector{z}^{'}_{t}=\left[\begin{array}{c}
\tbvector{z}_{t} \\
1
\end{array}\right] 
\tbvector{z}^{'}_{v}=\left[\begin{array}{c}
\tbvector{z}_{v} \\
1
\end{array}\right] \tbvector{z}^{'}_{a}=\left[\begin{array}{c}
\tbvector{z}_{a} \\
1
\end{array}\right].
\end{equation}
Then, as shown in Fig.~\ref{fig:Fusion1}, the TFL obtains a feature tensor $\tbtensor{Z}$ by calculating the outer product among the three concatenated vectors
\begin{equation}
\tbtensor{Z}=\tbvector{z}^{'}_{t} \circ \tbvector{z}^{'}_{v} \circ\tbvector{z}^{'}_{a} = \left[\begin{array}{c}
\tbvector{z}_{t} \\
1
\end{array}\right] \circ \left[\begin{array}{c}
\tbvector{z}_{v} \\
1
\end{array}\right] \circ\left[\begin{array}{c}
\tbvector{z}_{a} \\
1
\end{array}\right].
\label{eq:tensorfusion}
\end{equation}
Finally, the TFL processes the feature tensor $\tbtensor{Z}$ to obtain a prediction $\tbvector{y}$ via a two-layer fully connected NN.
Compared to direct concatenation-based fusion, which only considers unimodal interactions~\cite{zadeh2017tensor}, the TFL benefits from capturing both unimodal interactions and multimodal interactions.

Despite its success, the TFL suffers from an exponential increase in its computational complexity and number of parameters when the number of modalities increases. For example, in a multimodal sentiment analysis case~\cite{zadeh2017tensor}, the feature tensor $\tbtensor{Z} \in \mathbb{R}^{129 \times 33 \times 33}$ and the hidden vector $\tbvector{h} \in \mathbb{R}^{128}$ can result in $17,981,568$ parameters to be optimized.
To address these excessive parameters, low-rank multimodal fusion (LMF)~\cite{liu2018efficient}
adopts a special BTT layer to overcome the massive computational cost and overfitting risks of the TFL. For a general situation with $n$ modalities, the feature tensor $\tbtensor{Z}=\circ_{m=1}^{M} z^{'}_{m}$ can be processed.
The hidden vector $\tbvector{h}$ can be computed as follows
\begin{align}
    \tbvector{h} = ReLU\left(\bm{he}( \tbmatrix{W}_{1} z^{'}_{1},\tbmatrix{W}_{2} z^{'}_{2},\ldots,\tbmatrix{W}_{M} z^{'}_{M}, I)+\tbvector{b}\right), \notag
\end{align}
where $\tbmatrix{W}_{i} \in \mathbb{R}^{d_{i} \times d_{h}}$ is the weight matrix and $I\in \mathbb{R}^{d_{h} \times d_{h}}$ is an identity matrix. The LMF reduces the computational complexity of the TFL from $O\left(\prod_{m=1}^M d_m\right)$ to $O\left(d_h\times \sum_{m=1}^M d_m\right)$.

\begin{figure*}[t]
\centering
\includegraphics[width=0.85\textwidth]{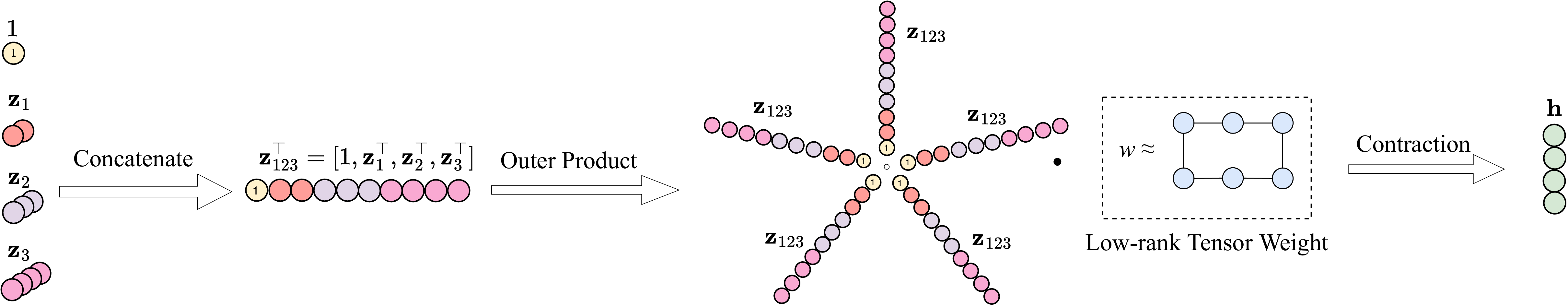}
\caption{Illustration of polynomial tensor pooling~(PTP)~\cite{hou2019deep}. PTP first concatenates all feature vectors $\tbvector{z}_1,\tbvector{z}_2,\tbvector{z}_3$ into a longer feature vector $\tbvector{z}_{123}^{\top}=\left[1, \tbvector{z}_{1}^{\top}, \tbvector{z}_{2}^{\top}, \tbvector{z}_{3}^{\top}\right]$, it then derives a polynomial feature tensor by repeatedly performing outer product operations on the feature vector $\tbvector{z}_{123}$, and finally adopts a tensorial layer (e.g., a TR layer) to merge the polynomial feature tensor into a vector $\tbvector{h}$.}
\label{fig:ptpFusion2}
\end{figure*}

\begin{figure}[t]
\centering
\includegraphics[width=0.4\textwidth]{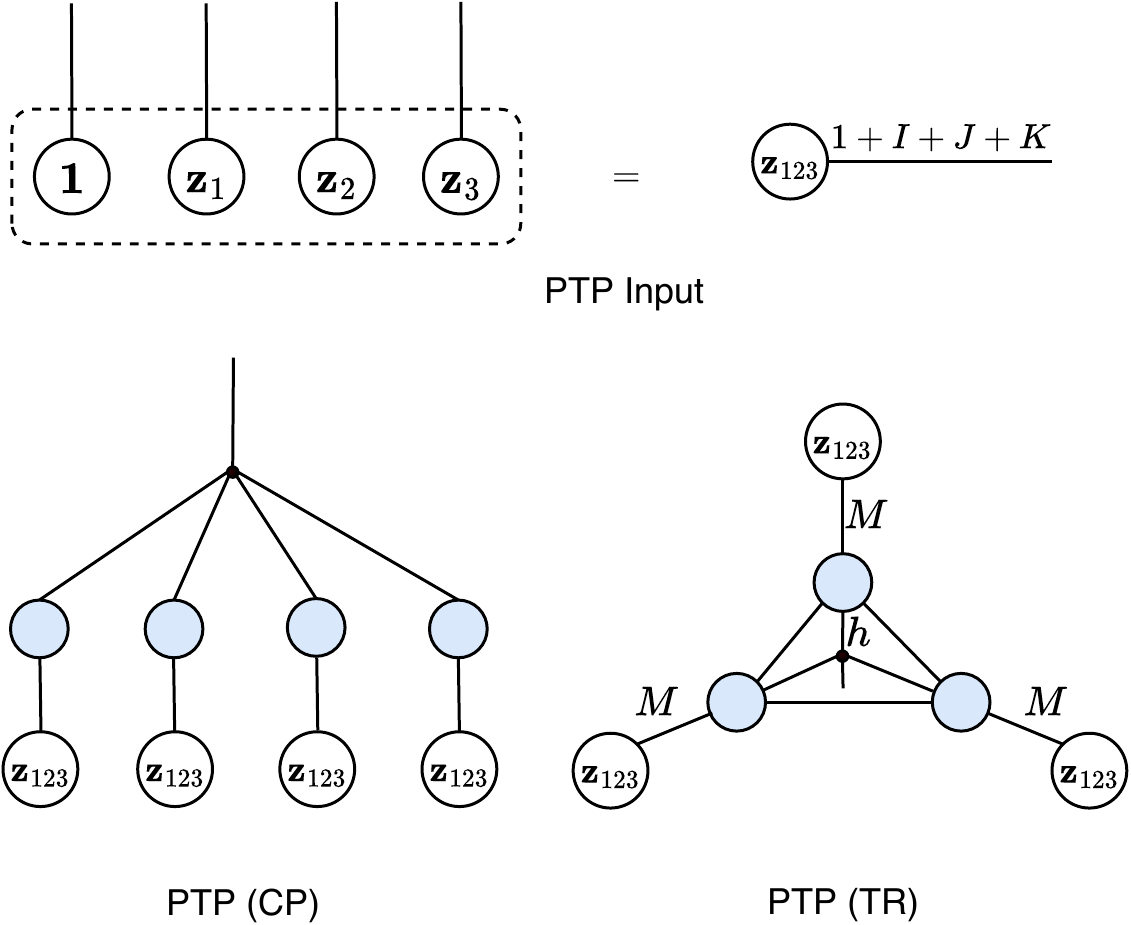}
\caption{TN diagrams of PTP. The CP and TR structures can be adopted in such a strategy.}
\label{fig:ptptensor}
\end{figure}

Although LMF and the TFL achieve better fusion results than other methods, they restrict the order of interactions, causing higher-order interactions to lack information. A PTP~\cite{hou2019deep} block has been proposed to tackle this problem. The whole procedure and TN diagram of PTP are shown in Fig.~\ref{fig:ptpFusion2} and Fig.~\ref{fig:ptptensor}, respectively.

The PTP first merges all feature vectors $\left\{\tbvector{z}_{m}\right\}_{m=1}^{M}$ into a long feature vector
\begin{equation}
    \tbvector{z}_{12 \cdots M}^{\top}=\left[1, \tbvector{z}_{1}^{\top}, \tbvector{z}_{2}^{\top}, \ldots, \tbvector{z}_{M}^{\top}\right].
\end{equation}
The polynomial feature tensor of degree $P$ is represented as
\begin{equation}
    \tbtensor{Z}^{P}=\tbvector{z}_{12 \ldots M} \circ\tbvector{z}_{12 \ldots M} \circ \cdots \circ \tbvector{z}_{12 \cdots M}.
    \label{eq:outerproduct}
\end{equation}
The PTP~\cite{hou2019deep} then adopts a tensorial layer (e.g., a CP layer) to process the polynomial feature tensor $\tbtensor{Z}^{P}$. The CP layer is represented as
\begin{align}
    \tbvector{h} = \bm{he}( \tbmatrix{W}_{1} \tbvector{z}_{12 \ldots M},\ldots \tbmatrix{W}_{P}\tbvector{z}_{12 \ldots M}, \tbmatrix{\Lambda})
\end{align}
where $\tbmatrix{W}_{i} \in \mathbb{R}^{d_{i} \times d_{h}}$ is the weight matrix and $\tbmatrix{\Lambda} \in \mathbb{R}^{d_{h} \times d_{h}}$ is a learnable diagonal matrix.
The structure of PTP is also equivalent to that of a deep polynomial NN~\cite{chrysos2021deep}, whereby
PTP models all nonlinear high-order interactions. For multimodal time series data, one approach uses a ``window'' to characterize local correlations and stack the PTP blocks into multiple layers. Such a model is called a hierarchical polynomial fusion network (HPFN)~\cite{hou2019deep}. The HPFN can recursively process local temporal-modality patterns to achieve a better information fusion effect.

The structure of a single-layer PTP block is similar to that of a shallow convolutional arithmetic circuit (ConvAC) network~\cite{zhang2018quantum} (see Section 3.5 and 4.5). The key distinction lies in their application domains - while ConvAC processes quantum location features, PTP handles temporal-modality patterns and polynomial concatenated multimodal features. The HPFN's architecture closely resembles a deeper ConvAC network, suggesting its enhanced expressive power through this structural connection. The recursive relationships in deep polynomial NNs have been extensively studied and implemented for efficient hierarchical computation~\cite{hou2019deep}, with similar findings reported by Chrysos et al.~\cite{chrysos2021deep}.

Varshneya \textit{et al.}~\cite{varshneya2024interpretable} use tensor fusion to derive relevance scores for individual modalities and their interactions in sentiment analysis and sarcasm detection. Nie \textit{et al.}~\cite{nie2024interpretable} combine mutual-information-based filtering with Tucker fusion to expose feature interactions in multimodal sentiment analysis, while Miao \textit{et al.}~\cite{miao2024low} integrate low-rank tensor fusion with self-supervised multi-task learning to capture inter-modal correlations. Wang \textit{et al.}~\cite{wang2024bearing} further combine mutual attention with feature-tensor construction for multi-sensor fault diagnosis.

\subsection{Multimodal Data Pooling}
\label{Sec:VQA}

Another group of information fusion methods originated from VQA tasks~\cite{antol2015vqa}.
In VQA tasks, the most important aspect is to parameterize bilinear the interactions between visual and textual representations. To address this aspect, some tensor fusion methods have been discovered in this area. Multimodal compact bilinear pooling (MCB)~\cite{fukui2016multimodal} is a well-known fusion method for VQA tasks and can be regarded as a special Tucker decomposition-based NN,
which tries to optimize the simple bilinear fusion operation
\begin{equation}
    \tbvector{z}=\tbmatrix{W}[\tbvector{v} \circ \tbvector{q}],
\end{equation}
where $\tbvector{v}$ and $\tbvector{q}$ are input vectors with different modalities and $\tbmatrix{W}$ is a learnable weight matrix. Moreover, MCB optimizes the computational cost of the outer product operation based on the property of the count sketch projection function.

Multimodal low-rank bilinear pooling (MLB)\cite{kim2016hadamard} adopts a CP layer in a data fusion step that can be formulated as 
\begin{equation}
\tbvector{z} = \tbvector{1}^{T} \left(\tbmatrix{W}_{v}\tbvector{v} \circ \tbmatrix{W}_{q}\tbvector{q}\right),
\end{equation}
where $\tbmatrix{W}_{q}$ and $\tbmatrix{W}_{v}$ are the prepossessing weight matrices for the inputs $\tbvector{q}$ and $\tbvector{v}$, respectively and $\tbvector{1}$ is a vector in which all values are 1.
The structure of the MLB method is a special case of LMF (see Sec.~\ref{Sec:TFL}). MLB fusion methods can also be regarded as simple product pooling when the number of modalities is equal to two.

The MUTAN~\cite{ben2017mutan} is a generalization of MCB and MLB, which adopts a Tucker layer to learn the bilinear interactions between visual and textual features as
\begin{align}
    \tbvector{z}&=\left(\left(\tbtensor{T}_{c} \times^{1}_{1}\left(\tbvector{q}^{\top} \tbmatrix{W}_{q}\right)\right)  \times^{1}_{2}\left(\tbvector{v}^{\top} \tbmatrix{W}_{v}\right)\right) \times^{1}_{3} \tbmatrix{W}_{o}, \notag\\
    \tbvector{z}&=\left(\tbtensor{T}_{c} \times^{1}_{1} \tilde{\tbvector{q}}\right) \times^{1}_{2} \tilde{\tbvector{v}},
\end{align}
where $\tilde{\tbvector{q}}=\tanh \left(\tbvector{q}^{\top} \tbmatrix{W}_{q}\right)$ and $\tilde{\tbvector{v}}=\tanh \left(\tbvector{v}^{\top} \tbmatrix{W}_{v}\right)$, $\tbtensor{T}_{c}$ is the fusion weight tensor, and $\tbmatrix{W}_{o}$ is the output processing weight matrix. Moreover, MUTAN~\cite{ben2017mutan} adopts a low rank for the fusion weight tensor $\tbtensor{T}_{c}$, as follows:
\begin{equation}
\tbtensor{T}_{c}[:,:, k]=\sum_{r=1}^{R} \tbvector{m}_{r}^{k} \circ \tbvector{n}_{r}^{k \top},
\end{equation}
where $\tbvector{m}_{r}^{k} $ and $\tbvector{n}_{r}^{k \top}$ are weight vectors and $\tbdim{R}$ is the number of ranks. In this way, MUTAN can represent comprehensive bilinear interactions while maintaining a reasonable model size by factorizing the interaction tensors into interpretable elements.

Furthermore, compact trilinear interaction (CTI)\cite{do2019compact} was proposed, which uses an attention-like structure. Instead of presenting the given data as a single vector, this method represents every modality as a matrix $\mathbf{A}\in\mathbb{R}^{n_1\times d_a}$, where $d_a$ corresponds to the feature dimension and $n_1$ denotes the number of states. The CTI simultaneously learns high-level trilinear joint representations in VQA tasks and overcomes both the computational complexity and memory issues in trilinear interaction learning\cite{do2019compact}.
Hua \textit{et al.}~\cite{hua2022high} use high-order tensor pooling to capture nonlinear node interactions in graph neural networks. For action recognition, Wang \textit{et al.}~\cite{wang2024high} combine mutual attention with eigenvalue power normalization to model high-order statistics while controlling burstiness. Chen \textit{et al.}~\cite{chen2025tdmfs} apply tensor networks and Tucker-based bilinear pooling to interactions between gene expression and copy-number variation for pan-cancer survival prediction.

\subsection{Multi-way Data Compression}
\label{Sec:Data_Compression}
TNNs present a powerful framework for addressing the unique challenges in multi-dimensional data compression. Unlike traditional compression methods that treat data as vectors or matrices, or conventional tensor methods that rely on fixed decomposition structures, TNNs leverage learnable neural architectures to adaptively preserve and exploit the natural multi-dimensional relationships in the data, leading to more efficient and accurate representations with theoretical guarantees.

The BNTD~\cite{he2018knowledge} first introduces TNNs and advances multi-way data compression through a principled probabilistic framework, effectively modelling complex entity-relation interactions and incorporating prior information via neural tensor architectures. Wang \textit{et al.}~\cite{wang2023federated} propose federated latent embedding sharing and tensor factorization (FLEST), which combines tensor factorization with embedding-matrix decomposition for communication-efficient federated knowledge-graph completion without directly sharing local graph data. The TTHRESH method~\cite{ballester2019tthresh} leverages HOSVD decomposition combined with bit-plane, run-length and arithmetic coding to efficiently compress high-dimensional gridded data for visualization, achieving smooth quality degradation and enabling low-cost compressed-domain manipulations while providing competitive compression ratios at low-to-medium bit rates.
Fan \textit{et al.}~\cite{fan2021multi} proposed a multi-mode deep matrix and tensor factorization approach (M2DMTF) that employs TKD with factor matrices generated using multilayer perceptrons, effectively handling complex tensor data with missing values and noise. Lee and Shin~\cite{lee2021robust} developed a robust factorization method specifically designed for real-world tensor streams containing patterns, missing values, and outliers. Lamba \textit{et al.}~\cite{lamba2016incorporating} introduced a method for incorporating side information into tensor factorization, improving the quality of compression and representation learning.

NeuKron~\cite{kwon2023neukron} extends tensor neural networks by introducing auto-regressive neural networks to generalize Kronecker products, enabling constant-size lossy compression of sparse reorderable matrices and tensors. TensorCodec~\cite{kwon2024compact,kwon2023tensorcodec} extends tensor neural networks for efficient data compression by introducing neural tensor-train decomposition, tensor folding, and mode-index reordering techniques, enabling accurate compression without strong data assumptions.
Light-IT and Light-IT++~\cite{kwon2024compact} extend tensor neural networks for efficient data compression by introducing vocabulary-based compression and core tensor operations, enabling compact and accurate representation of irregular tensors.
\rthreehl{Loeschcke \textit{et al.}~\cite{loeschcke24a} construct coarse-to-fine tensor trains that can be progressively refined for image fitting, denoising, inpainting, and novel-view synthesis. Their results show that TT parameterizations support flexible continuous visual modeling and storage reduction.}

\subsection{Multi-task Data Training}
\label{Sec:Multi-task}
For multitask learning applications, WISDOM~\cite{xu2016wisdom} pioneered an incremental learning algorithm that performs supervised tensor decomposition on spatio-temporal data encoded as third-order tensors, simultaneously training spatial and temporal prediction models from extracted latent factors while incorporating domain knowledge, demonstrating superior performance over baseline algorithms in global-scale climate data prediction across multiple locations. Yang \textit{et al.}~\cite{yang2017deep} then proposed the Tensor Train multitask (TTMT) and Tucker multitask (TMT) models using TT and Tucker formats, respectively, to alleviate the negative transfer problem in a hard sharing architecture and reduce the parameter volume in a soft structure. The M2TD method~\cite{li2018m2td} stitches patterns from partitioned parameter subspaces of large simulation ensembles to efficiently discover underlying dynamics and interrelationships while maximizing accuracy under limited simulation budgets. 

Zhang \textit{et al.}~\cite{zhang2019tensor} proposed a tensor network-based multi-task model that decomposes person Re-ID into camera-specific classification tasks and leverages low-rank tensor decomposition to capture cross-camera correlations while aligning feature distributions across different views. 
The SMART method~\cite{xu2019spatio} decomposes spatio-temporal data into interpretable latent factors and trains an ensemble of spatial-temporal predictors while incorporating domain constraints to handle large-scale spatio-temporal prediction tasks efficiently. A PEPS-like concatenated TN layer~\cite{wang2020concatenated} for multitask missions was also proposed, which, unlike the TTMT and TMT models that suffer from the negative transfer problem due to their hard sharing architectures, only contains a soft sharing layer, thereby achieving better performance. 

The MTCN method~\cite{duan2020novel} achieves superior face multi-attribute prediction by sharing all features in lower layers while differentiating attribute features in higher layers, by incorporating tensor canonical correlation analysis to exploit inter-attribute relationships. The CTNN method~\cite{jin2020ctnn} combines depthwise separable CNN and low-rank tensor networks to efficiently extract both local and global features from multi-task brainprint data, achieving high recognition accuracy with limited training samples while providing interpretable channel-specific biomarkers. The GTTN method~\cite{zhang2021multi} combines matrix trace norms from all possible tensor flattenings to automatically discover comprehensive low-rank structures in deep multi-task learning models, eliminating the need for manual specification of component importance. Zhang \textit{et al.}~\cite{zhang2021tensor} propose a tensor-based multi-task learning framework that leverages spatio-temporal similarities between brain biomarkers to predict Alzheimer's disease progression by encoding MRI morphological changes into a third-order tensor and extracting shared latent factors through tensor decomposition. 

Ren \textit{et al.}~\cite{ren2023multipar} extend PARAFAC2 with multi-task supervision for EHR mining, improving phenotype extraction and prediction through jointly modeled static and dynamic tasks. Wang \textit{et al.}~\cite{wang2024multi} combine Tucker-based fusion with self-supervised multi-task learning and a dual multimodal--unimodal training mechanism. Liu \textit{et al.}~\cite{liu2024non} map speech-quality features into a higher-dimensional tensor-network representation and jointly optimize regression, classification, and correlation objectives.

\subsection{Quantum (State-based) Data Representation}
\label{sec:quantum1}

To process machine learning tasks in a quantum system, the input data should be converted into a linear combination of some quantum states as an orthogonal basis; in the form
\begin{align}
\label{eq:quantum}
&|\psi\rangle=\sum_{d_{1} \ldots d_{N}=1}^{M} \tbtensor{A}_{d_{1} \ldots d_{N}}\left|\psi_{d_{1}}\right\rangle \otimes \cdots \otimes\left|\psi_{d_{N}}\right\rangle,\notag
\\&
s.t.\quad \sum_{d_{1} \ldots d_{N}=1}^{M} \left|\tbtensor{A}_{d_{1} \ldots d_{N}}\right|^2 = 1,\quad \tbtensor{A}_{d_{1} \ldots d_{N}}\in\mathbb{C},
\end{align}
where $|\cdot\rangle$ denotes a ket in Dirac notation~\cite{divincenzo1995quantum}, $\otimes$ denotes the tensor product, and $\tbtensor{A}$ is the complex amplitude tensor. The squared magnitudes of its entries sum to one, and the amplitude tensor can be represented and analyzed through a low-rank TN~\cite{biamonte2017tensor}.
To embed classic data into a quantum state for adapting quantum systems,
Stoudenmire and Schwab~\cite{stoudenmire2016supervised} proposed a quantum state mapping function $\phi^{i}(x_i)$ for the $i$-th pixel $x_i$ in a grayscale image as
\begin{align}
\label{eq:feature map1}
\phi^{i}(x_i)=[\cos(\frac{\pi}{2}x_i),\sin(\frac{\pi}{2}x_i)].
\end{align}
The values of pixels are transformed into the range from 0.0 to 1.0 via the mapping function.
Furthermore, a full grayscale image $\tbvector{x}$ can be represented as outer products of the mapped quantum states of each pixel as
\begin{align}
\label{eq:feature map1_2}
\Phi^{1,2,...N}(\tbvector{x})=\phi^{1}(x_1)\circ\phi^{2}(x_2)\circ\cdots\phi^{N}(x_N),
\end{align}
where $\Phi^{1,2,...,N}(\tbvector{x}) \in \mathbb{R}^{\overbrace{2\times 2\cdots \times 2}^{N}}$.
Through Eq.~(\ref{eq:feature map1_2}), it is feasible to associate realistic images with real quantum systems. 

\begin{figure}[t]
\centering
 \includegraphics[width=\columnwidth]{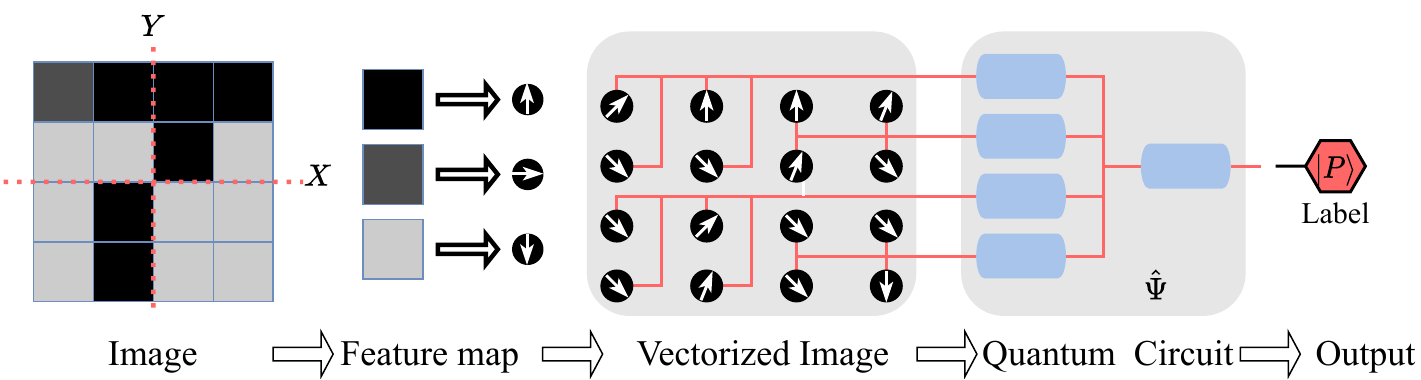}
\caption{The processing procedure employed for quantum embedded data~\cite{benedetti2019parameterized}. Quantum circuits can be simulated via TNNs on classic electronic computers, and some special TNNs (such as ConvAC) can also be theoretically implemented on a realistic quantum circuit.}
\label{fig:TTN2}
\end{figure}

For a natural language document, the $i$-th word $\left|x_{i}\right\rangle$ can also be represented as the sum of orthogonal quantum state bases $ \left|\phi_{h_{i}}\right\rangle\left(h_{i}=1, \ldots, M\right)$~\cite{ganmorphte12, li2019cnm, zhang2018quantum, miller2020tensor} corresponding to a specific semantic meaning $M$ as
\begin{align}
\label{eq:word}
&\left|x_{i}\right\rangle=\sum_{h_{i}=1}^{M} \alpha_{i, h_{i}}\left|\phi_{h_{i}}\right\rangle, \notag
\\& 
s.t.\quad \sum_{h_{i}=1}^{M} \left|\alpha_{i, h_{i}}\right|^2 = 1,\quad
\alpha_{i, h_{i}}\in\mathbb{C},
\end{align}
where $\alpha_{i,h_i}$ is the amplitude associated with each semantic basis state. Normalization requires the squared magnitudes of these amplitudes to sum to one; restricting them to nonnegative real values would be a model-specific assumption rather than a general requirement of quantum mechanics.
After completing data mapping, the embedded quantum data can be processed by TNNs on a realistic quantum circuit, as shown in Fig.~\ref{fig:TTN2}. The loss functions of TNNs can also be defined through the properties of quantum circuits. Such a procedure can be simulated on classic electronic computers via TNs and can be theoretically efficiently implemented on realistic quantum systems.

\section{Sustainable AI through TNNs: Compact Model Structures}
\label{Sec:CTDNN}

DNNs have extraordinarily high spatial and temporal complexity levels, as deeply stacked layers contain large-scale matrix multiplications. As a result, DNNs usually require several days for training while occupying a large amount of memory for inference purposes.
In addition, large weight redundancy has been proven to exist in DNNs~\cite{zeiler2014visualizing}, indicating the possibility of compressing DNNs while maintaining performance.
Motivated by this, a wide range of compression techniques have been developed, including pruning~\cite{DBLP:conf/cvpr/MolchanovMTFK19},
%,DBLP:journals/tnn/Karnin90
quantization~\cite{DBLP:conf/cvpr/YangSXTLDH019}, 
% ,DBLP:conf/aaai/ZhouMCF18
distillation~\cite{DBLP:conf/emnlp/JiaoYSJCL0L20}
% ,DBLP:journals/corr/HintonVD15
and low-rank decomposition~\cite{DBLP:conf/aaai/PanXWYWBX19}. Among them, applying TNs to DNNs to construct TNNs can be a good choice since TNNs have excellent abilities to approximate the original weights with much fewer parameters~\cite{DBLP:journals/pieee/PanagakisKCONAZ21}.
In this direction, researchers have completed many studies, especially concerning the reconstruction of convolutional and fully connected layers through a variety of TD formats~\cite{DBLP:conf/aaai/PanXWYWBX19,li2017bt,novikov2015tensorizing,DBLP:conf/nips/HayashiYSM19}.
With compact architectures, these TNNs can achieve improved performance with less redundancy.

\rthreehl{Pham \textit{et al.}~\cite{pham2026coupledcompact} exploit shared low-rank structure across related network parameter tensors through coupled tensor decomposition, rather than compressing each weight tensor independently.}
\rthreehl{Yang \textit{et al.}~\cite{yang2026kgn} treat connectivity topology as an algebraic design object and recursively generate architectures that interpolate between sparse and dense aggregation regimes. For scientific machine learning, Zhou \textit{et al.}~\cite{zhou2026tuckerfno} apply Tucker decomposition to a high-dimensional Fourier neural operator, yielding a sequence of lower-dimensional operators while retaining a universal-approximation guarantee. Together, these studies broaden the compact-model perspective from factorizing conventional weights to designing structured connectivity and efficient operator models within a common architectural framework.}

In this section, we examine how TNNs enable more sustainable AI through compact model structures. Modern DNNs, while powerful, often require extensive computational resources and memory due to their deep architectures and large-scale matrix operations. Additionally, studies have shown significant weight redundancy in DNNs, suggesting opportunities for compression without sacrificing performance. Here, we explore five key TNN architectures that address these challenges: TCNNs (Section~\ref{Sec:CNN}), TRNNs (Section~\ref{Sec:RNN}), tensorial Transformers (Section~\ref{Sec:transformer}), TGNNs (Section~\ref{Sec:GNNs}), and tensorial quantum neural networks (Section~\ref{Sec:TQNNs}). By leveraging tensor decomposition techniques and efficient parameter sharing, these approaches achieve significant model compression while maintaining or even enhancing performance compared to their conventional counterparts. We also examine applications of tensor networks in large language models (Section~\ref{Sec:LLMs}), where they enable parameter-efficient fine-tuning.

\subsection{TCNNs\label{Sec:CNN}}

\begin{figure*}[t]

\centering
\includegraphics[width=0.83\textwidth]{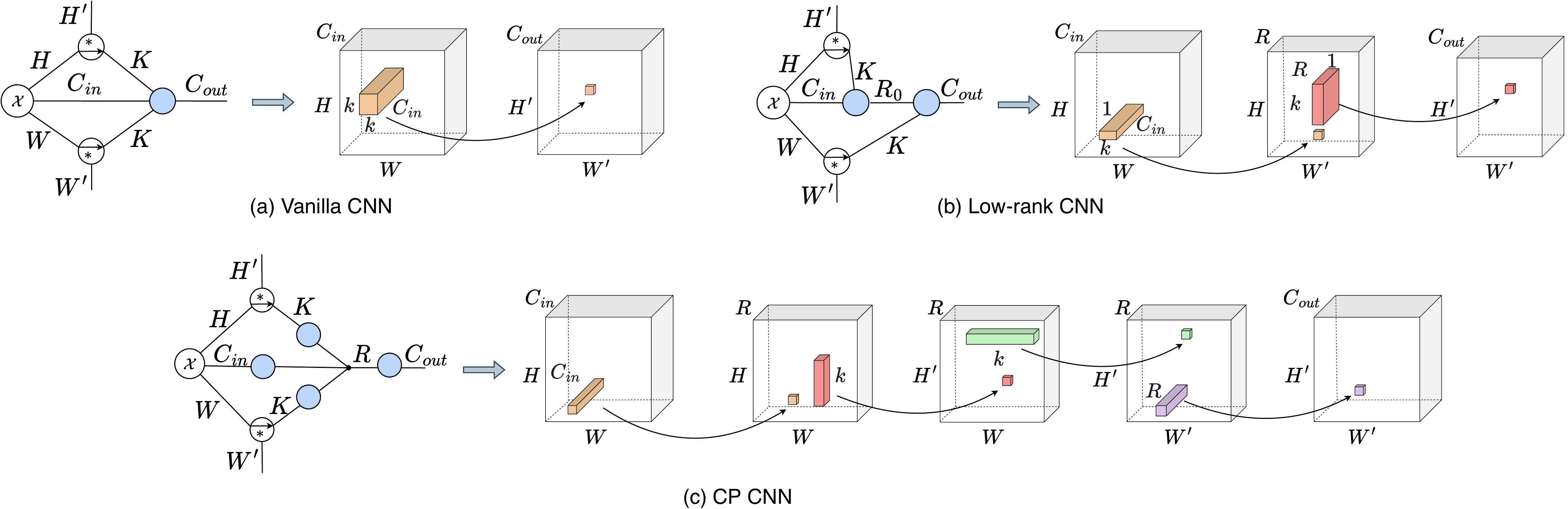}
\caption{\rwone{Correspondence between TN diagrams and convolutional procedures. In each subfigure, the left part is a TN diagram, and the right part is the associated feature representation. \textbf{(a)} depicts a standard convolutional layer with a fourth-order weight tensor. \textbf{(b)} illustrates a low-rank convolution in which the channel transformation is factorized through an intermediate rank. \textbf{(c)} shows CP-CNN, where the convolutional weight is represented by a sum of rank-one components and implemented through compact factor vectors.}}
\label{fig:conv-layers}
\end{figure*}

CNNs have achieved strong performance across a broad range of tasks. However, their large parameter counts cause weight redundancy and superfluous computations, affecting both performance and efficiency. TD methods provide one approach to this problem, and CNNs represented with tensor formats are called TCNNs.
Prior to introducing TCNNs, we first present a formulation of a vanilla CNN, shown in Fig.~\ref{fig:conv-layers} (a), in a standard tensorial form as
\begin{align}
    \tbtensor{Y}=\tbtensor{X}\circledast\tbtensor{C}+\mathbf{b},
\end{align}
where $\tbtensor{C} \in \mathbb{R}^{K\times K\times I\times O}$ denotes a convolutional weight, $\tbtensor{X}\in \mathbb{R}^{I\times H\times W}$ denotes an input, $\tbtensor{Y}\in \mathbb{R}^{O\times H'\times W'}$ denotes an output, $\mathbf{b}\in \mathbb{R}^{O}$ represents a bias, and $\circledast$ denotes a convolutional operator. $K$ represents the kernel window size, $I$ is an input channel, $H$ and $W$ denote the height and width of $\tbtensor{X}$, $O$ is an output channel, and $H'$ and $W'$ denote the height and width of $\tbtensor{Y}$, respectively.
TCNNs mainly focus on decomposing the channels $I$ and $O$. In detail, the weight $\tbtensor{C}$ is first reshaped to $\tilde{\tbtensor{C}} \in \mathbb{R}^{K\times K\times I_1 \times I_2 \times\dots I_M \times O_1 \times O_2 \times\dots O_N}$, where $\prod_{k=1}^M I_k=I$ and $\prod_{k=1}^N O_k=O$.
Then, TCNNs can be derived by tensorizing the reshaped convolutional kernel $\tilde{\tbtensor{C}}$.

To accelerate the CNN training and inference process, CP-CNN~\cite{denton2014exploiting, DBLP:journals/corr/LebedevGROL14,DBLP:conf/nips/HayashiYSM19,phan2020stable,nekooei2022compression} is constructed by decomposing the convolutional weight into the CP format, as shown in Fig.~\ref{fig:conv-layers}~(c).
CP-CNN contains vector factors as its principal subcomponents and can therefore achieve a high compression ratio when a sufficiently small CP rank is adequate.
As with CP-CNN, additional TCNNs can be constructed by applying the tensor formats in Fig.~\ref{fig:tensor-formats} to convolutional weights. Tucker decomposition is widely used for Tucker-CNNs~\cite{DBLP:journals/ijon/PanWX22,Liu2022DeepNN}. A BTT-CNN instead uses a hyperedge indexed by $R_c$ to represent a sum of Tucker blocks~\cite{li2017bt}. This block-term structure provides greater flexibility than a single Tucker block, but introduces additional rank choices and optimization cost.
Highly compact TT formats have also been introduced to CNNs to implement TT-CNNs\cite{liu2022tt,qi2023exploiting}.
Relative to TT, the cyclic topology of TR removes the fixed rank-one endpoints and can provide a more flexible parameterization at comparable rank settings~\cite{DBLP:conf/cvpr/WangSEWA18}; whether this improves compression or accuracy depends on the selected tensorization and ranks. To address degeneracy in tensorial layers, CPD-EPC~\cite{phan2020stable} uses a minimal-sensitivity design for CP convolutional layers and hybrid Tucker2-CP convolutional layers. The TR-Compress method~\cite{xie2024neural} combines tensor-ring factorization with execution scheduling to reduce parameters while controlling inference overhead.
There are also some tensorial convolutional neural networks that decompose more than just the convolution cores. The tensorized network (T-Net)~\cite{DBLP:conf/cvpr/KossaifiBTP19} treats the whole network as a one-layer architecture and then decomposes it. As a result, the T-Net achieves better results with a lighter structure. The CP-higher-order convolution (CP-HOConv)~\cite{DBLP:conf/cvpr/KossaifiTBPHP20} utilizes the CP format to efficiently model higher-order data, such as spatial, temporal, and channel-wise correlations, for tasks like spatiotemporal emotion estimation. \rthreehl{Obukhov \textit{et al.}~\cite{obukhov20a} model a \emph{set} of neural tensors with shared tensor-ring-style basis cores, showing that parameter sharing across multiple tensors can itself provide effective compression under resource constraints.}
\rwfour{Hamreras \textit{et al.}~\cite{hamreras2025tensorization} provide a comprehensive analysis of tensorization as a powerful but underexplored tool for compression and interpretability of neural networks, highlighting its potential for advancing mechanistic interpretability and proposing the vision of fully tensorized neural networks for future efficient and transparent model design.}

\aehl{To provide a quantitative example that does not depend on a particular benchmark, consider a stride-one, padded $3\times 3$ convolution with $256$ input channels, $256$ output channels, and a $56\times56$ output feature map. A dense kernel contains $3\times3\times256\times256=589{,}824$ parameters and requires approximately $1.850$ GMACs. Table~\ref{tab:analytical_cnn_example} compares this baseline with Tucker-2 and CP parameterizations over three rank settings. For the factorized implementations, the parameter counts are $C_i r_i+r_i r_o K^2+r_oC_o$ for Tucker-2 and $R(C_i+C_o+2K)$ for CP. Multiplying these expressions by $H_oW_o$ gives the corresponding forward-pass MAC counts when the intermediate feature maps retain the same spatial resolution. Bias additions, activations, data movement, and framework overhead are excluded so that every entry is directly verifiable.}

\begin{table*}[t]
\color{AEOrange}
\caption{Rank sensitivity in an analytical $3\times3$ convolutional example with $C_i=C_o=256$ and $H_o=W_o=56$. GMACs count the multiply-accumulate operations of the dense or factorized forward pass under equal spatial resolution.}
\label{tab:analytical_cnn_example}
\centering
\resizebox{0.90\textwidth}{!}{%
\begin{tabular}{lccccc}
\hline
\textbf{Method} & \textbf{Rank setting} & \textbf{Parameter formula} & \textbf{Params} & \textbf{Compression} & \textbf{GMACs} \\
\hline
Dense & -- & $K^2C_iC_o$ & $589{,}824$ & $1.00\times$ & $1.850$ \\
Tucker-2 & $r_i=r_o=32$ & $C_i r_i+r_i r_oK^2+r_oC_o$ & $25{,}600$ & $23.04\times$ & $0.080$ \\
Tucker-2 & $r_i=r_o=64$ & $C_i r_i+r_i r_oK^2+r_oC_o$ & $69{,}632$ & $8.47\times$ & $0.218$ \\
Tucker-2 & $r_i=r_o=128$ & $C_i r_i+r_i r_oK^2+r_oC_o$ & $212{,}992$ & $2.77\times$ & $0.668$ \\
CP & $R=32$ & $R(C_i+C_o+2K)$ & $16{,}576$ & $35.58\times$ & $0.052$ \\
CP & $R=64$ & $R(C_i+C_o+2K)$ & $33{,}152$ & $17.79\times$ & $0.104$ \\
CP & $R=128$ & $R(C_i+C_o+2K)$ & $66{,}304$ & $8.90\times$ & $0.208$ \\
\hline
\end{tabular}%
}
\end{table*}
\aehl{The rank sweep makes two effects explicit. First, compression is highly rank dependent. Increasing the Tucker-2 rank from $32$ to $128$ changes the compression ratio from $23.04\times$ to $2.77\times$, while increasing the CP rank over the same range changes it from $35.58\times$ to $8.90\times$. At rank $64$, Tucker-2 requires $69{,}632$ parameters and $0.218$ GMACs, whereas CP requires $33{,}152$ parameters and $0.104$ GMACs. The lower CP count follows from its stronger separability constraint; Tucker-2 retains a dense $r_i\times r_o\times K\times K$ interaction core and therefore allocates more capacity to cross-channel and spatial interactions.}

\aehl{Second, the benefit depends on layer geometry rather than rank alone. For a $1\times1$ convolution with the same channel dimensions and rank $64$, the dense layer contains $65{,}536$ parameters, Tucker-2 contains $36{,}864$ parameters, and CP contains $32{,}896$ parameters. The corresponding compression ratios fall to only $1.78\times$ and $1.99\times$, respectively, because there is little spatial structure to factorize. These calculations do not predict accuracy or measured latency, but they show why rank, tensor format, kernel size, and hardware execution must be evaluated jointly instead of reporting a single compression ratio.}

\aehl{We further conducted a controlled task-level benchmark to connect these analytical counts with accuracy and measured execution. The experiment uses CIFAR-10 and a three-convolution SimpleCNN with channel dimensions $3\rightarrow64\rightarrow128\rightarrow256$. The first convolution remains dense; only the second and third convolutions are replaced by Tucker-2 or CP layers. We use a fixed split of $45{,}000$ training, $5{,}000$ validation, and $10{,}000$ test images, train for 20 epochs with AdamW and a cosine schedule, and select the best validation checkpoint. Table~\ref{tab:empirical_cifar_tnn} reports the mean and sample standard deviation over seeds 11, 22, and 33. Training uses a batch size of 256 and mixed precision on one Tesla V100 PCIe GPU with 32GB memory under PyTorch 2.3.0 and CUDA 12.1. Supplementary Notes~S6 and~S7 provide the full protocol and per-seed records, and the accompanying package includes the execution script and machine-readable outputs.}

\begin{table*}[t]
\color{AEOrange}
\caption{Controlled CIFAR-10 benchmark for Dense, Tucker-2, and CP versions of SimpleCNN. The first convolution remains dense and the second and third convolutions are tensorized. Accuracy, total 20-epoch training time including validation, and full-test-set GPU inference time are reported as mean $\pm$ standard deviation over three seeds. MACs are counted per image. Peak denotes maximum allocated GPU memory during training.}
\label{tab:empirical_cifar_tnn}
\centering
\resizebox{0.98\textwidth}{!}{%
\begin{tabular}{llccccccc}
\hline
\textbf{Method} & \textbf{Tensor setting} & \textbf{Params(M)} & \textbf{CR} & \textbf{GMACs} & \textbf{Test Acc.(\%)} & \textbf{Train(s)} & \textbf{Infer(ms)} & \textbf{Peak(MB)} \\
\hline
Dense & None & $0.374$ & $1.00\times$ & $0.153$ & $78.78\pm0.03$ & $127.7\pm10.2$ & $316.9\pm53.2$ & $1176$ \\
Tucker-2 & rank ratio $0.25$ & $0.054$ & $6.94\times$ & $0.022$ & $72.59\pm0.27$ & $136.8\pm20.6$ & $155.0\pm34.6$ & $666$ \\
Tucker-2 & rank ratio $0.50$ & $0.149$ & $2.52\times$ & $0.060$ & $74.23\pm0.34$ & $130.6\pm15.6$ & $230.0\pm45.5$ & $764$ \\
CP & $R=32$ & $0.024$ & $15.57\times$ & $0.011$ & $71.46\pm0.11$ & $137.9\pm11.9$ & $166.5\pm36.8$ & $665$ \\
CP & $R=64$ & $0.043$ & $8.73\times$ & $0.021$ & $74.56\pm0.19$ & $136.3\pm23.1$ & $188.1\pm48.1$ & $666$ \\
\hline
\end{tabular}%
}
\end{table*}

\aehl{The empirical results expose trade-offs that parameter counts alone cannot show. CP with rank 64 retains $74.56\%$ accuracy while reducing parameters by $8.73\times$, MACs by $7.23\times$, and peak training memory by $43.4\%$. Its mean full-test inference time is $1.69\times$ faster than Dense, but its total training time is slightly longer and its accuracy is lower by $4.21$ percentage points. CP with rank 32 reaches the largest parameter compression, $15.57\times$, but increases the accuracy gap to $7.31$ points. Tucker-2 with rank ratio $0.25$ yields a $6.94\times$ parameter reduction and a $2.05\times$ inference speedup, yet its training time is again not lower than the dense baseline. These observations reinforce that theoretical MAC reductions do not translate proportionally into training speed because factorized layers introduce additional kernels, intermediate activations, and synchronization. A useful TNN comparison should therefore report accuracy, parameters, arithmetic cost, memory, and measured execution together.}

\begin{figure}[!b]
\centering

\includegraphics[width=0.42\textwidth]{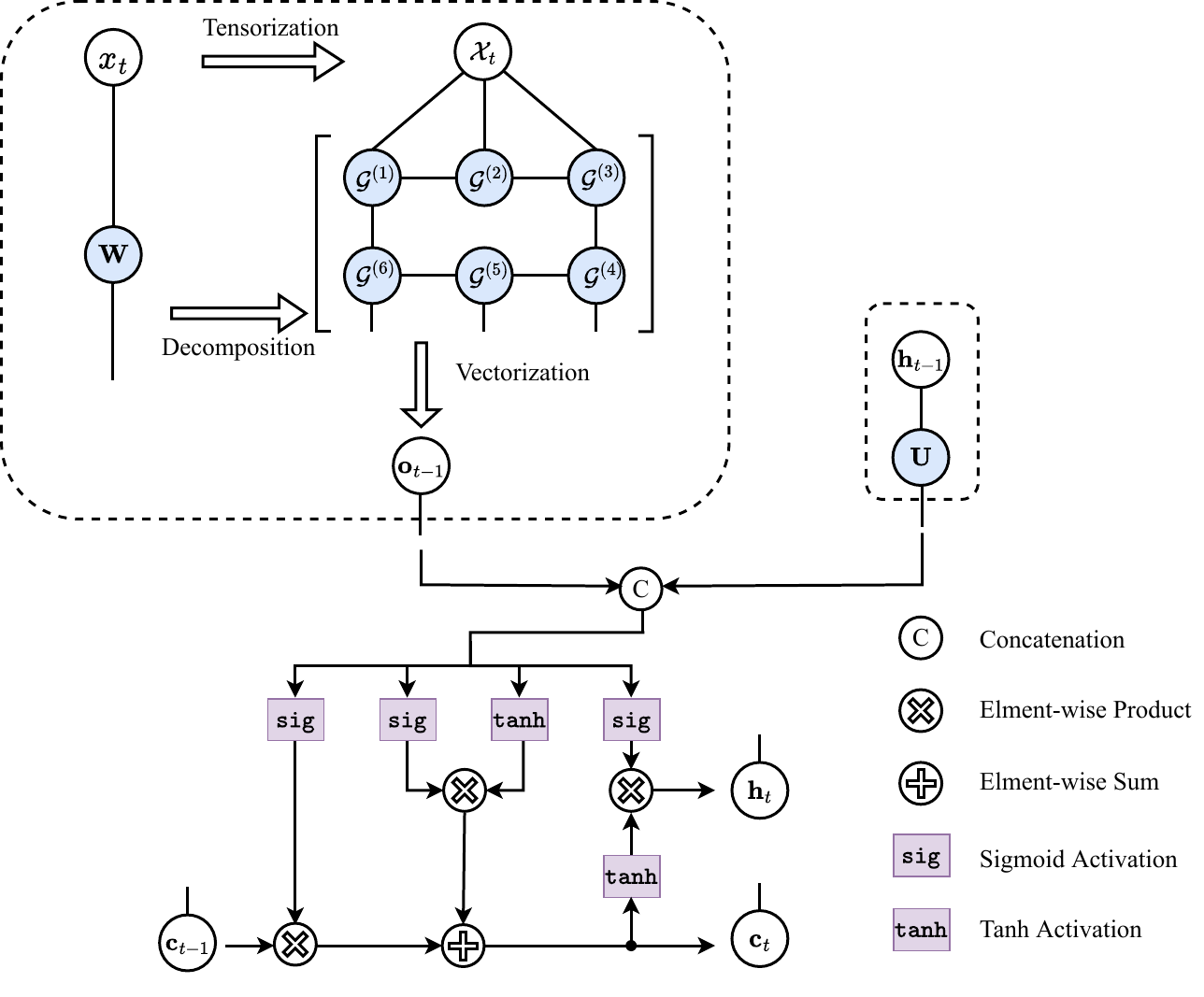}
\caption{The tensor ring LSTM. It is effective at reducing the parameters of an LSTM model by replacing the input-to-hidden transformation weights with TR decomposition.}
\label{fig:rnn-layers}
\end{figure}

\subsection{Tensor Recurrent Neural Networks\label{Sec:RNN}}

RNNs, such as the vanilla RNN and LSTM, have achieved promising performance on sequential data. However, when dealing with high-dimensional input data (e.g., video and text data), the input-to-hidden and hidden-to-hidden transformations in RNNs will result in high memory usage rates and computational costs. To solve this problem, low-rank TD is efficient for compressing the transformation process in practice. First, we formulate an RNN as
\begin{align}
\label{eq:rnn}
    \mathbf{h}^{(t+1)} = \phi(\mathbf{Wx}^{(t)} + \mathbf{Uh}^{(t)} + \mathbf{b}),
\end{align}
where $\mathbf{h}^{(t)} \in \mathbb{R}^O$ and $\mathbf{x}^{(t)} \in \mathbb{R}^I$ denote the hidden state and input feature at time $t$, respectively, $\mathbf{W} \in \mathbb{R}^{O\times I}$ is the input-to-hidden matrix, $\mathbf{U} \in \mathbb{R}^{O\times O}$ represents the hidden-to-hidden matrix, $\mathbf{b} \in \mathbb{R}^{O}$ is a bias, while $\phi(\cdot)$ indicates a series of operations that form RNN variants, including the vanilla RNN and LSTM~\cite{hochreiter1997long}. Eq.~\eqref{eq:rnn} can also be reformulated in a concatenated form that is widely used in TD, given by
\begin{align}
    \mathbf{h}^{(t+1)} = \phi([\mathbf{W}, \mathbf{U}] [\mathbf{x}^{(t)}, \mathbf{h}^{(t)}] + \mathbf{b}),
\end{align}
where $[\mathbf{W}, \mathbf{U}] \in \mathbb{R}^{O\times (I+O)}$ and $[\mathbf{x}^{(t)}, \mathbf{h}^{(t)}] \in \mathbb{R}^{(I + O)}$ denote the concatenation of $\mathbf{W}, \mathbf{U}$ and $\mathbf{x}^{(t)}, \mathbf{h}^{(t)}$, respectively. As shown in Fig.~\ref{fig:rnn-layers}, there are usually two ways to decompose RNNs: (a) only tensorizing $\mathbf{W}$, which is often the largest component in an RNN, and (b) tensorizing $[\mathbf{W}, \mathbf{U}]$ for extreme compression.
Because $\mathbf{U}$ is usually smaller than $\mathbf{W}$, the surveyed methods typically tensorize $\mathbf{W}$ alone or tensorize the concatenated transformation $[\mathbf{W},\mathbf{U}]$, rather than decomposing $\mathbf{U}$ in isolation. The process of implementing a TRNN is analogous to that used for a TCNN: the relevant weights are reshaped into higher-order tensors and replaced with tensor-network parameterizations.

The most direct compression strategy is to decompose the input-to-hidden matrix $\mathbf{W}$. CP-RNN and Tucker-RNN~\cite{DBLP:journals/ijon/PanWX22} apply CP and Tucker parameterizations, respectively. At a sufficiently small rank, CP-RNN can be among the most compact alternatives, although its stronger rank-one structure may also reduce approximation accuracy.
The TT-RNN~\cite{DBLP:conf/icml/YangKT17} implements the TT format on an RNN to obtain a high parameter compression ratio. However, the TT-RNN suffers from a linear structure with two smaller endpoints, which hinders the representation ability and flexibility of TT-based models. To release the power of a linear architecture, TRs were proposed to link the endpoints to create a ring format~\cite{zhao2016tensor}. The TR-RNN~\cite{DBLP:conf/aaai/PanXWYWBX19} uses this ring structure to obtain a more compact recurrent model. The BTT-RNN~\cite{Ye_2018_CVPR,li2017bt} was constructed on the generalized TD approach, the BTT decomposition~\cite{de2008decompositions2}. The BTT-RNN can automatically learn interparameter correlations to implicitly prune redundant dense connections and simultaneously achieve better performance.
\rthreehl{Lizaire \textit{et al.}~\cite{lizaire2024secondorderrnn} interpret second-order RNNs through tensors, clarifying when multiplicative recurrent interactions are useful and how they relate to classical TN parameterizations rather than treating decomposition only as a post-hoc compression tool.}

\begin{figure*}[t]
\centering
\includegraphics[width=0.88\textwidth]{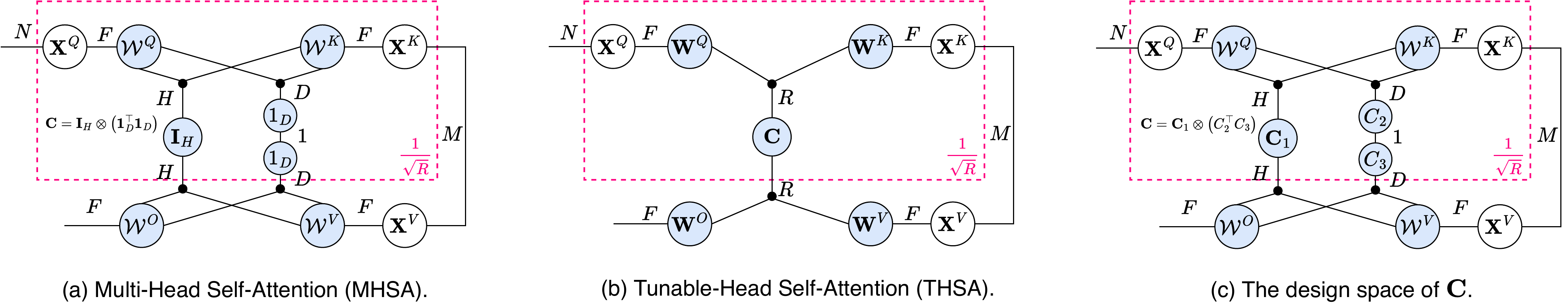}
\caption{Tensor diagrams for SA modules~\cite{DBLP:conf/acl/LiuGZXLW20}. (a) It is feasible to represent a classic multihead SA (MHSA) mechanism as a tensor diagram. MHSA can be treated as a special case of tunable-head self-attention (THSA) by setting $\mathbf{C}=\mathbf{I}_H\otimes (\mathbf{1}_D^\top\mathbf{1}_D)$. (b) The THSA of the Tuformer can be a more generalized version of SA through a trainable matrix $\mathbf{C}$. (c) THSA has a design space formulated as $\mathbf{C}=\mathbf{C}_1\otimes (C_2^\top C_3)$, which is the direct generalized form of MHSA.}
\label{fig:tuformer}
\end{figure*}

Moreover, studies are utilizing TD to compress an RNN's two transformation layers, and some have even developed decomposition methods that are suitable for both RNNs and CNNs. The TT-GRU~\cite{DBLP:journals/ieicet/TjandraSN20} and the HT-RNN~\cite{DBLP:conf/cvpr/YinLLW021} methods decompose $[\mathbf{W}, \mathbf{U}]$ to attain a higher compression ratio. Specifically, TT-GRU~\cite{DBLP:journals/ieicet/TjandraSN20} applies a TT for decomposition, and the HT-RNN~\cite{DBLP:conf/cvpr/YinLLW021} adopts HT decomposition. Unlike prior works that decompose hidden matrices, Conv-TT-LSTM~\cite{DBLP:conf/nips/SuBKHKA20} utilizes the idea of a TT to represent convolutional operations. As shown in Fig.~\ref{fig:rnn-layers}, through a TT-like convolution, Conv-TT-LSTM can replace convolutional LSTM with fewer parameters while achieving good results on action benchmarks.
For the adaptation of both CNNs and RNNs, a hybrid TD (termed HT-TT) method that combines HT and TT decomposition~\cite{DBLP:journals/nn/WuWZDL20} was adopted to compress both the CNN and RNN $[\mathbf{W}, \mathbf{U}]$ matrices. 
The MPS-NLP~\cite{tangpanitanon2022explainable} proposes tensor recurrent neural networks (TRNNs) through matrix product states and entanglement entropy, enabling explainable natural language processing while maintaining model performance.
In addition, the tensor contraction layer (TC-Layer)~\cite{kossaifi2017tensor1} was designed to replace the fully connected layer and therefore can be utilized as the last layer of a CNN and the hidden layers in RNNs. Interestingly, TC-Layer can be viewed as a special case of a TT-based layer obtained by setting all TT ranks to 1, which yields a simpler factorized structure with reduced expressive flexibility.

\subsection{Tensorial Transformers\label{Sec:transformer}}

Transformers~\cite{vaswani2017attention,lubich2013dynamical} are well known for processing sequence data. Compared with CNNs and RNNs, Transformers can be stacked into large-scale sizes to achieve significant performance gains~\cite{DBLP:conf/naacl/DevlinCLT19}. However, Transformers are still redundant, similar to classic DNNs, and can be made smaller and more efficient~\cite{wangExploringExtremeParameter2021}. Therefore, tensor networks can be explored to reduce the numbers of parameters in Transformers~\cite{ma2019tensorized,DBLP:conf/acl/LiuGZXLW20,liuTuformerDataDrivenDesign2021}.

Classic Transformers mainly consist of the self-attention (SA) mechanism and feedforward Networks (FFNs).
The SA processes the given query matrix $\tbmatrix{q}$, key matrix $\tbmatrix{k}$ and value matrix $\tbmatrix{v}$ with parameters $\tbmatrix{w}^Q, \tbmatrix{w}^{K}, \tbmatrix{w}^V, \tbmatrix{w}^{O}$. More generally, SA is separated into $n$ heads: $\{\tbmatrix{w}^Q_i\}^n, \{\tbmatrix{w}^{K}_i\}^n, \{\tbmatrix{w}^V_i\}^n, \{\tbmatrix{w}^{O}_i\}^n$. Each head can be calculated as
\begin{align}
\operatorname{Att}_i(\tbmatrix{q}, \tbmatrix{k}, \tbmatrix{v}) = \operatorname{softmax}\left(\frac{\tbmatrix{q}\tbmatrix{w}^Q_i\tbmatrix{w}^{K^T}_i \tbmatrix{K}^{T}}{\sqrt{d}}\right) \tbmatrix{v}\tbmatrix{w}^V_i\tbmatrix{w}^{O^T}_i.
\end{align}
Then, $\operatorname{SA}((\tbmatrix{q}, \tbmatrix{k}, \tbmatrix{v})) = \sum^n_{i=1}{\operatorname{Att}_i(\tbmatrix{q}, \tbmatrix{k}, \tbmatrix{v})}$. Another important component, the FFN, is formulated as
\begin{align}
\operatorname{FFN}(\tbmatrix{x}) = 
\operatorname{ReLU}(\tbmatrix{x}\tbmatrix{w}^{in}+\tbvector{b}^{in})\tbmatrix{w}^{out}+\tbvector{b}^{out},
\end{align}
where $\tbmatrix{X}$ is the input, $\tbvector{b}^{in}$ and $\tbvector{b}^{out}$ are biases, and $\tbmatrix{w}^{in}$ and $\tbmatrix{w}^{out}$ are weights. The number of parameters in a Transformer is mainly based on its linear transformation matrices, i.e., $\tbmatrix{w}^Q, \tbmatrix{w}^{K}, \tbmatrix{w}^V, \tbmatrix{w}^{O}$, $\tbmatrix{w}^{in}$ and $\tbmatrix{w}^{out}$.

Therefore, most compression studies focus on eliminating the parameters of these matrices. For instance,
the MPO structure was proposed to decompose each matrix in a Transformer~\cite{DBLP:conf/acl/LiuGZXLW20}, generating central tensors (containing the core information) and small auxiliary tensors. A tuning strategy was further adopted to continue training the auxiliary tensors to achieve a performance improvement while freezing the weight of the central tensor to retain the main information of the original matrix.
Moreover, observing that a low-rank MPO structure can cause a severe performance drop, Hypoformer~\cite{benyouformer} was proposed based on hybrid TT decomposition; this approach concatenates a dense matrix part with a low-rank MPO part.
Hypoformer retains the full-rank property while reducing the required numbers of operations and parameters to compress and accelerate the base Transformer. In addition, by concatenating all matrices into one larger tensor, Tucker-Bert~\cite{wangExploringExtremeParameter2021} decomposes the concatenated tensor with Tucker decomposition to greatly reduce the number of parameters, leading to extreme compression and maintaining comparably good results. 

\rthreehl{Xie \textit{et al.}~\cite{xie2026sweet} extend tensorized compression to parameter transfer across model scales by learning a shared self-supervised weight template and size-specific lightweight scalers under Tucker factorization. Their formulation initializes vision models with different depths and widths across classification, detection, segmentation, and generation tasks, indicating that tensorization can serve as a reusable cross-scale parameter prior rather than only as a post-hoc compression operator.}
Compared to compressing the original attention operation, multiway multimodal transformer~(MMT)~\cite{tang2022mmt} explores a novel generalized tensorial attention operation to model modality-aware multiway correlations for multimodal datasets.
Interestingly, Tuformer~\cite{liuTuformerDataDrivenDesign2021} generalizes MHSA into the Tucker form, thus containing more expressive power and achieving better results, as shown in Fig.~\ref{fig:tuformer}. Advances in tensorial causal learning have emerged through the implementation of causal capsules and tucker-format tensor transformers for latent variable interaction control. 
\rthreehl{Klein \textit{et al.}~\cite{klein2026tuckerattention} interpret MHA, GQA, and MLA within a unified Tucker-style low-rank view of approximate attention. Their analysis connects classical tensor decomposition with attention design, KV-cache reduction, and parameter-efficient Transformer deployment. Zhang \textit{et al.}~\cite{zhang2025tensornew} use tensor-product parameterization to compress the KV cache for long-context language modeling while retaining performance against MHA, MQA, GQA, and MLA.}

% We can design novel neural network architectures by utilizing various low-rank forms, including those without well-established numerical decomposition algorithms.

\subsection{TGNNs\label{Sec:GNNs}}
Graph Neural Networks~(GNNs) have achieved groundbreaking performances across a range of applications and domains~\cite{zhou2020graph}.
One classic GNN layer consists of an aggregation function for aggregating the neighbor node information and an update function for updating the current node information. For example, the processing step for node $v$ in the $k$-th layer of a GNN can be formulated as
\begin{equation}
\begin{aligned}
& \tbvector{a}_{v}^{(k)} \leftarrow \operatorname{Aggregate}_{(k)}\left(\left\{\tbvector{h}_u^{(k-1)}, \forall u \in \mathcal{N}(v)\right\}\right), \\
& \tbvector{h}_v^{(k)} \leftarrow \operatorname{Update}_{(k)}\left(\tbvector{h}_v^{(k-1)}, \tbvector{a}_{v}^{(k)}\right),
\end{aligned}
\end{equation}
where $\tbvector{a}_{v}^{(k)}$ is
an aggregated embedding vector, $\tbvector{h}_v^{(k-1)}$ is a node embedding vector, and $\mathcal{N}(v)$ is a neighbor node set. A typical choice for the update function is a simple one-layer perceptron, and simple summation/maximization is always chosen as the aggregation function.
Classic GNNs suffer from low model expressivity since high-order nonlinear information among nodes is missed~\cite{hua2022high}.
Because of the merits of the tradeoff between expressivity and computing efficiency, the usage of TGNNs for graph data processing is quite beneficial.

To efficiently parameterize permutation-invariant multilinear maps for modeling the interactions among neighbors in an undirected graph structure, a TGNN~\cite{hua2022high} makes use of a symmetric CP layer as its node aggregation function. It has been demonstrated that a TGNN has a strong capacity to represent any multilinear polynomial that is permutation-invariant, including the sum and mean pooling functions.
Nimble GNN~\cite{yin2022nimble} innovatively applies tensor-train decomposition to GNN embeddings with graph-aware tensor operations, achieving up to 81,362$\times$ compression while maintaining accuracy.
Compared to undirected graph processing, TGNNs are more naturally suited for high-order graph structures, such as knowledge graphs or multi-view graph.
Traditional relational graph convolutional networks neglect the trilinear interaction relations in knowledge graphs and additively combine the information possessed by entities. 

The RTGNN~\cite{zhao2022multi}, which applies a Tucker format structure to extract the graph structure features in the common feature space, was introduced to capture the potential high order correlation information in multi-view graph learning tasks.
\rthreehl{Wang \textit{et al.}~\cite{wang2026tginee} extend tensor-based graph learning to multilayer and multi-graph settings in which relation layers should neither be processed independently nor collapsed by simple aggregation. They combine CP-derived shared latent factors with generalized estimating equations and task-specific losses to model cross-layer correlations explicitly. Their analysis broadens the role of tensorization in graph learning from parameter compression and high-order aggregation to statistically coupled representation learning across multiple networks, with consistency and asymptotic normality guarantees.}
TGNNs are also appropriate for high-order correlation modeling in dynamic spatial-temporal graph processing situations. For example,
tensorized spatiotemporal GNN designs further explore dynamic time relations and spatial relations jointly, including PEPS-inspired modules that reduce parameter counts while capturing richer cross-time and cross-space dependencies.

% \subsection{Tensorial RBMs\label{Sec:RBM}}
% RBMs~\cite{hinton2012practical} are generative stochastic NNs that can learn a probability distribution from an input set.
% A standard RBM consists of one visible factor $\tbvector{v} \in \mathbb{R}^M$ and one hidden factor $\tbvector{h} \in \mathbb{R}^N$ and assigns the following energy function for a joint vector $\{\tbvector{v},\tbvector{u}\}$ as
% \begin{equation}
%     E(\tbvector{v}, \tbvector{h})=-\tbvector{v}^T \tbvector{W} \tbvector{h}-\tbvector{v}^T \tbvector{b}-\tbvector{c}^T \tbvector{h},
% \end{equation}
% where $\tbvector{b} \in \mathbb{R}^M$ and $\tbvector{c} \in \mathbb{R}^N$ are the biases of the visible layer and hidden layer, respectively, and $\tbmatrix{W} \in \mathbb{R}^{M \times N}$ is the mapping weight matrix. The probability distribution of
% the joint vector $\{\tbvector{v},\tbvector{u}\}$ can be defined as
% \begin{equation}
% P(v, h)=\frac{1}{Z} e^{-E(\tbvector{v}, \tbvector{h})},
% \end{equation}
% where $Z$ is a partition function. Then, some loss function can be defined via the learnable distribution formulation to optimize the parameters. The RBM parameters appear to be based mostly on the mapping weight matrix.

\subsection{Tensorial Quantum Neural Networks\label{Sec:TQNNs}}
\label{Sec:ConvAC}
\begin{figure}[t]
\centering
 \includegraphics[width=\columnwidth]{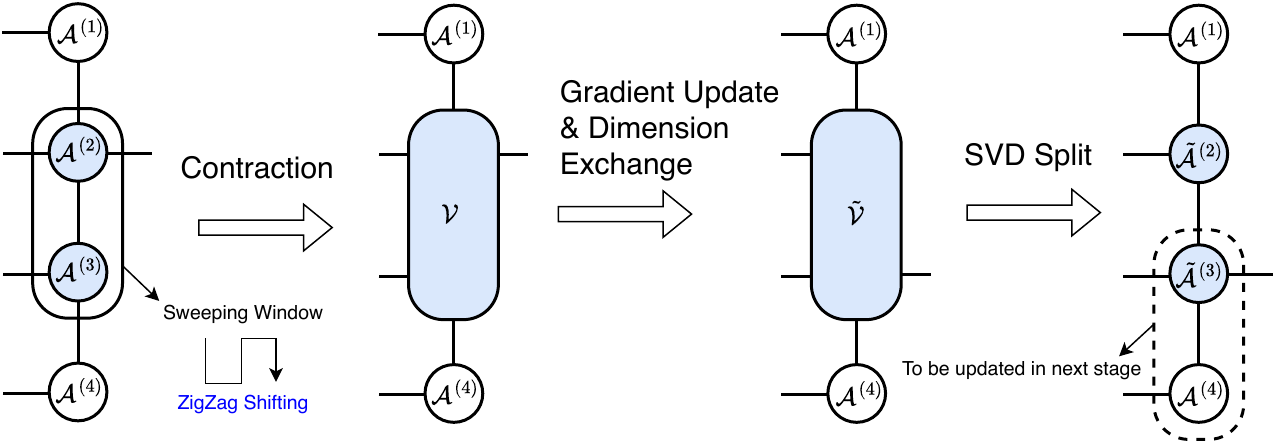}
\caption{A single stage of the Sweeping method~\cite{stoudenmire2016supervised}. In each stage, Sweeping only updates the nodes in the sweeping window, which shifts along a zigzag trajectory.}
\label{fig:sweep2}
\end{figure}
Quantum neural networks aim to process quantum data directly in quantum systems. One representative work bridging TNNs with quantum data processing is the MPS-based architecture proposed by Stoudenmire and Schwab~\cite{stoudenmire2016supervised}, which formulates the classification task of quantum mapped image data~(as introduced in Sec~\ref{sec:quantum1}) as optimizing functions indexed by labels, given by

\begin{align}
    c=\frac{1}{2} \sum_{n=1}^{N_{T}} \sum_{\ell}\left(f^{\ell}\left(\tbvector{x}_{n}\right)-\tbvector{y}_{n \ell}\right)^{2},
\end{align}
where $N_{T}$ denotes the number of training samples, and $\tbvector{y}_{n}$ denotes the true one-hot label vector of $\tbvector{x}_{n}$. The optimization process is carried out to minimize this cost function in stages with stochastic gradient descent.
A single stage is shown in Fig.~\ref{fig:sweep2}. In each stage, two MPS tensors $\tbtensor{A}^{(2)}$ and $\tbtensor{A}^{(3)}$ are combined into a single bond tensor $\tbtensor{V}$ via tensor contraction. Then, the tensor $\tbtensor{V}$ is updated with gradients. Finally, $\Tilde{\tbtensor{V}}$ is decomposed back into separate tensors with the SVD algorithm.
This work establishes a crucial connection between quantum physics and machine learning: the MPS structure, originally developed for quantum many-body systems, naturally bridges quantum-inspired tensor methods with neural architectures, where the bond dimensions serve as model complexity controls. The Sweeping method demonstrates how quantum-inspired optimization techniques can be effectively adapted for machine learning tasks. Furthermore, this framework's extensibility to other tensor network structures like PEPS~\cite{cheng2021supervised} suggests its potential for advancing both quantum and classical architectures while maintaining computational tractability.

The expressive power of previously developed quantum data processing models, e.g., the MPS models~\cite{torchmps} and the Born machine~\cite{born1926quantenmechanik}, suffers from a lack of nonlinearity. Classic nonlinear operators, e.g., activation functions (such as the rectified linear unit (ReLU) function) and average/max pooling, can significantly benefit model performance. However, classic nonlinearity cannot be directly implemented in a quantum circuit. 
To solve this problem, the ConvAC network~\cite{cohen2016expressive,levine2019quantum} was proposed to adopt quantum deployable product pooling as a nonlinear operator, proving that ConvAC can be transformed into ConvNets with ReLU activations and average/max pooling. The whole structure of ConvAC can be represented by an HT format and has been proven to be theoretically deployable in realistic quantum systems.

\begin{figure}[t]
\centering
 \includegraphics[width=0.42\textwidth]{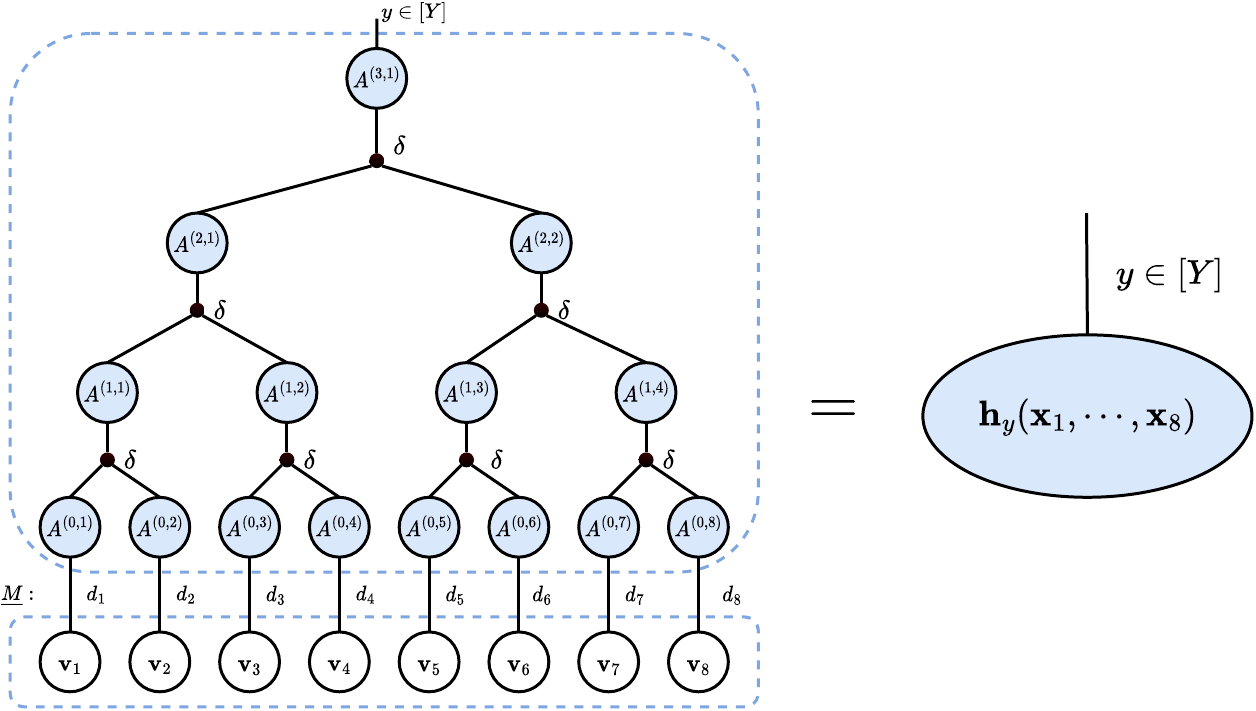}
\caption{ConvAC is equivalent to an HT-like TN~\cite{cohen2016convolutional}. $\tbvector{x}_{1}, \ldots, \tbvector{x}_{N} \in \mathbb{R}^{s}$, where $\tbvector{x}_{1}$ corresponds to a local patch from the input image or the feature map, and $v^{(0,j)}$ is the linear transformation of $\tbvector{x}_{j}$. The width is 2 for a single block. Notably, a single layer is equivalent to a CP format.~}
\label{fig:ConvAC}
\end{figure}

A tensor diagram example of ConvAC is shown in Fig.~\ref{fig:ConvAC}, where one hidden layer of ConvAC is in a CP format. ConvAC can also handle language data \cite{zhang2018quantum} by mapping natural language sentences into quantum states via Eq.~\eqref{eq:word}.
ConvAC is a milestone in that deep convolutional networks, along with nonlinear modules, are implemented on quantum circuits. It serves as an inspiration for the integration of more NNs into quantum systems. This has led to several important developments.
First, Zhang \textit{et al.}~\cite{zhang2019generalized} introduced the tensor space language model (TSLM), which generalizes the n-gram language model. 

Chen \textit{et al.}~\cite{chen2023antn} subsequently bridge tensor networks and autoregressive neural networks through matrix product states, enabling efficient quantum many-body simulation while preserving physical priors and model expressivity.
Chen \textit{et al.}~\cite{chen2024machine} combine tree tensor networks with CP-rank constraints and tensor dropout to improve classification efficiency and accuracy. Lu \textit{et al.}~\cite{lu2025variational} use a variational Monte Carlo algorithm based on R\'enyi free energy to approximate finite-temperature quantum states with tensor and neural networks, and show that string-bond states outperform restricted Boltzmann machines.

\subsection{Tensor Networks in Large Language Models\label{Sec:LLMs}}

\aehl{While LLMs share the same core tensor operations as the tensorized Transformers discussed in Section~\ref{Sec:transformer}, we keep a separate subsection for three reasons: (i) \emph{scale}, because LLMs operate at parameter counts that make memory-aware tensorization a first-order concern; (ii) \emph{deployment}, because inference-time issues such as KV-cache growth and multi-token prediction are especially prominent; and (iii) \emph{adaptation}, because tensorized parameter-efficient fine-tuning has become a distinct sub-literature. Because a substantial fraction of this literature remains in preprint form, we summarize representative directions rather than attempting an exhaustive catalogue.}

\aehl{Xu \textit{et al.}~\cite{xu2023tensorgpt} tensorize token embeddings for model compression, while Yang \textit{et al.}~\cite{yang2024loretta} use tensor-train adaptation for parameter-efficient fine-tuning. At inference time, KV-cache reduction and dynamic tensorization are especially relevant because they target one of the dominant memory bottlenecks in LLM deployment.}
\aehl{More broadly, tensorized PEFT methods suggest that structured adapters can preserve more multilinear structure than matrix-only LoRA variants. Structured matrix parameterizations such as Monarch Mixer~\cite{fu2023monarch} are also relevant here because they reveal a broader continuum between tensor decompositions, Kronecker/butterfly factorizations, and hardware-efficient sequence models. \rthreehl{Qiu \textit{et al.}~\cite{qiu24f} reinforce this connection by showing that BTT-style structured matrices can improve the compute--accuracy trade-off in vision transformers and small GPT-style models.}}

\rthreehl{Ye \textit{et al.}~\cite{ye2026trac} share or freeze selected TT cores across layers and use lightweight controllers to preserve layer-specific flexibility. Zheng \textit{et al.}~\cite{zheng2026refta} avoid repeated reconstruction of tensorized weights during adaptation to improve the accuracy--memory trade-off, whereas Li \textit{et al.}~\cite{li2026lestd} combine a shared orthogonal basis with an importance-pruned sparse core to alleviate the storage bottleneck of dense Tucker-style cores. These studies show that tensorized adaptation depends not only on the decomposition format, but also on cross-layer sharing, reconstruction cost, and core sparsity.}

\rthreehl{Mixture-of-experts models provide another important setting. Xu \textit{et al.}~\cite{xu2026tdmoe} jointly tensorize all experts within a layer so that a shared three-dimensional factorization captures cross-expert redundancy. In a complementary formulation, Xu \textit{et al.}~\cite{xu2026time} represent lifelong multimodal knowledge edits as compact CP-decomposed experts and use the input factors for intrinsic routing. These results connect tensor decomposition with expert sharing, routing, continual adaptation, and memory-efficient model editing.}

\rwone{\textbf{On the Relationship Between Tensor Representations and Architectures.} A natural question arises regarding how different tensor representations interact with different neural architectures. We offer the following observations: (1) \emph{For the same architecture, different tensor representations offer different trade-offs.} For instance, in CNNs, TR decomposition generally outperforms TT due to its ring structure providing more flexibility at endpoints, while CP offers the highest compression ratio but may sacrifice accuracy. Tucker decomposition provides a balanced middle ground. (2) \emph{The same tensor representation manifests differently across architectures.} TT decomposition in CNNs primarily targets convolutional kernel channels; in RNNs, it decomposes input-to-hidden and hidden-to-hidden transformation matrices; in Transformers, it compresses attention weight matrices and FFN layers. Despite sharing the same mathematical framework, the specific design choices (e.g., reshaping strategies, rank allocation) differ substantially across architectures. (3) \emph{Architectural affinity of tensor decompositions.} Certain decompositions show natural affinity with specific architectures: BTT is particularly well-suited for CNNs, as its hyperedge structure naturally captures channel-wise summation of Tucker blocks, enabling implicit pruning of redundant connections~\cite{li2017bt}; MPO is especially effective for Transformers, as its structure aligns with the multi-head attention paradigm, allowing compression of attention weights while preserving head-specific information~\cite{DBLP:conf/acl/LiuGZXLW20}; and MPS/TT formats naturally fit sequential data processing in RNNs due to their linear chain topology. However, whether certain decompositions are \emph{exclusively} suitable for certain architectures remains an open question that warrants further investigation.}

\begin{figure*}[t]
    \centering
    \includegraphics[width=0.9\linewidth]{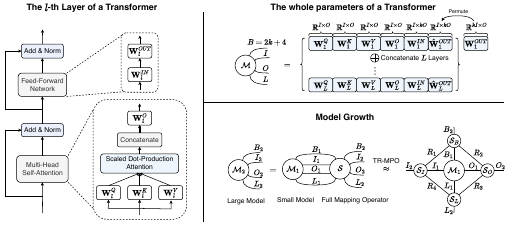}
    \caption{The architecture of MANGO, showing how it establishes comprehensive linear correlations across the entire Transformer structure, including Multi-Head Self-Attention (MHSA), Feed-Forward Networks (FFN), and normalization layers. The full mapping operator leverages TR-MPO to correlate parameters between small and large models.}
    \label{fig:mango}
\end{figure*}
\begin{figure*}[t]
\centering
\includegraphics[width=0.96\textwidth]{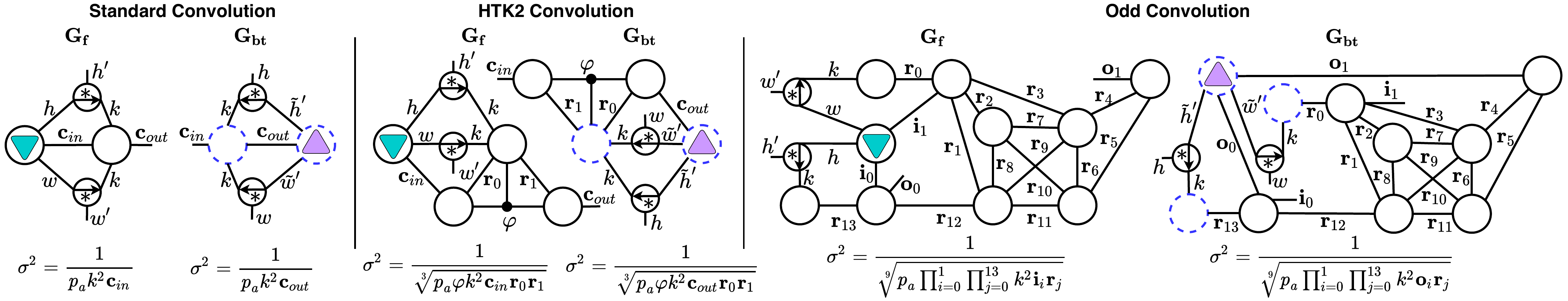}
\caption{Three cases of unified TCNN initialization~\cite{DBLP:conf/icml/0005SLW0X22}, where $\sigma^2$ denotes the initial variance of each weight vertex. $\mathbf{G}_f$ denotes a forward procedure, and $\mathbf{G}_b$ denotes a backward procedure.
(i)~Standard convolution. The method in \cite{DBLP:conf/icml/0005SLW0X22} degenerates to Xavier/Kaiming initialization on the standard convolution for the same weight variance formulation.
(ii)~Hyper Tucker-2 (HTK2) convolution. Tucker-2 (TK2) is a common TD that is utilized in ResNet as the bottleneck module~\cite{DBLP:conf/cvpr/HeZRS16}. HTK2 is formed by applying a hyperedge to the weight vertices of TK2.
(iii)~Odd convolution. The odd TD was originally proposed by Li and Sun~\cite{li20l}. The connections among the vertices are irregular, making weight initialization a complex problem. These three successful initialization cases can better demonstrate the potential adaptability of unified initialization to diverse TCNNs.}
\label{fig:yuinit}
\end{figure*}

\vspace{3mm}
\section{Training Strategies for TNNs\label{Sec:strategies}}

While the aforementioned TNNs can perform well on various tasks and machines, it is also worth exploring training strategies with more stability, better performance and higher efficiency.
In this section, we introduce such strategies in three groups: (1) strategies for stabilizing the training processes of TNNs are presented in Section~\ref{sec:stable-training}, (2) strategies for selecting and searching the ranks of TNNs are provided in Section~\ref{Sec:select}, and (3) strategies for applying hardware speedup are shown in Section~\ref{sec:hardware-implementation}.

\subsection{Stable Training Approaches}\label{sec:stable-training}
Despite their success, TNNs face significant training challenges stemming from their inherent multilinear characteristics. While traditional neural networks primarily rely on simple linear operations like matrix multiplication, TNNs involve tensor contractions that result in exponentially scaling data flows as the number of modes increases linearly~\cite{DBLP:conf/icml/0005SLW0X22}. This exponential scaling affects both the forward propagation of features and the backward propagation of gradients, creating substantial computational and numerical stability challenges.
Several approaches have been proposed to address these issues. One straightforward solution involves using full-precision float64 format to represent large weights, which helps mitigate numerical instability problems. However, this approach comes with significant drawbacks - the higher precision format requires more computational resources and increases processing time compared to lower-precision alternatives like float16. Conversely, while lower precision formats offer computational efficiency, they can introduce numerical stability issues that compromise training effectiveness.

To balance these competing concerns, Panagakis~\textit{et al.}~\cite{DBLP:journals/pieee/PanagakisKCONAZ21} introduced an innovative mixed-precision strategy. This dynamic precision approach adaptively adjusts numerical precision during different phases of computation, effectively creating a trade-off between computational efficiency and numerical stability. 
By selectively applying higher precision only where necessary, this strategy successfully reduces memory requirements while maintaining training stability. This approach has proven particularly effective in handling the complex tensor operations characteristic of TNNs, enabling more efficient and reliable training processes. MANGO~\cite{pan2023reusing} accelerates large model training by establishing comprehensive linear correlations between all weights of pretrained and target models, rather than using only partial weight mapping as in earlier transfer schemes such as bert2BERT~\cite{chen2022bert2bert}.
As shown in Figure~\ref{fig:mango}, MANGO operates on the entire Transformer, including Multi-Head Self-Attention blocks, Feed-Forward Networks, and normalization layers, applying its mapping operator to correlate parameters between small and large models through TR-MPO.

% stable decomposition

Another feasible way to solve the training problem lies in developing a suitable initialization method for tensorial neural networks~(TNNs). Two widely adopted adaptive initialization methods in deep learning are Xavier~\cite{DBLP:journals/jmlr/GlorotB10} initialization and Kaiming~\cite{DBLP:conf/iccv/HeZRS15} initialization. Xavier initialization, proposed by Glorot and Bengio in 2010, regulates the variances of data flows between layers to prevent the vanishing gradient problem in deep networks. Similarly, Kaiming initialization, introduced by He~\textit{et al.} in 2015, was specifically designed for networks using ReLU activation functions.
However, these conventional initialization methods face two major challenges when applied to TNNs. First, they cannot accurately calculate the appropriate scales for TNNs due to their inability to account for the complex interactions occurring in tensor contractions. 

Second, the diversity of tensor formats (e.g., CP decomposition, Tucker decomposition, Tensor Train) makes it challenging to develop a universally applicable initialization method that fits all tensorial layers.
To address these limitations, Yu initialization~\cite{DBLP:conf/icml/0005SLW0X22} was proposed as a unified initialization paradigm. This method extends the principles of Xavier initialization while introducing adaptive mechanisms specifically designed for arbitrary Tensor-based Convolutional Neural Networks (TCNNs). The key innovation of Yu initialization lies in its systematic approach to handling tensor operations.
Specifically, Pan~\textit{et al.} developed a two-step process: First, they extract a backbone graph (BG) from a tensorial convolution hypergraph~\cite{DBLP:conf/nips/HayashiYSM19}, which captures the essential structure of tensor operations. Second, they encode an arbitrary TCNN into an adjacency matrix using this BG. Through this adjacency matrix representation, the method can directly calculate a suitable initial variance for any TCNN, taking into account its specific tensor structure and operations.
We illustrate three representative cases of applying these unified initializations in Fig.~\ref{fig:yuinit}. These examples demonstrate how the method adapts to different tensor formats and network architectures. Although Yu initialization was initially developed specifically for TCNNs, its applicability extends far beyond this original scope. The method has shown remarkable versatility and robustness across different settings, and it can be effectively applied to a wide range of various neural network architectures, including both convolutional and Transformer-based models.

\subsection{Rank Selection and Search\label{Sec:select}}
Prior studies~\cite{DBLP:conf/aaai/PanXWYWBX19,li2017bt,novikov2015tensorizing} focused on finding efficient TN formats (e.g., TTs and TRs) for compressing NNs and have achieved significant efficiency for their natural compact structures.
However, despite these remarkable successes, efficient algorithms for adjusting or selecting suitable ranks for a TN are lacking since rank selection is an NP-hard problem~\cite{hillar2013most}. As a result, many approaches~\cite{DBLP:conf/icml/YangKT17,li2017bt,DBLP:conf/cvpr/WangSEWA18,DBLP:journals/ijon/PanWX22} can only set values for all ranks manually, which severely affects the resulting models' training procedures.
Fortunately, the rank selection problem can still be optimized through heuristic strategies,
such as Bayesian optimization\cite{hawkins2021bayesian,sobolev2022pars}, reinforcement learning (RL)~\cite{cheng2020novel} and evolutionary algorithms (EAs)~\cite{li2021heuristic}. Here, we introduce some rank selection methods for TNNs.

DNNs utilize neural architecture search (NAS)~\cite{elsken2019neural} to search for the optimal network hyperparameters, achieving significant success.
As ranks can be treated as architecture hyperparameters, NAS is applicable to searching for optimal tensorial layers with better rank settings. Following this idea, the progressive searching TR network (PSTRN)~\cite{li2021heuristic} employs NAS with an EA to select suitable ranks for a TR network (TRN). In detail, the PSTRN employs a heuristic hypothesis for searching: ``when a shape-fixed TRN performs well, part or all of its rank elements are sensitive, and each of them tends to aggregate in a narrow region, which is called an interest region''.
Instructed by the interest region hypothesis, the PSTRN can reach the optimal point with a higher probability than a plain EA method.
The PSTRN consists of an evolutionary phase and a progressive phase. During the evolutionary phase, this method validates the ranks in the search space on benchmarks and picks the rank that yields the best performance. Then, in the progressive phase, the PSTRN samples new ranks around the previously picked rank and inserts them into a new search space. After several rounds, the heuristic EA can find a high-performance solution. With such an efficient design, the PSTRN successfully achieves better performance than manual setting, which demonstrates that its hypothesis is practical.
\rthreehl{Li \textit{et al.}~\cite{li20l,li2022tnls,li23ar} formulate structure search through evolutionary topology optimization, local-sampling-based permutation search, and alternating local enumeration with fewer evaluations, respectively. Zeng \textit{et al.}~\cite{zeng24b} further study language-model-assisted discovery of tensor-network structure-search algorithms. Together with the core-shape results in Section~\ref{Sec:pre}, these studies show that practical TNN design depends not only on rank values, but also on tensorization shape, mode order, and topology.}

In addition to NAS, some other efficient methods are also available for rank selection.
Zhao~\textit{et al.}~\cite{zhao2015bayesian} inferred a CP rank by implementing a reduction process on a large rank value via a variational Bayesian optimization procedure.
Hawkins and Zhang~\cite{hawkins2021bayesian} extended this CP procedure \cite{zhao2015bayesian} to TT-based TNNs and adopted the Stein variational gradient descent method, which combines the flexibility of the Markov chain Monte Carlo (MCMC) approach with the speed of variational Bayesian inference to construct a Bayesian optimization method. In pretrained networks, Kim~\textit{et al.}~\cite{DBLP:journals/corr/KimPYCYS15} and Gusak~\textit{et al.}~\cite{DBLP:conf/iccvw/GusakKPMBCO19} derived approximate ranks by employing Bayesian matrix factorization (BMF)~\cite{DBLP:conf/nips/NakajimaTSB12} to unfolding weight tensors. Konstantin~\textit{et al.}~\cite{sobolev2022pars} utilize a proxy-based Bayesian optimization approach to find the best combination of ranks for NN compression. 
Unlike Bayesian methods, Cheng~\textit{et al.}~\cite{cheng2020novel} treated the rank searching task as a game process whose search space was irregular, thus applying RL to find comparably suitable ranks for a trained CNN.
However, this algorithm is TD-dependent, which indicates that its performance may be influenced by the selected TD method. 

Yin~\textit{et al.}~\cite{DBLP:conf/cvpr/YinSL021} leveraged the alternating direction method of multipliers (ADMM) to gradually transfer the original weight to a low-rank representation (i.e., a TT). 
Farnaz~\textit{et al.}~\cite{sedighin2021adaptive} proposed an adaptive rank search framework for TR format in which TR ranks gradually increase in each iteration rather than being predetermined in advance. Wang \textit{et al.}~\cite{wang2026renormalization} apply renormalization-group flows to automate tensor-network structure search, obtaining more compact decompositions with substantially lower search cost.
\rthreehl{For TR models in particular, latent-space rank minimization has also been shown to be effective for controlling model complexity while preserving representation quality~\cite{yuan2019trlatent}. Taken together, these results reinforce a broader conclusion of this survey: rank, tensorization shape, and mode order should be treated as a coupled design problem rather than independent knobs.}

\rwone{We briefly discuss the strengths and limitations of the above rank-selection approaches. Manual/Grid Search is the simplest baseline. It is easy to implement but computationally expensive and prone to suboptimal results. Heuristic methods (e.g., PSTRN~\cite{li2021heuristic}) leverage domain-specific assumptions (e.g., interest region hypothesis) to guide the search, offering better efficiency than brute-force methods, but may miss global optima when the assumptions do not hold. Bayesian methods (e.g., CP-Bayes~\cite{zhao2015bayesian}, TT-Bayes~\cite{hawkins2021bayesian}, proxy-based Bayesian~\cite{sobolev2022pars}) provide principled uncertainty quantification over rank choices, enabling informed decisions with limited evaluations; however, they are computationally demanding and require careful prior specification. RL-based methods (e.g., TR-RL~\cite{cheng2020novel}) offer flexible exploration of irregular search spaces but are tensor-decomposition-dependent, meaning their performance may vary across different TD formats. ADMM-based methods~\cite{DBLP:conf/cvpr/YinSL021} enable gradual transfer from full-rank to low-rank representations, providing smooth optimization trajectories, but require careful tuning of penalty parameters. Adaptive methods~\cite{sedighin2021adaptive} automatically grow ranks during training, avoiding the need for pre-specification, but risk over-parameterization if growth criteria are not properly calibrated. In practice, the choice of rank-selection strategy should be guided by the specific requirements: Bayesian methods are recommended when evaluation budgets are limited and uncertainty quantification is important; heuristic methods are suitable for rapid prototyping; and adaptive methods are preferred when the optimal rank is unknown a priori.}

\subsection{Hardware Speedup}
\label{sec:hardware-implementation}
Accelerating the training and inference procedures of TNNs can benefit resource consumption and experimental adjustment, thereby achieving economic gains and green research. A direct and effective approach is to optimize the speed of tensor operations in TNNs to realize hardware acceleration.
As inferring TT-format TNNs inevitably results in enormous quantities of redundant calculations, the TIE scheme~\cite{DBLP:conf/isca/DengSQLWY19} was proposed to accelerate TT layers by splitting the working SRAM into numerous groups with a well-designed data selection mechanism.
Huang~\textit{et al.}~\cite{huang2017ltnn} designed a parallel computation scheme with higher I/O bandwidth, improving the speed of tensor contractions. Later, they proposed an LTNN~\cite{huang2017ltnn} to map TT-format TNNs into a 3D accelerator based on CMOS-RRAM, leading to significantly increased bandwidth via vertical I/O connections. As a result, they simultaneously attained high throughput and low power consumption for TNNs. Qu~\textit{et al.}~\cite{DBLP:journals/tcad/QuDWCLLLZX22} proposed a spatial 2D processing element (PE) array architecture and built a hardware TT engine consisting of off-chip DRAM.
% , a global buffer, a tensor multiplication unit (TMU), and an SVD core
 Kao~\textit{et al.}~\cite{DBLP:conf/mwscas/KaoHCY22} proposed an energy-efficient hardware accelerator for CP convolution with a mixing method that combines the Walsh-Hadamard transform and the discrete cosine transform. ETTE~\cite{gong2023ette} proposes a novel algorithm-hardware co-optimization framework for TT based TNN acceleration, featuring new tensor core construction, computation ordering mechanisms, and lookahead-style processing schemes, thereby achieving significant improvements in computational efficiency, memory consumption, and data movement compared to existing acceleration solutions across various DNN architectures and deployment settings.

Many more fascinating methods have been developed for the acceleration of generic tensor operations, which are correlated with TNNs.
For instance, Huang~\textit{et al.}~\cite{DBLP:conf/acssc/HuangDI0021} observed that the tensor matricization operation is usually resource-consuming since its DRAM access is built on a random reading address; thus, they proposed a tensor storage scheme with a sequential address design for better DRAM accessibility.
Both T2s-tensor~\cite{srivastava2019t2s} and Tensaurus~\cite{srivastava2020tensaurus} mainly focus on designing general computation kernels for dense and sparse tensor data. Xie~\textit{et al.}~\cite{xie2017optimized} and Liang~\textit{et al.}~\cite{liang2021fast} accelerated search procedures for obtaining an optimal sequence of tensor contractions. Xie~\textit{et al.}~\cite{xie2017optimized} solved the massive computational complexity problem of double-layer TN contraction in quantum analysis and mapped such a double-layer TN onto an intersected single-layer TN. Liang~\textit{et al.}~\cite{liang2021fast} implemented multithread optimization to improve the parallelism of contractions. \rthreehl{Meirom \textit{et al.}~\cite{meirom22a} optimize tensor-network contraction paths using reinforcement learning, complementing the broader RL-based tensor-operation design perspective of Fawzi \textit{et al.}~\cite{fawzi2022discovering}.}
\rthreehl{Xu \textit{et al.}~\cite{xu2026lego} develop a hierarchical optimizer that combines top-down decomposition with bottom-up fusion to search GPU kernel implementations while preserving local correctness and global fusion opportunities. Their results suggest that TNN acceleration can combine tensor-aware compiler representations with learned, structure-aware kernel synthesis rather than relying only on hand-designed contraction schedules.}
Although many TNNs have low calculation complexity levels in theory, realistic hardware deployments usually fall short of this objective due to their numerous permutation operations~\cite{li2024thc,DBLP:journals/ijon/PanWX22} and the absence of sufficient parallelism~\cite{huang2017ltnn}. Current hardware architectures are primarily designed for matrix operations, making them suboptimal for tensor operations that involve complex permutations and contractions. Frequent data movement between memory hierarchies creates substantial bottlenecks, especially for tensor transposition and reshaping, which require extensive reorganization but contribute little arithmetic computation. Promising directions include specialized tensor processing units~\cite{jouppi2023tpu}, tensor-aware memory hierarchies, and efficient contraction and permutation primitives that jointly optimize computation and data movement.
In the future, it is expected that more general hardware acceleration schemes based on tensor operations will be developed to implement TNNs with smaller storage and time consumption levels.

\rwtwo{\textbf{Remark.} The training strategies discussed in this section address three complementary aspects of TNN development: (1) training stability through mixed-precision approaches and unified initialization; (2) rank selection via heuristic, Bayesian, RL-based, and adaptive methods; and (3) hardware acceleration through specialized tensor operation engines. An important observation is that these three aspects are interrelated---for instance, the choice of rank directly affects both training stability and computational efficiency, and hardware-aware rank allocation strategies could potentially unify all three directions.}

\section{TNN Toolboxes\label{Sec:Tool}}

\rwone{In this section, we present a \emph{systematic taxonomy and critical analysis} of the existing toolbox ecosystem for TNNs rather than introducing new toolboxes. Specifically, our contributions are twofold: (1) we propose a unified categorization of existing toolboxes into three levels, namely basic tensor operations, deep model implementations, and quantum simulations, which helps readers quickly identify the tools most suitable for their needs; and (2) we analyze the limitations and gaps of current toolboxes, highlighting key directions for future improvement.}

In 1973, Pereyra and Scherer~\cite{pereyra1973efficient}, as pioneers in this field, developed a programming technique for basic tensor operations.
Modern computing frameworks support a broad range of basic tensor-operation toolboxes and higher-level TNN libraries for network compression and quantum-circuit simulation.
In this section,
toolboxes for TNNs are presented in three categories according to their design purposes: (1) toolboxes for basic tensor operations,  which contain important and fundamental operations (e.g., tensor contraction and permutation) in TNNs (Section~\ref{sec:basictensortoolbox}); (2) toolboxes for network compression are high-level TNN architecture toolboxes based on other basic operation tools (Section~\ref{sec:tensorneuraltoolbox}); and (3) toolboxes for quantum circuit simulation are software packages for the quantum circuit simulation or quantum machine learning processes that use TNs from a quantum perspective (Section~\ref{sec:quantumtoolbox}).

\subsection{ Toolboxes for Basic Tensor Operations}
\label{sec:basictensortoolbox}
Toolboxes for basic tensor operations aim to implement some specific TD algorithms.
Many basic tensor toolboxes based on different programming languages and backends have been designed for this purpose. For example,
the online stochastic framework for TD~(OSTD)~\cite{DBLP:conf/iccvw/SobralJJBZ15} and Tensor Toolbox~\cite{kolda2006matlab} were constructed for low-rank decomposition and implemented with MATLAB.
Regarding Python-based toolboxes,
TensorTools based on NumPy~\cite{DBLP:journals/cse/WaltCV11} implements CP only, while T3F~\cite{DBLP:journals/jmlr/NovikovIKFO20} was explicitly designed for TT decomposition on TensorFlow~\cite{DBLP:conf/osdi/AbadiBCCDDDGIIK16}. Similarly, based on TensorFlow, TensorD~\cite{DBLP:journals/ijon/HaoLYX18} supports CP and Tucker decomposition. Tntorch~\cite{tntorch} is a PyTorch-based library for tensor modeling in the CP, Tucker and TT formats. TorchMPS~\cite{torchmps}, TT-Toolbox~\cite{oseledets2016ttbox} and Scikit-TT~\cite{scikittt} are all powerful Python-based specific TT solvers that efficiently implement the DMRG algorithm. Tensorly is a powerful general TD library that supports many decomposition formats and various Python backends including CuPy, Pytorch, TensorFlow and MXNet. TensorNetwork~\cite{roberts2019tensornetwork} is a powerful general-purpose TN library that supports a variety of Python backends, including JAX, TensorFlow, PyTorch and NumPy.
\rwfour{Complementarily, CoLA~\cite{potapczynski2023cola} automates structure-aware numerical linear algebra by dispatching efficient routines for compositional matrix structures, such as Kronecker products, block diagonals, and sums, enabling accelerated solves, eigendecompositions, and trace estimation with automatic differentiation support in both JAX and PyTorch.}
In addition, some toolboxes based on C++ are also available.
TenDeC++\cite{tedc++} leverages a unique pointer technology called PointerDeformer in C++ to support the efficient computation of TD functions. ITensor~\cite{itensor} is an efficient and flexible C++ library for general TN calculations. Tensor4ML~\cite{chen2024tensor4ml} provides a comprehensive overview of tensor decomposition models, algorithms, and optimization techniques, along with Python implementations and datasets, serving as a bridge between theoretical foundations and practical applications in machine learning and data science.

\subsection{Toolboxes for Network Compression}
\label{sec:tensorneuraltoolbox}
Specific TNN toolboxes are used to assist with the development of tensorial layers.
Although some general tensor toolboxes such as Tensorly~\cite{DBLP:journals/jmlr/KossaifiPAP19} are powerful for TD processing and can use their TD operations to help initialize TNN modules to a certain extent, they still lack support for application programming interfaces (APIs) for building TNNs directly. Therefore, a TNN library (Tensorly-Torch) based on Tensorly was developed to build some tensor layers within any PyTorch network.
Pan \textit{et al.} also developed a powerful TNN library called TedNet~\cite{DBLP:journals/ijon/PanWX22}. TedNet can quickly set up TNN layers by directly calling the API.
In addition, TedNet supports the construction of TCNNs and TRNNs in single lines of code.

\subsection{Toolboxes for Quantum Simulation}
\label{sec:quantumtoolbox}

A number of quantum circuit simulation toolboxes have been designed.
For example, some TT toolboxes such as Scikit-TT and TorchMPS can partially simulate quantum circuits to some extent, although they were not specifically designed for quantum circuits simulation. In contrast, general TN toolboxes, e.g., TensorNetwork and ITensor, can simulate any quantum circuit. In addition, with optimized tensor contraction, TeD-Q~\cite{JDquantum}, a TN-enhanced open-source software framework for quantum machine learning, enables the simulation of large quantum circuits.
Furthermore, Yao~\cite{luo2020yao}, an extensible and efficient library for designing quantum algorithms, can provide support for dumping a quantum circuit into a TN. 

\ronehl{In addition, \emph{TensorNetworkQuantumSimulator.jl}~\cite{tindall2024tnqsjl}, a Julia toolbox for quantum circuit simulation with PEPS-style tensor-network backends, further broadens the software landscape beyond the Python/C++ ecosystems and is directly relevant to TN-based quantum simulation workflows. Although no practical hardware implementations of quantum TNNs are currently available, these classical quantum circuit simulations are still potentially useful and valuable for simulations.}

\textbf{Remark.}
Despite the success of current toolboxes, some areas for improvement remain.
\ronehl{(1) Existing basic tensor operation toolboxes are mostly implemented on top of high-level runtime stacks, which makes it difficult to fuse reshaping, permutation, and contraction into layout-aware kernels that fully exploit modern accelerators.} (2) Existing deep model implementation toolboxes for TNNs can only contain a limited number of predefined TNN structures and cannot allow users to design structures freely. (3) Existing quantum simulation toolboxes focus more on the simulation of quantum circuits using TNs and do not facilitate the processing of embedded quantum data via TNNs.

\rwtwo{\subsection{Benefits of Tensor Networks for Interpretability}
\label{sec:interpretability}
Tensor networks offer significant potential for interpretability, a dimension largely underexplored in the TNN literature. Their structured, low-rank form enables the extraction of quantities such as mutual information and entanglement entropy between input features. Puljak \textit{et al.}~\cite{puljak2025tensor} use MPS representations for transparent classification and anomaly detection. Hamreras \textit{et al.}~\cite{hamreras2025tensorization} analyze low-rank factorizations of neural weights and activations as tools for both compression and interpretability. Pareja Monturiol \textit{et al.}~\cite{pareja2025tensorization} construct MPS representations of trained black-box networks through sketching and cross interpolation, enabling structural analysis while obfuscating parameter-level information associated with the training distribution. Hamdi and Angryk~\cite{hamdi2019interpretable} further show that tensor factorization can expose meaningful latent structure in graph data.}
\rthreehl{Heidari and Rabusseau~\cite{heidari2026tnshap} construct a graph-aligned tensor-network surrogate for a black-box predictor and deterministically recover first- and higher-order Shapley interactions without repeated model queries. Their formulation uses TN topology directly as an interpretable surrogate structure rather than only as a compressed parameterization.}

\section{Discussion and Future Perspectives}
\label{sec:discussion}

\aehl{The value of a survey lies not only in organizing existing methods, but also in identifying the design principles that recur across otherwise different applications. The literature reviewed in this article indicates that the central questions in TNN research concern how tensors are formed, how network topology affects expressiveness and computation, how learning algorithms exploit multilinear structure, and when tensorization provides an appropriate inductive bias. We therefore organize the discussion around five cross-cutting themes, followed by a perspective on the exchange of ideas between tensor-network learning and quantum physics.}
\rthreehl{These themes should be interpreted together with several unresolved limitations. Tensorized models remain nonconvex and can be sensitive to initialization, rank choice, mode ordering, and numerical conditioning. Theoretical parameter reductions do not necessarily translate into wall-clock or energy savings because permutation, contraction scheduling, and memory traffic can dominate execution. Robustness to perturbations, calibration, cross-domain transfer, and reproducibility across common benchmarks also remain less established than compression results. Accordingly, the following perspectives emphasize testable design criteria and matched-budget evaluation rather than treating tensorization as an automatic advantage.}

\subsection{Tensorization Design Beyond Rank}
\label{sec:insight-design}

\rthreehl{Tensorization design is a multi-dimensional model-selection problem. Rank is an important capacity parameter, but it does not uniquely determine either approximation quality or computational cost. The mapping from a dense tensor to a collection of modes determines which variables interact locally, the dimensions of the cores determine where representational capacity is allocated, and the network topology determines which correlations can be expressed through short contraction paths. Ghadiri \textit{et al.}~\cite{ghadiri2023core_shapes} show that core shape alone can materially change decomposition efficiency, while Li \textit{et al.}~\cite{li2022tnls} demonstrate that mode permutation can alter the quality of a fixed tensor-network structure. These results imply that reporting only the decomposition family and rank is insufficient for reproducible comparison.}

\rthreehl{Topology selection further couples statistical and computational considerations. Li \textit{et al.}~\cite{li20l,li23ar} search over tensor-network structures using evolutionary and local-enumeration strategies, whereas Wang \textit{et al.}~\cite{wang2026renormalization} use renormalization-group flows to guide structure search. Such methods move beyond the assumption that TT, TR, Tucker, or PEPS should be selected in advance. However, the resulting search spaces are highly nonuniform: two structures with the same number of parameters may have very different contraction widths, memory requirements, and optimization landscapes. Structure search should therefore optimize a vector of objectives rather than parameter count alone.}

\rthreehl{A useful evaluation protocol should report the tensorization map, mode ordering, core dimensions, rank profile, topology, and contraction strategy together with accuracy, peak memory, and measured latency. Comparisons should control at least one of parameter count, arithmetic cost, or memory traffic, because matching rank does not imply matching computational complexity. An important research direction is to develop differentiable or bilevel procedures that jointly select tensorization, topology, ranks, and contraction order. Such procedures would turn architectural choices that are currently treated as implementation details into explicit, testable components of the learning problem.}

\subsection{Expressiveness Versus Contractibility}
\label{sec:insight-contraction}

\rthreehl{The expressive power of a tensor network is inseparable from the cost of contracting it. Chain and tree structures, including TT/MPS and HT, remain common because they admit efficient exact contraction and stable local manipulation. Their low treewidth, however, restricts the pathways through which distant variables interact. Loopy structures such as PEPS can encode richer spatial and multiway correlations, but exact contraction is computationally intractable in the general case~\cite{biamonte2017tensor,schuch2007computational}. Consequently, a topology that is more expressive in representation space may be less useful as a trainable neural component if its forward and backward contractions cannot be evaluated within practical resource limits.}

\rthreehl{This trade-off is not binary. Gray and Chan~\cite{slattery2024hyperoptimized} show that contraction for arbitrary geometries can be improved substantially through hyperoptimization, while Alkabetz and Arad~\cite{alkabetz2021tensor} connect belief propagation with tensor-network contraction. Tindall \textit{et al.}~\cite{tindall2024efficient} further demonstrate that message-passing approximations can support large loopy simulations. These studies indicate that the effective model class is determined jointly by topology and contraction algorithm. Approximation error, contraction-path quality, numerical stability, and hardware parallelism therefore belong to the architectural analysis rather than to a separate implementation stage.}
\rthreehl{Future TNN studies should evaluate expressiveness under a fixed contraction budget. Relevant quantities include maximum intermediate tensor size, contraction width, peak memory, communication volume, and the error introduced by approximate contraction. Hybrid structures may offer a productive compromise: tree-like backbones can preserve tractability, while a limited number of additional edges can encode selected long-range dependencies. Another open problem is to connect local contraction error with task-level prediction error and gradient bias. Such theory would clarify when approximate contraction preserves the benefit of a richer topology and when it merely introduces an uncontrolled source of optimization noise.}

\subsection{The Algorithmic Shift in TNN Research}
\label{sec:insight-algorithms}

\rthreehl{TNN research is increasingly defined by algorithms for constructing, searching, contracting, and learning tensor networks rather than by the insertion of a fixed decomposition into another neural architecture. Bharadwaj \textit{et al.}~\cite{bharadwaj2024leverage_tt} reduce the cost of TT construction through leverage-score sampling, and Tao \textit{et al.}~\cite{tao2023upmtd} formulate probabilistic tensor decomposition as an undirected graphical model. Hameed and Rabusseau~\cite{hameed2025efficientptn} develop an automatic-differentiation-compatible procedure for probabilistic tensor networks; because this work remains a preprint, its empirical conclusions should be interpreted together with the peer-reviewed probabilistic decomposition literature. These examples show that sampling, uncertainty estimation, and numerical linear algebra are becoming first-class parts of TNN methodology.}

\rthreehl{A parallel algorithmic development concerns contraction and execution. Meirom \textit{et al.}~\cite{meirom22a} learn contraction strategies with reinforcement learning, complementing the broader demonstration by Fawzi \textit{et al.}~\cite{fawzi2022discovering} that search can discover efficient multilinear algorithms. At the computation-graph level, Xu \textit{et al.}~\cite{xu2026lego} combine hierarchical decomposition and fusion to search tensor-oriented GPU kernels. These contributions blur the conventional boundary between model architecture, compiler optimization, and hardware mapping: changing the contraction schedule can alter memory consumption by orders of magnitude without changing the mathematical tensor network.}

\rthreehl{The next step is to optimize these layers jointly. A complete pipeline would select the tensorization and topology, estimate ranks, choose exact or approximate contraction, and generate hardware-aware kernels under a shared resource objective. Such integration requires reproducible benchmarks that separate algorithmic gains from implementation effects and report wall-clock time, memory traffic, numerical error, and energy consumption in addition to parameter count. Theoretical analysis is also needed for convergence when the model structure or contraction approximation changes during training. Establishing these foundations would position TNNs as a general framework for structured computation rather than a collection of compression techniques.}

\subsection{Tensorization as an Inductive Bias}
\label{sec:insight-bias}

\rthreehl{Tensorization should be interpreted as an inductive bias, not merely as a mechanism for reducing parameters. A decomposition constrains which interactions are represented directly, which correlations must be transmitted through intermediate cores, and how parameters are shared across modes. When these constraints align with the structure of the task, a tensorized model may generalize better than a dense model even at a comparable parameter budget. Wang \textit{et al.}~\cite{wang2023transformed_lowrank} provide theoretical and empirical evidence that transformed low-rank parameterizations can improve robust generalization. This perspective explains why compression ratio alone is an incomplete measure of success: two equally compact models may encode very different assumptions about the data.}

\rthreehl{Several application settings illustrate this principle. Xie \textit{et al.}~\cite{xie2026sweet} use shared Tucker-structured templates as transferable priors across vision-model scales. Wang \textit{et al.}~\cite{wang2026tginee} encode correlations across multiple graphs through shared latent tensor factors rather than treating graph layers independently. Klein \textit{et al.}~\cite{klein2026tuckerattention} and Zhang \textit{et al.}~\cite{zhang2025tensornew} impose structured low-rank interactions on attention and its key-value representations. In each case, the tensor format specifies which forms of sharing and interaction are preferred before optimization begins.}

\rthreehl{Matrix factorization provides a necessary baseline hierarchy for evaluating this bias. Truncated SVD and related matrix-low-rank methods operate on a selected unfolding and can be preferable when a weight is intrinsically two-dimensional. TNNs generalize this idea by retaining multiple modes and distributing capacity across cores and topology, but the advantage is not automatic: reshaping can introduce artificial correlations, and contraction overhead can offset parameter savings. Evaluations should therefore compare dense, matrix-low-rank, and tensor-network models under matched parameter, compute, and data budgets, while also reporting out-of-distribution performance, robustness, and calibration. Such comparisons help distinguish task-aligned structural generalization from gains caused only by regularization or reduced model size.}

\subsection{Unified Learning and Scientific Machine Learning}
\label{sec:insight-unified}

\rthreehl{Tensor networks provide a common representation language for supervised prediction, unsupervised density estimation, probabilistic inference, and operator learning. Zeng \textit{et al.}~\cite{zeng2026mtnl} connect supervised and unsupervised tensor-network learning through mixed structures, structure-search priors, and lightweight uncertainty quantification. This perspective is important because many applications require more than a point prediction: they must also represent latent structure, generate samples, or quantify uncertainty. A unified framework can share cores or intermediate representations across these objectives while preserving the explicit multilinear structure of the model.}

\rthreehl{Scientific machine learning places additional demands on such a framework. Models must often respect separability, locality, conservation structure, or symmetries while operating on high-dimensional fields. Zhou \textit{et al.}~\cite{zhou2026tuckerfno} decompose a high-dimensional Fourier neural operator into lower-dimensional operators and establish a universal-approximation guarantee. Yang \textit{et al.}~\cite{yang2026kgn} show that Kronecker structure can also be used to generate parameter-efficient connectivity patterns rather than merely factorize existing weights. These examples indicate that tensorization can serve simultaneously as a computational device and as a mechanism for encoding scientific structure.}

\rthreehl{Important open problems concern the preservation of physical and statistical properties after factorization. Approximation guarantees should be accompanied by analyses of stability, discretization transfer, uncertainty calibration, and conservation error. Benchmarks should compare tensorized operators not only by prediction accuracy, but also by memory, resolution generalization, rollout stability, and sensitivity to rank and mode ordering. Another promising direction is to combine probabilistic tensor networks with neural operators so that uncertainty is represented in the same structured space as the solution field. Such developments would connect TNN research with inverse problems, surrogate modeling, and data assimilation without reducing scientific machine learning to another application-specific compression task.}

\subsection{Quantum Physics Perspectives and Physics-Inspired Algorithms}
\label{sec:quantum-perspectives}

\rthreehl{Tensor networks originated as structured representations of quantum many-body states, and this origin continues to provide both mathematical tools and algorithmic intuition for TNNs. In quantum simulation, tensor networks capture entanglement and high-order correlations without explicitly materializing the exponentially large state tensor~\cite{cai2018approximating}. Neural and tensor-network models therefore offer complementary mechanisms: neural components provide flexible nonlinear parameterization, while tensor structure provides controlled correlation pathways and access to physically meaningful quantities. The challenge is to combine these advantages without discarding the tractability and interpretability that motivate tensor networks in the first place.}

\rthreehl{Loconte \textit{et al.}~\cite{loconte2026circuits} place tensor factorizations and tractable circuit representations in a generalized hierarchical framework, clarifying how modular components can be composed while preserving efficient inference. Physics-originated optimization algorithms offer another connection. The density matrix renormalization group optimizes an MPS locally through sweeping updates, providing a direct analogue of alternating core optimization in TNN training~\cite{ganahl2023density}. Message-passing and approximate-contraction methods similarly suggest ways to train models with loopy structures that would otherwise be computationally inaccessible~\cite{alkabetz2021tensor,tindall2024efficient}.}
\rthreehl{MERA provides a particularly instructive example because its multiscale representation combines tensor contractions with disentanglers and isometries~\cite{cincio2008multiscale,batselier2021meracle,reyes2021multi}. These components encode scale-dependent correlations that are not naturally represented by a simple chain or ring. Translating such structures into practical TNNs requires differentiable constraints, stable joint optimization, and efficient contraction on classical hardware. Evaluation should determine whether physics-inspired components improve representation or optimization beyond what can be explained by parameter count. More broadly, future work should identify which physical concepts, including entanglement measures, renormalization, gauge freedom, and locality, remain meaningful when tensor networks are embedded in neural systems. This distinction is necessary to turn physical analogy into testable algorithmic design.}

\section{Conclusion}
TNs and NNs provide complementary perspectives on high-dimensional learning: TNs offer explicit multilinear structure and controllable complexity, whereas NNs provide flexible nonlinear representations and scalable data-driven optimization. This survey has examined their theoretical connections, methodological integration, and principal applications within TNNs. In particular, tensor convolution, convolutional tensors, tensorized parameterizations, and tensor-network diagrams provide a common mathematical language for analyzing neural operations and clarifying how network structure affects representation, computation, and optimization.

The practical value of this integration extends beyond parameter compression. For data representation, TNNs preserve higher-order dependencies in multimodal, multi-view, multi-source, and multi-task settings without requiring the underlying structure to be flattened prematurely. For model design, tensor decompositions replace dense transformations with structured multilinear mappings, reducing parameter count, memory, and computational cost. More importantly, tensorization acts as an inductive bias: mode organization, core shape, rank allocation, and network topology determine which interactions are represented efficiently and how information is shared across inputs, tasks, layers, or modalities. Consequently, a TNN's effectiveness cannot be judged by compression ratio alone, but also by predictive performance, robustness, contraction cost, memory traffic, and compatibility with the target problem's structure.
TNNs also connect machine learning with quantum many-body modeling and physics-inspired computation. Concepts such as entanglement, locality, renormalization, message passing, and structured contraction provide useful principles for understanding and designing tensorized learning systems. Conversely, differentiable optimization and neural parameterization can extend the range of tensor-network models that can be trained from data. This relationship should nevertheless be treated as a source of testable computational and representational principles rather than as a purely conceptual analogy. Establishing when physics-inspired structures provide measurable advantages over conventional tensor decompositions remains an important research question.

\rthreehl{Future progress will require treating tensorization, topology selection, rank adaptation, contraction, and hardware execution as a coupled design problem rather than optimizing them in isolation. Recent advances in tensorized attention, mixture-of-experts models, graph learning, neural operators, and multiscale structures such as MERA illustrate the breadth of this design space, but their practical value should be established through matched-budget and hardware-aware evaluation. Parameter and arithmetic reductions should therefore be assessed together with predictive performance, robustness, memory traffic, latency, and energy consumption. Under these criteria, tensor structure can evolve from a post hoc compression device into a principled architectural choice for efficient, interpretable, and structurally grounded learning, while also clarifying when multilinear structure provides measurable advantages in neural computation.}

% \appendices

% % you can choose not to have a title for an appendix
% % if you want by leaving the argument blank
% \section{}

% \newpage

% \vspace{-3 mm}
% % use section* for acknowledgment
\ifCLASSOPTIONcompsoc
  \section*{Acknowledgments}
\else
  \section*{Acknowledgment}
\fi
This paper was partially supported by National Key Research and Development Program of China (No. 2018AAA0100204). G. Li was supported by National Natural Science Foundation of China (Grant No. 62402206). This paper was also partially supported by the Hong Kong Institute of AI for Science, City University
of Hong Kong.

\ifCLASSOPTIONcaptionsoff
  \newpage
\fi



%\bibliographystyle{IEEEtran}

% \clearpage

\bibliographystyle{IEEEtran}
\bibliography{nntn}

% \newpage

% %\vspace{-10 mm}
\begin{IEEEbiographynophoto}{Maolin Wang} is a Research Assistant Professor at the Hong Kong Institute of AI for Science (HKAI-Sci), City University of Hong Kong. He received his Ph.D. from the Department of Data Science, City University of Hong Kong, under the supervision of Prof. Ruocheng Guo, Prof. Xiangyu Zhao, and Prof. Junhui Wang, and earlier received his master's and bachelor's degrees from the University of Electronic Science and Technology of China (UESTC) under the guidance of Prof. Zenglin Xu. His research focuses on graph learning, model compression, tensor/matrix decomposition, and large language models (LLMs). His Ph.D. dissertation, Tensor Criterion Based Neural Networks, was named Runner-Up for the 2026 INNS Best Doctoral Dissertation Award. He also received the Best Paper Award Runner-Up at KDD 2025 and the Best Paper Award at the 1st TensorKDD Workshop, co-located with KDD 2026, along with the College of Computing Research Excellence Award at City University of Hong Kong and Best Reviewer at the KDD 2026 ADS Track. 
\end{IEEEbiographynophoto}

%\vspace{-10 mm}
\begin{IEEEbiographynophoto}{Yu Pan}
received a Ph.D. degree from the School of Computer Science and Technology, Harbin Institute of Technology, Shenzhen. Prior to that, he received his M.E. degree from the School of Computer Science and Engineering, University of Electronic Science and Technology of China. He is currently a Research Scientist at Huawei Noah's Ark Lab in Shenzhen, China. His research interests mainly include TD techniques and DNNs.
\end{IEEEbiographynophoto}
%\vspace{-10 mm}

\begin{IEEEbiographynophoto}{Zenglin Xu} is a Distinguished Professor at Fudan University and an AI Scientist at the Shanghai Academy of AI for Science, where his work centers on trustworthy machine learning and AI for Science. Prior to joining Fudan University, he conducted research and held faculty positions at leading institutions across North America, Europe, and Asia, including Michigan State University, the Cluster of Excellence at Saarland University and the Max Planck Institute for Informatics, Purdue University, the University of Electronic Science and Technology of China, and Harbin Institute of Technology, Shenzhen. He earned his Ph.D. in Computer Science and Engineering from The Chinese University of Hong Kong. Dr. Xu’s research has been recognized with the APNNS Young Researcher Award (2016), an AAAI Outstanding Student Paper Honorable Mention (2015), the ACML Best Student Paper Runner-Up Award (2016), and a Best Paper Finalist distinction at ICONIP 2023. Beyond his research, he provides extensive leadership to the global AI community as Vice President for Education of the International Neural Network Society (INNS), Editor-in-Chief of Academia AI and Applications, Senior Action Editor of Neural Networks, and a recurring Area Chair for premier conferences such as NeurIPS, ICML, ICLR, and ACL.
\end{IEEEbiographynophoto}
%\vspace{-10mm}

\begin{IEEEbiographynophoto}{Guangxi Li}
is currently an Assistant Researcher at the Quantum Science Center of Guangdong-Hong Kong-Macao Greater Bay Area. He received his Ph.D. degree from the Centre for Quantum Software and Information, University of Technology Sydney, in 2023. Prior to that, he earned his M.S. degree in 2018 and his B.S. degree in 2015, both in computer science and engineering from the University of Electronic Science and Technology of China. His research interests include quantum machine learning, QNNs, TD, TN and recommendation systems.
\end{IEEEbiographynophoto}

\begin{IEEEbiographynophoto}{Xiangli Yang}
received a Ph.D. degree at the School of Computer Science and Engineering, University of Electronic Science and Technology of China, Chengdu, China. She is currently a lecturer employed by the Information Engineering University in Zhengzhou, China. Her research is in Semi-supervised learning, category discovery, and data mining. Her research interests are semi-supervised learning, data mining, and machine learning.
\end{IEEEbiographynophoto}
%\vspace{-10 mm}

\begin{IEEEbiographynophoto}{Danilo P. Mandic} (Fellow, IEEE) is a Professor of Machine Intelligence with Imperial College London, UK, and has been working in the areas of machine intelligence, statistical signal processing, big data, data analytics on graphs,  bioengineering, and financial modelling. He is a Fellow of the IEEE and a past President of the International Neural Network Society (INNS). Dr Mandic is a Director of the Laboratory for Artificial Intelligence and Data Analytics (AIDA-LAb, www.aidalab.co.uk), and has more than 600 publications in international journals and conferences. Dr Mandic is a 2018 recipient of the Dennis Gabor Award for "Outstanding Achievements in Neural Engineering", given by the International Neural Networks Society. He is the 2023 Winner of The Prize Paper Award, given by the IEEE Engineering in Medicine and Biology Society for his Smart Helmet article, the 2018 winner of the Best Paper Award in IEEE Signal Processing Magazine for his article on tensor decompositions for signal processing applications, and the 2021 winner of the Outstanding Paper Award in the International Conference on Acoustics, Speech and Signal Processing (ICASSP) series of conferences. He has given more than 80 Keynote and Tutorial lectures in international conferences and was appointed by the World University Service (WUS), as a Visiting Lecturer within the Brain Gain Program (BGP), in 2015. Danilo is currently serving as a Distinguished Lecturer for the IEEE Computational Society and a Distinguished Lecturer for the IEEE Signal Processing Society. Dr Mandic is a 2014 recipient of President Award for Excellence in Postgraduate Supervision at Imperial College and holds eight patents.
\end{IEEEbiographynophoto}
%\vspace{-10 mm}
\begin{IEEEbiographynophoto}{Andrzej Cichocki} (Fellow, IEEE) received the M.Sc. (hons.), Ph.D. and Dr.Sc. (Habilitation) degrees, in electrical engineering from Warsaw University of Technology, Warsaw, Poland, in 1972,
1976, and 1982, respectively. He spent several
years at University Erlangen (Germany) as an
Alexander von-Humboldt Research Fellow and Guest
Professor. He was a Senior Team Leader and Head
of the laboratory for Advanced Brain Signal Processing, at RIKEN Brain Science Institute (Japan) and Visiting Professor professor with Skolkovo Institute of Science and Technology (Skoltech) and an Adjunct Professor with Hangzhou
Dianzi University. Currently he is working as professor in Systems Research Institute, Polish Academy of Science. He has authored more than 500 technical journal papers
and 6 monographs in English (two of them translated to Chinese). His current research interests include multi-way blind source separation, tensor decomposition, tensor networks, deep learning, human robot interactions and brain-computer interface. Dr. Cichocki served as an Associate Editor of,
IEEE Transactions on Signal Processing, IEEE Transactions on Neural Networks and Learning Systems, IEEE Transactions on Cybernetics, Journal of Neuroscience Methods. 
\end{IEEEbiographynophoto}

% ======================================================================
% Supplementary Material
% (merged from the former supplementary_material.tex so that the arXiv
%  version compiles as a single document)
% ======================================================================
\clearpage
\setcounter{section}{0}
\renewcommand{\thesection}{S\arabic{section}}
\renewcommand{\theHsection}{S\arabic{section}}
\setcounter{equation}{0}
\renewcommand{\theequation}{S\arabic{equation}}
\renewcommand{\theHequation}{S\arabic{equation}}
\setcounter{table}{0}
\renewcommand{\thetable}{S\arabic{table}}
\renewcommand{\theHtable}{S\arabic{table}}
\setcounter{figure}{0}
\renewcommand{\thefigure}{S\arabic{figure}}
\renewcommand{\theHfigure}{S\arabic{figure}}
\twocolumn[{\centering{\LARGE Supplementary Material for\\[0.3em]
Tensor Networks Meet Neural Networks:\\[0.3em] A Survey and Future Perspectives\par}\vspace{2em}}]

\section*{Detailed Tensor Decomposition Formulations}

The main article retains representative formulas for CP, Tucker, and tensor train decompositions and summarizes the remaining formats through their topology, ranks, and contraction properties. This supplement records detailed formulations omitted from the main article.

\section*{Background, Notation, and Scope}

An order-$N$ tensor $\mathcal{X}\in\mathbb{R}^{I_1\times\cdots\times I_N}$ contains $\prod_{n=1}^{N}I_n$ entries when represented densely. Tensor decomposition replaces this dense array with a collection of lower-order factors whose shared indices are summed during contraction~\cite{kolda2009tensor,cichocki2015tensor}. An index associated with an original data dimension is often called a \emph{physical index}, whereas an internal index joining two factors is called a \emph{bond} or \emph{virtual index}. The dimension of a bond is a tensor-network rank. These ranks determine how much information can pass between different groups of physical modes and therefore control both approximation capacity and resource use.

The topology of the factor graph specifies which cores share bonds. A chain gives the tensor train or matrix product state, closing the chain gives a tensor ring, recursively grouping modes gives a hierarchical Tucker tree, and placing cores on a two-dimensional lattice gives PEPS. Block-term models instead represent the tensor as a sum of several structured components. Two decompositions with similar parameter counts can consequently have different contraction widths, parallelism, numerical conditioning, and sensitivity to mode ordering. Parameter count should therefore be interpreted as one resource measure rather than as a complete prediction of execution cost.

In a conventional numerical decomposition, the factors are fitted to approximate a given dense tensor. In a TNN, the same factors may instead be initialized directly and optimized end to end with the task loss. The latter setting avoids materializing a pretrained dense weight but changes the optimization problem: scale and gauge freedoms can produce flat directions, small ranks can restrict task-relevant interactions, and a mathematically compact factorization can require several sequential kernels and intermediate activations. The formulas below describe the represented tensors, while the accompanying discussion highlights the additional choices that arise when the formats are used as trainable neural components.

Throughout this supplement, mode sizes are denoted by $I_n$ or by paired input and output sizes $(I_n,J_n)$. Core or bond dimensions are denoted by $R_n$. The notation $\times_n$ denotes a mode product, and repeated virtual indices are summed. Complexity statements refer to stored scalar parameters unless contraction or execution cost is stated explicitly. Biases, normalization parameters, and nonlinear activations are not part of the decomposition formulas.

\section{Block-Term Tucker and General Block-Term Decompositions}

Block-Term Tucker (BTT) represents a tensor as a sum of Tucker blocks with a shared multilinear-rank specification~\cite{de2008decompositions1}. For a fourth-order tensor $\mathcal{X}\in\mathbb{R}^{I_1\times I_2\times I_3\times I_4}$, a BTT model with $R_C$ blocks and uniform Tucker rank $R_T$ can be written elementwise as
\begin{align}
\mathcal{X}_{i_1,i_2,i_3,i_4}
\approx
\sum_{c=1}^{R_C}
\sum_{r_1,r_2,r_3,r_4=1}^{R_T}
\mathcal{G}^{(c)}_{r_1,r_2,r_3,r_4}
\prod_{n=1}^{4}
\mathbf{A}^{(n,c)}_{i_n,r_n},
\label{eq:supp:btt}
\end{align}
where $\mathcal{G}^{(c)}\in\mathbb{R}^{R_T\times R_T\times R_T\times R_T}$ and $\mathbf{A}^{(n,c)}\in\mathbb{R}^{I_n\times R_T}$. Setting $R_T=1$ gives a CP-type sum of rank-one terms, while setting $R_C=1$ gives a Tucker model.

The general BTD permits every block to use a different multilinear-rank profile~\cite{de2000multilinear}:
\begin{align}
\mathcal{X}
\approx
\sum_{c=1}^{R_C}
\mathcal{G}_c
\times_1\mathbf{A}_c^{(1)}
\times_2\mathbf{A}_c^{(2)}
\cdots
\times_N\mathbf{A}_c^{(N)},
\label{eq:supp:btd}
\end{align}
where $\mathcal{G}_c\in\mathbb{R}^{R_{1,c}\times\cdots\times R_{N,c}}$ and $\mathbf{A}_c^{(n)}\in\mathbb{R}^{I_n\times R_{n,c}}$. Thus, each block can allocate a different amount of capacity to each mode.

BTT and BTD interpolate between globally shared low-rank structure and a mixture of specialized components. A single Tucker block uses one multilinear subspace in each mode. Adding blocks allows different components to occupy different combinations of subspaces, which can represent heterogeneous patterns more efficiently than increasing every Tucker rank uniformly. CP appears as the special case in which every block core is a scalar and all multilinear ranks equal one. This relationship is useful in neural layers because the number of blocks and the within-block ranks provide two distinct capacity controls.

For the fourth-order uniform BTT model in Eq.~\eqref{eq:supp:btt}, the stored parameter count is
\begin{equation}
P_{\mathrm{BTT}}
=
R_C\left(R_T^4+R_T\sum_{n=1}^{4}I_n\right).
\end{equation}
For general BTD, it becomes
\begin{equation}
P_{\mathrm{BTD}}
=
\sum_{c=1}^{R_C}
\left(
\prod_{n=1}^{N}R_{n,c}
+
\sum_{n=1}^{N}I_nR_{n,c}
\right).
\end{equation}
These counts can be far below $\prod_n I_n$, but the core term can grow quickly with order and multilinear rank. A BTD layer is therefore not automatically cheaper than CP or TT at the same nominal rank. Contraction order also determines activation memory.

When BTD is trained end to end, different blocks may become redundant or may learn highly correlated subspaces. Block permutation is immaterial to the represented tensor, and reciprocal rescaling of a core mode and its factor matrix leaves the output unchanged. Initialization, normalization, and regularization can therefore matter as much as the raw parameter count. Practical model selection should report both $R_C$ and the complete rank profile of every block rather than using a single ambiguous ``BTD rank.''

In TNN applications, block-term layers are most attractive when one Tucker core is too restrictive but a dense core is unaffordable. They can model multiple interaction regimes, share structure across tasks, or provide a controllable path between CP-like sparsity and Tucker-like coupling. Their main limitations are the enlarged hyperparameter space, possible block redundancy, and uneven support for optimized implementations across current learning frameworks.

\section{Matrix Product Operator}

An MPO tensorizes both the input and output modes of a linear operator~\cite{pirvu2010matrix}. For paired local input and output modes, its elementwise representation is
\begin{align}
\mathcal{X}_{i_1,j_1,\ldots,i_N,j_N}
&\approx
\sum_{r_1=1}^{R_1}\cdots\sum_{r_{N-1}=1}^{R_{N-1}}
\notag\\
&\quad
\mathcal{G}^{(1)}_{1,i_1,j_1,r_1}
\mathcal{G}^{(2)}_{r_1,i_2,j_2,r_2}
\cdots
\mathcal{G}^{(N)}_{r_{N-1},i_N,j_N,1},
\label{eq:supp:mpo}
\end{align}
The dimensions of each MPO core are
\begin{equation}
\mathcal{G}^{(n)}
\in
\mathbb{R}^{R_{n-1}\times I_n\times J_n\times R_n},
\qquad R_0=R_N=1.
\end{equation}
The paired modes $(I_n,J_n)$ distinguish an MPO core from a third-order TT/MPS core.

An MPO can be understood as a tensor train applied to a matrix after the row and column indices have each been factorized. If a dense linear map has input dimension $I=\prod_n I_n$ and output dimension $J=\prod_n J_n$, the dense representation stores $IJ$ values. The corresponding MPO stores
\begin{equation}
P_{\mathrm{MPO}}
=
\sum_{n=1}^{N}R_{n-1}I_nJ_nR_n.
\end{equation}
With bounded local dimensions and ranks, this count grows linearly in the number of cores rather than exponentially in $N$~\cite{oseledets2011tensor}. This scaling is the principal reason that MPO and TT-matrix layers are used to replace large fully connected transformations.

The saving depends on how the original input and output dimensions are split and paired. Modes representing strongly related features should ideally be placed close in the chain so that their interaction does not have to pass through many small bonds. A poor factorization of $I$ and $J$, or an unfavorable permutation of the paired modes, can require much larger ranks for the same approximation quality. The tensorization map is consequently part of the model definition, not merely a reshape operation.

For a neural layer, an MPO contraction can be evaluated without reconstructing the dense matrix. The input is reshaped into local modes and contracted sequentially with the cores. The best contraction order depends on batch size, mode sizes, rank profile, and backend. A parameter-efficient MPO may still underutilize a GPU when it produces many small contractions or repeated permutations. Conversely, fusing contractions and choosing balanced local dimensions can make the structured layer substantially more practical.

Like other chain tensor networks, an MPO has gauge freedom. An invertible matrix can be inserted on a bond and its inverse absorbed into the adjacent core without changing the represented operator. Orthogonalization or canonicalization can reduce this ambiguity and improve numerical stability, particularly for alternating optimization or rank truncation~\cite{schollwock2011density}. End-to-end gradient training does not require an explicit canonical form, but monitoring core norms and avoiding extreme scale imbalance remains useful.

\section{Tensor Ring}

Tensor Ring (TR) closes the open boundary of a tensor train and contracts the cores cyclically~\cite{zhao2016tensor}:
\begin{align}
&\mathcal{X}_{i_1,\ldots,i_N}
\approx
\sum_{r_0=1}^{R_0}\sum_{r_1=1}^{R_1}\cdots\sum_{r_{N-1}=1}^{R_{N-1}}
\nonumber\\
&\qquad
\mathcal{G}^{(1)}_{r_0,i_1,r_1}
\mathcal{G}^{(2)}_{r_1,i_2,r_2}
\cdots
\mathcal{G}^{(N)}_{r_{N-1},i_N,r_0},
\label{eq:supp:tr}
\end{align}
where $\mathcal{G}^{(n)}\in\mathbb{R}^{R_{n-1}\times I_n\times R_n}$ and the cyclic boundary condition is $R_0=R_N$. Unlike TT, no core is forced to terminate in a rank-one boundary bond.

For fixed physical indices, each TR core becomes a matrix, and the ring admits the following trace form used in tensor-network analysis:
\begin{equation}
\mathcal{X}_{i_1,\ldots,i_N}
\approx
\operatorname{Tr}\!\left(
\mathbf{G}^{(1)}(i_1)
\mathbf{G}^{(2)}(i_2)\cdots
\mathbf{G}^{(N)}(i_N)
\right).
\end{equation}
The trace closes the final bond and makes the representation invariant to cyclic shifts of the core sequence, provided that the associated physical modes are shifted with it. This cyclic symmetry removes the distinguished endpoints of TT and can distribute representational capacity more evenly across all modeled modes.

The TR parameter count is
\begin{equation}
P_{\mathrm{TR}}
=
\sum_{n=1}^{N}R_{n-1}I_nR_n,
\qquad R_0=R_N.
\end{equation}
For uniform mode size $I$ and rank $R$, this is $O(NIR^2)$. The asymptotic form resembles TT, but TT has rank-one boundaries and therefore fewer parameters at the two endpoints. Removing those boundaries can improve flexibility at a fixed internal rank, although it does not guarantee better approximation, accuracy, or optimization.

The loop also introduces practical complications. A TT can be placed in a left-canonical, right-canonical, or mixed-canonical form around a selected core. A ring has no natural boundary at which to start this construction, and cutting the ring for optimization can produce a larger effective boundary rank. TR decompositions also have substantial gauge and cyclic-permutation nonuniqueness. These properties can make rank adaptation, initialization, and comparison across implementations less straightforward than the simple storage formula suggests.

Mode ordering remains important even though the topology is cyclic. Nearby cores communicate through one bond, whereas distant modes must interact through a sequence of bonds around the ring. For neural weights, reporting only a uniform ring rank omits the chosen tensorization, mode order, and contraction path. TR is most useful when the data or parameter interactions have an approximately cyclic organization or when TT boundary effects are undesirable. It is less attractive when a chain ordering has a natural beginning and end or when a mature TT kernel is substantially more efficient on the target hardware.

\section{Hierarchical Tucker}

Hierarchical Tucker (HT) recursively partitions the mode set with a dimension tree~\cite{kressner2012htucker}. Let the root subset be $S=\{1,\ldots,N\}$ with children $S_1$ and $S_2$. If $S_1$ has children $D_1,D_2$ and $S_2$ has children $D_3,D_4$, the first levels can be expressed as
\begin{align}
\mathcal{U}^{S}
&\approx
\mathcal{G}^{S}
\times_1^2\mathcal{U}^{S_1}
\times_1^2\mathcal{U}^{S_2},\\
\mathcal{U}^{S_1}
&\approx
\mathcal{G}^{S_1}
\times_1^2\mathcal{U}^{D_1}
\times_1^2\mathcal{U}^{D_2},\\
\mathcal{U}^{S_2}
&\approx
\mathcal{G}^{S_2}
\times_1^2\mathcal{U}^{D_3}
\times_1^2\mathcal{U}^{D_4}.
\label{eq:supp:ht}
\end{align}
Each internal transfer tensor combines the frames associated with its two children. The recursion continues until the leaves correspond to individual modes. For a balanced binary tree with uniform mode dimension $d$ and rank bounded by $R$, the commonly stated storage cost is $O(NdR+NR^3)$.

The dimension tree is the defining design choice of HT. Each internal node represents the subspace associated with a subset of modes, and its transfer tensor combines the subspaces of its two children. Modes that are grouped near the leaves can interact through short paths, while interactions between distant branches pass through higher-level ranks. The tree therefore encodes a hierarchy of conditional dependencies rather than only reducing storage~\cite{grasedyck2013literature}.

More generally, if leaf $n$ has rank $R_{\{n\}}$ and internal node $t$ with children $t_1,t_2$ has rank $R_t$, the parameter count has the form
\begin{equation}
P_{\mathrm{HT}}
=
\sum_{n=1}^{N}I_nR_{\{n\}}
+
\sum_{t\ \mathrm{internal}}
R_tR_{t_1}R_{t_2},
\end{equation}
with the root rank usually set to one for a single output tensor. A balanced tree limits path length to $O(\log N)$, but an unbalanced tree can approach a chain. Two HT models with the same rank bound can allocate capacity differently because their selected dimension trees differ.

Orthogonal frames at the nodes are often used to stabilize truncation and local optimization. They separate basis conditioning from the singular values associated with a tree edge and support quasi-optimal hierarchical truncation. In a neural implementation, maintaining exact orthogonality after every gradient update may be expensive, so alternatives include periodic reorthogonalization, manifold-aware updates, or soft orthogonality penalties. The appropriate choice depends on whether the factors are used primarily for compression, direct training, or repeated rank adaptation.

HT is well suited to data with known multiscale or grouped structure, such as variables arranged by spatial region, sensor hierarchy, or feature family. Its main modeling risk is a mismatched dimension tree. If strongly coupled modes are placed in distant branches, the upper-level ranks must grow to recover their interaction. Automated tree search and task-aware mode grouping are therefore natural extensions of rank selection.

\section{Projected Entangled Pair State}

A PEPS places cores on an $M\times N$ grid~\cite{biamonte2017tensor,schuch2007computational}. Let $h^{(R)}_{m,n}$ denote the horizontal bond between sites $(m,n)$ and $(m,n+1)$, and let $h^{(C)}_{m,n}$ denote the vertical bond between sites $(m,n)$ and $(m+1,n)$. The resulting elementwise representation is
\begin{align}
&\mathcal{X}_{i_{1,1},\ldots,i_{M,N}}
=
\sum_{\{h^{(R)}\},\{h^{(C)}\}}
\nonumber\\
&\qquad
\prod_{m=1}^{M}\prod_{n=1}^{N}
\mathcal{G}^{(m,n)}_{
i_{m,n},
h^{(R)}_{m,n-1},h^{(R)}_{m,n},
h^{(C)}_{m-1,n},h^{(C)}_{m,n}},
\label{eq:supp:peps}
\end{align}
where the missing bonds at the four grid boundaries have dimension one. An interior core therefore has one physical mode and four bond modes, while edge and corner cores have fewer nontrivial bonds. The product in Eq.~\eqref{eq:supp:peps} is contracted over all shared horizontal and vertical bond indices. The loops in this grid distinguish PEPS from chain and tree formats and make exact contraction \#P-hard for arbitrary lattice geometries~\cite{schuch2007computational}.

For a uniform square lattice with physical dimension $d$ and bond dimension $D$, an interior PEPS core has approximately $dD^4$ parameters. Boundary cores have fewer nontrivial bonds, so the total storage is $O(MNdD^4)$ for an $M\times N$ grid. This can still be exponentially smaller than the dense state or coefficient tensor, whose size is $d^{MN}$. The additional lattice bonds allow correlations to propagate through multiple paths and make PEPS a natural representation for two-dimensional spatial structure in physics-inspired learning problems~\cite{orus2014practical}.

Storage, however, is not the dominant difficulty. Exact contraction of a loopy two-dimensional network generally creates intermediates whose dimensions grow exponentially with a lattice width. Practical algorithms therefore contract approximately by truncating boundary MPS representations, using corner transfer matrices, applying tensor renormalization, or performing message passing. Each approach introduces additional approximation ranks and stopping criteria. A PEPS-based neural layer is consequently defined jointly by its cores and by the algorithm used for the forward and backward contractions.

Approximate contraction raises questions that do not occur in the same form for exactly contractible chains. Truncation changes the computed output, and differentiating through the truncated procedure can bias the gradient. The error can depend on input, parameter state, contraction direction, and hardware precision. Reproducible studies should report the PEPS bond dimensions, boundary or environment ranks, contraction ordering, truncation rule, numerical precision, and peak intermediate size. Reporting only the number of stored core parameters substantially understates the computational model.

PEPS is most compelling when the task has a genuine two-dimensional interaction pattern that a chain would represent only with large ranks. For ordinary dense-layer compression, the contraction overhead may outweigh this representational advantage. Hybrid networks can provide a compromise by using tree or chain backbones with a limited number of additional edges, or by restricting PEPS contractions to small local patches. These designs preserve some spatial inductive bias while bounding contraction width.

Taken together, BTD, MPO, TR, HT, and PEPS illustrate why a tensor format should not be selected by compression ratio alone. BTD controls capacity through blocks and multilinear ranks, MPO and TR emphasize sequential factorization, HT encodes a hierarchy, and PEPS prioritizes richer spatial connectivity. Their practical comparison requires the tensorization map, topology, rank profile, contraction algorithm, intermediate memory, and task performance to be evaluated together under application-specific constraints.

\section{Controlled CIFAR-10 Benchmark Protocol}

\subsection*{Purpose and Scope}

The benchmark is designed as a controlled illustration of the trade-off between predictive accuracy and several notions of efficiency. It is not intended to establish a state-of-the-art CIFAR-10 result or to rank all tensor formats. A small common architecture makes the parameter and operation counts transparent, permits repeated training within a reasonable resource budget, and exposes the difference between theoretical compression and measured GPU execution. Dense, Tucker-2, and CP layers are included because their convolutional implementations are direct and their parameter formulas can be independently checked.

The experiment complements the analytical example in the main article. The analytical table isolates rank and kernel geometry while holding the layer dimensions fixed. The task-level benchmark adds optimization, validation-based model selection, nonlinearities, memory allocation, framework overhead, and prediction accuracy. Neither example alone is sufficient: analytical counts are reproducible but cannot predict accuracy, whereas empirical timing without formulas can conceal differences in represented capacity.

\subsection*{Model Architecture and Factorized Layers}

The empirical example uses a three-convolution SimpleCNN with channel dimensions $3\rightarrow64\rightarrow128\rightarrow256$. Each convolution is followed by batch normalization and ReLU. Max pooling follows the second and third convolutions, reducing the spatial resolution from $32\times32$ to $16\times16$ and then to $8\times8$. Global average pooling produces a 256-dimensional feature vector, and a linear classifier maps it to the ten CIFAR-10 logits. The first $3\times3$ convolution remains dense in every configuration so that the low-dimensional RGB input is processed identically. Only the second and third convolutions are changed.

A dense $K\times K$ convolution from $C_i$ to $C_o$ channels stores $K^2C_iC_o$ kernel parameters. The Tucker-2 implementation applies a $1\times1$ projection from $C_i$ to $r_i$, a $K\times K$ convolution from $r_i$ to $r_o$, and a $1\times1$ expansion from $r_o$ to $C_o$. Its kernel count is
\begin{equation}
P_{\mathrm{T2}}=C_ir_i+K^2r_ir_o+r_oC_o.
\end{equation}
Rank ratio $0.25$ uses $(r_i,r_o)=(16,32)$ in the second convolution and $(32,64)$ in the third. Rank ratio $0.50$ uses $(32,64)$ and $(64,128)$ for the same two layers.

The CP implementation applies a $1\times1$ projection from $C_i$ channels to rank $R$, separate depthwise $K\times1$ and $1\times K$ spatial convolutions, and a final $1\times1$ expansion to $C_o$. Its kernel count is
\begin{equation}
P_{\mathrm{CP}}=R(C_i+C_o+2K).
\end{equation}
The benchmark uses $R=32$ or $R=64$ in both tensorized layers. All configurations are trained directly in their dense or factorized form. No method receives a pretrained dense checkpoint, post-training decomposition, or additional fine-tuning stage.

The parameter values in the main table count every trainable model parameter, including the common stem, batch-normalization affine parameters, and classifier. The compression ratio therefore compares complete models rather than only the two replaced kernels. This convention is more conservative than a layer-only compression ratio and makes clear how non-tensorized components limit end-to-end model compression.

\subsection*{Data Split and Optimization}

The standard CIFAR-10 training split is deterministically divided into $45{,}000$ training and $5{,}000$ validation images using a fixed split generator seed of 20260716. The original $10{,}000$ test images are not used for rank selection, hyperparameter selection, or checkpoint selection. Training images use random crops with four-pixel padding and random horizontal flips. Training, validation, and test images are normalized with the standard CIFAR-10 channel statistics.

Every configuration uses AdamW with learning rate $3\times10^{-3}$, weight decay $5\times10^{-4}$, cosine learning-rate decay, label smoothing $0.1$, batch size 256, and 20 epochs. The validation set is evaluated after every epoch, and the checkpoint with the highest validation accuracy is retained. That checkpoint is evaluated once on the test set. Seeds 11, 22, and 33 control Python, NumPy, CPU PyTorch, CUDA PyTorch, data-loader shuffling, and worker initialization. cuDNN benchmarking is disabled and deterministic cuDNN behavior is enabled. These settings reduce avoidable variation, although complete bitwise reproducibility is not guaranteed across different GPU models, drivers, or library versions.

Training uses automatic mixed precision. The execution environment is PyTorch 2.3.0, CUDA 12.1, cuDNN 8.9.2, and one Tesla V100 PCIe GPU with 32GB memory. The same software environment, batch sizes, augmentations, optimizer, schedule, and checkpoint rule are used for all five configurations. The ranks were chosen to cover moderate and strong compression regimes rather than to tune each format to the same parameter count.

\subsection*{Parameter, Arithmetic, Time, and Memory Metrics}

MACs count the convolutional and linear multiply-accumulate operations for one image by registering hooks on the corresponding PyTorch modules. The count includes grouped depthwise convolutions but excludes batch normalization, ReLU, pooling comparisons, memory movement, kernel launch overhead, and optimizer operations. It is therefore an arithmetic model, not a direct latency prediction. One MAC is treated as one multiply-accumulate operation; readers using a two-FLOP convention can multiply the reported MAC count by two.

Training time starts immediately before the first optimization epoch and ends after the final validation pass. It includes forward and backward computation, optimizer and scheduler updates, data loading during the epochs, and all 20 validation evaluations. It excludes initial dataset decoding, NumPy cache construction, model creation, and final test evaluation. Training peak memory is the maximum memory allocated by the PyTorch CUDA allocator after the model is placed on the GPU and before training begins. It includes model parameters, gradients, optimizer state, activations, and loaded batches, but it does not represent total board-level memory reserved by the CUDA context or driver.

Inference timing is intended to isolate model execution from storage and host-to-device input transfer. All test inputs are first preloaded onto the GPU. Three complete warm-up passes are executed, after which each run measures 20 complete forward passes through all $10{,}000$ test images. CUDA synchronization is applied around every measured pass, and the median of the 20 timings is retained for that run. The main table then reports the mean and sample standard deviation of the three run-level medians. Consequently, the reported time is the full-test-set latency rather than per-image latency or throughput under a streaming input pipeline.

\subsection*{Fairness, Reproducibility, and Limitations}

The comparison holds the macro-architecture, data split, optimizer, training budget, and hardware constant. It does not match every pair of methods by parameter count or MAC count because the purpose is to show several practically meaningful rank settings. A stricter format comparison could interpolate ranks to match parameters, tune learning rates separately, repeat on additional architectures, or compare post-training decomposition with direct factorized training. Those extensions would answer different questions and should not be inferred from this controlled example.

The benchmark is intentionally modest. Twenty epochs are enough to reveal stable relative trends in this architecture, but several runs select the final epoch, indicating that the experiment should not be interpreted as a convergence study. CIFAR-10 and SimpleCNN also do not capture the optimization and systems behavior of large vision models, Transformers, or distributed training. Measured timing is specific to the V100, PyTorch version, CUDA stack, batch size, and implementation. The robust conclusion is therefore qualitative: reductions in stored parameters and MACs can reduce memory and inference time, but the gains are not proportional and need not reduce end-to-end training time.

\section{Per-Seed CIFAR-10 Results}

Table~\ref{tab:supp:perseed} exposes the run-level measurements underlying the aggregate values in the main article. ``Best epoch'' is selected only from validation accuracy, while test accuracy is evaluated after checkpoint selection. Reporting the individual records prevents a small standard deviation or a favorable mean from concealing an outlying seed, and it makes the timing variability visible.

\begin{table*}[!t]
\centering
\caption{Per-seed records underlying the empirical summary, including total 20-epoch training and validation time, median full-test-set GPU inference time, and peak allocated training memory.}\label{tab:supp:perseed}
\begin{tabular}{lrrrrrr}
\hline
\textbf{Method} & \textbf{Seed} & \textbf{Best epoch} & \textbf{Test Acc.(\%)} & \textbf{Train(s)} & \textbf{Infer(ms)} & \textbf{Peak(MB)} \\
\hline
Dense & 11 & 20 & 78.80 & 119.2 & 342.6 & 1175 \\
Dense & 22 & 20 & 78.75 & 124.8 & 352.5 & 1176 \\
Dense & 33 & 20 & 78.78 & 139.1 & 255.7 & 1176 \\
Tucker-2 (0.25) & 11 & 20 & 72.62 & 156.8 & 167.4 & 666 \\
Tucker-2 (0.25) & 22 & 20 & 72.31 & 137.9 & 181.6 & 666 \\
Tucker-2 (0.25) & 33 & 20 & 72.84 & 115.6 & 115.9 & 666 \\
Tucker-2 (0.50) & 11 & 20 & 74.26 & 140.0 & 205.2 & 764 \\
Tucker-2 (0.50) & 22 & 20 & 73.88 & 139.2 & 282.5 & 764 \\
Tucker-2 (0.50) & 33 & 19 & 74.56 & 112.5 & 202.2 & 764 \\
CP ($R=32$) & 11 & 19 & 71.37 & 137.0 & 178.2 & 665 \\
CP ($R=32$) & 22 & 20 & 71.58 & 150.1 & 196.0 & 665 \\
CP ($R=32$) & 33 & 20 & 71.44 & 126.5 & 125.3 & 665 \\
CP ($R=64$) & 11 & 20 & 74.44 & 110.4 & 243.6 & 666 \\
CP ($R=64$) & 22 & 20 & 74.47 & 154.6 & 160.5 & 666 \\
CP ($R=64$) & 33 & 20 & 74.78 & 143.9 & 160.2 & 666 \\
\hline
\end{tabular}
\end{table*}

\subsection*{Accuracy and Compression Trends}

The test-accuracy variation across seeds is small relative to the gaps between the dense and factorized models. The sample standard deviations are $0.03$ percentage points for Dense, $0.27$ and $0.34$ for the two Tucker-2 settings, and $0.11$ and $0.19$ for the two CP settings. The dense-to-factorized accuracy differences, which range from $4.21$ to $7.31$ points, are therefore not explained by one unusually favorable or unfavorable seed in this experiment.

Increasing factorized capacity improves accuracy but weakens compression. Tucker-2 improves from $72.59\%$ at rank ratio $0.25$ to $74.23\%$ at ratio $0.50$, while compression decreases from $6.94\times$ to $2.52\times$. CP improves from $71.46\%$ at rank 32 to $74.56\%$ at rank 64, while compression decreases from $15.57\times$ to $8.73\times$. This is the empirical counterpart of the analytical rank-sensitivity example: rank is a capacity parameter, and compression must be reported together with its associated accuracy.

Within the tested settings, CP rank 64 reaches slightly higher mean accuracy than Tucker-2 ratio $0.50$ while using fewer parameters and MACs. This observation should not be generalized into a claim that CP is universally superior. The two settings are not parameter matched, the optimization hyperparameters were shared rather than separately tuned, and different architectures may favor the denser interaction core retained by Tucker-2.

\subsection*{Execution and Memory Trends}

All four factorized models reduce the measured full-test inference time relative to Dense on average, but the speedups are much smaller than their MAC reductions. Tucker-2 ratio $0.25$ has the lowest mean inference time, $155.0$ ms, corresponding to a $2.05\times$ speedup over Dense despite a theoretical MAC reduction of approximately $7.04\times$. CP rank 64 reduces MACs by $7.23\times$ but improves mean inference time by only $1.69\times$. Small sequential convolutions, kernel launch overhead, intermediate tensors, and limited arithmetic intensity help explain this gap.

Inference timing also varies more than accuracy across seeds. Although model structure is fixed within each configuration, GPU timing can be influenced by runtime scheduling, thermal state, and other system-level effects. Reporting the run-level medians and their sample standard deviation makes this variation explicit. The timing values should be compared as measurements from the documented environment rather than as architecture-independent constants.

The factorized models do not reduce total training time in this protocol. Their mean training times range from $130.6$ to $137.9$ seconds, compared with $127.7$ seconds for Dense. Backpropagation through several factorized operations, additional kernel launches, validation variation, and optimizer bookkeeping can offset the reduction in arithmetic. This result is important because parameter compression is sometimes presented as if it implied proportional training acceleration.

Peak allocated training memory is more stable across seeds and shows a clearer benefit. Dense uses approximately $1176$ MB, Tucker-2 ratio $0.50$ uses $764$ MB, and the more compressed Tucker-2 and CP settings use approximately $665$ to $666$ MB. These measurements include activations and optimizer state, so the reduction is smaller than the parameter compression ratio but remains operationally meaningful for resource-constrained training.

\subsection*{Checkpoint and Statistical Interpretation}

Most runs select epoch 20, with only Tucker-2 ratio $0.50$ at seed 33 and CP rank 32 at seed 11 selecting epoch 19. This consistency supports the use of a common validation rule, but it also shows that the 20-epoch budget often ends near the best observed checkpoint. Longer training or format-specific schedules could change the absolute accuracies and possibly the ordering. The present experiment is a fixed-budget comparison rather than a convergence study.

The reported standard deviations summarize three independent training seeds and should not be interpreted as narrow confidence intervals for all possible implementations or hardware. The benchmark controls one source of uncertainty, stochastic training, while leaving architecture choice, rank selection, software kernels, and device behavior fixed. Broader empirical claims would require more seeds, additional datasets, deeper models, parameter-matched settings, and measurements across multiple accelerator platforms.

% \begin{IEEEbiography}[{\includegraphics[width=1in,height=1.25in,clip,keepaspectratio]{mshell}}]{Michael Shell}
% % or if you just want to reserve a space for a photo:

% % \begin{IEEEbiography}{Michael Shell}
% % Biography text here.
% \end{IEEEbiography}


% % insert where needed to balance the two columns on the last page with
% \newpage


% % You can push biographies down or up by placing
% % a \vfill before or after them. The appropriate
% % use of \vfill depends on what kind of text is
% % on the last page and whether or not the columns
% % are being equalized.

% \vfill

% % Can be used to pull up biographies so that the bottom of the last one
% % is flush with the other column.
% %\enlargethispage{-5in}

\end{document}